\documentclass[pdflatex,sn-apa, iicol]{sn-jnl}

\usepackage{mathtools} 
\usepackage{amsmath} 

\usepackage{booktabs}       
\usepackage{amsfonts}       
\usepackage{nicefrac}       
\usepackage{microtype}      
\usepackage{xcolor}         
\usepackage{multirow}
\usepackage{tabularx}
\usepackage{makecell}
\usepackage{graphicx}
\usepackage{caption}
\usepackage{wrapfig}
\usepackage{lipsum}
\usepackage{colortbl}
\usepackage{makecell, multirow}
\newcolumntype{L}{>{\raggedright\arraybackslash}X}

\usepackage[capitalize,noabbrev]{cleveref}

\usepackage{pifont}
\newcommand{\cmark}{\ding{51}}%
\newcommand{\xmark}{\ding{55}}%

\usepackage{array}
\newcolumntype{L}[1]{>{\raggedright\let\newline\\\arraybackslash\hspace{0pt}}m{#1}}
\newcolumntype{C}[1]{>{\centering\let\newline\\\arraybackslash\hspace{0pt}}m{#1}}
\newcolumntype{R}[1]{>{\raggedleft\let\newline\\\arraybackslash\hspace{0pt}}m{#1}}

\usepackage{colortbl}
\definecolor{Orange}{rgb}{0.9,0.5,0}
\definecolor{NavyBlue}{rgb}{0.1, 0.4, 0.8}
\definecolor{Magenta}{rgb}{0.8, 0.1, 0.6}
\definecolor{F7E0D5}{RGB}{247,224,213}
\colorlet{Light}{white!0!F7E0D5}

\definecolor{cb-black}      {RGB}{  0,   0,   0}
\definecolor{cb-blue-green} {RGB}{  0,  073,  073}
\definecolor{cb-green-sea}  {RGB}{  0, 146, 146}
\definecolor{cb-rose}       {RGB}{255, 109, 182}
\definecolor{cb-salmon-pink}{RGB}{255, 182, 119}
\definecolor{cb-purple}     {RGB}{ 73,   0, 146}
\definecolor{cb-blue}       {RGB}{ 0, 109, 219}
\definecolor{cb-lilac}      {RGB}{182, 109, 255}
\definecolor{cb-blue-sky}   {RGB}{109, 182, 255}
\definecolor{cb-blue-light} {RGB}{182, 219, 255}
\definecolor{cb-burgundy}   {RGB}{146,   0,   0}
\definecolor{cb-brown}      {RGB}{146,  73,   0}
\definecolor{cb-clay}       {RGB}{219, 209,   0}
\definecolor{cb-green-lime} {RGB}{ 36, 255,  36}

\hypersetup{colorlinks = true, linkcolor = NavyBlue,
            urlcolor  = gray,
            citecolor = NavyBlue,
            anchorcolor = NavyBlue}

\newcommand{\posimprov}[1]{\scriptsize{(\textcolor{teal}{+#1})}}
\newcommand{\negimprov}[1]{\scriptsize{(\textcolor{brown}{-#1})}}
\newcommand{\negimprovgreen}[1]{\scriptsize{(\textcolor{teal}{-#1})}}

\usepackage{setspace}
\usepackage[frozencache]{minted}
\usepackage{cuted}
\usepackage{appendix}

\jyear{2022}%

\theoremstyle{thmstyleone}%
\theoremstyle{thmstyletwo}%

\theoremstyle{thmstylethree}%

\begin{document}

\title[Spatial Entropy]{Spatial Entropy as an Inductive Bias for Vision Transformers}


\author[1]{\fnm{Elia} \sur{Peruzzo}}

\author*[2]{\fnm{Enver} \sur{Sangineto}}\email{enver.sangineto@unimore.it}\equalcont{These authors contributed equally to this work.}

\author[1,3]{\fnm{Yahui} \sur{Liu}}\equalcont{These authors contributed equally to this work.}

\author*[3]{\fnm{Marco} \sur{De Nadai}}

\author[4]{\fnm{Wei} \sur{Bi}}

\author[3]{\fnm{Bruno} \sur{Lepri}}

\author[1]{\fnm{Nicu} \sur{Sebe}}

\affil[1]{\orgname{University of Trento},  \country{Italy}}

\affil[2]{\orgname{University of Modena and Reggio Emilia}, \country{Italy}}

\affil[3]{\orgname{Bruno Kessler Foundation}, \country{Italy}}

\affil[4]{\orgname{Tencent AI Lab}, \country{China}}


\abstract{Recent work on Vision Transformers (VTs) 
showed that introducing a local inductive bias in the VT {\em architecture} helps reducing the number of  samples necessary for training.
However, the architecture modifications lead to a loss of generality of the Transformer backbone, partially contradicting the push towards the development of uniform architectures, shared, e.g., by both the Computer Vision and the Natural Language Processing areas.  
In this work, we propose a different and complementary direction, in which a local bias is introduced using {\em an auxiliary self-supervised task}, performed jointly with standard supervised training.
Specifically, we exploit the observation that
the attention maps of VTs, when trained with self-supervision,  can contain a semantic segmentation structure which does not spontaneously emerge when training is supervised. 
Thus, we
{\em explicitly} encourage the emergence of this spatial clustering as a form of training regularization.
In more detail, we 
exploit the assumption that, in a given image, objects usually correspond to  few connected regions, and we
propose 
a spatial formulation of the information entropy to quantify this {\em object-based inductive bias}.
 By minimizing the proposed spatial entropy, we 
 include an additional self-supervised signal
 during training. Using extensive experiments, we show that the proposed regularization 
 leads to equivalent or better results than other VT proposals which include a local bias by changing the basic Transformer architecture, and it 

can drastically boost the VT final accuracy when using small-medium training sets.
 The code is available at \href{https://github.com/helia95/SAR}{https://github.com/helia95/SAR}.}





\maketitle

\section{Introduction}
\label{sec:intro}

 \begin{figure*}[ht]
    \centering
\includegraphics[width=0.55\linewidth]{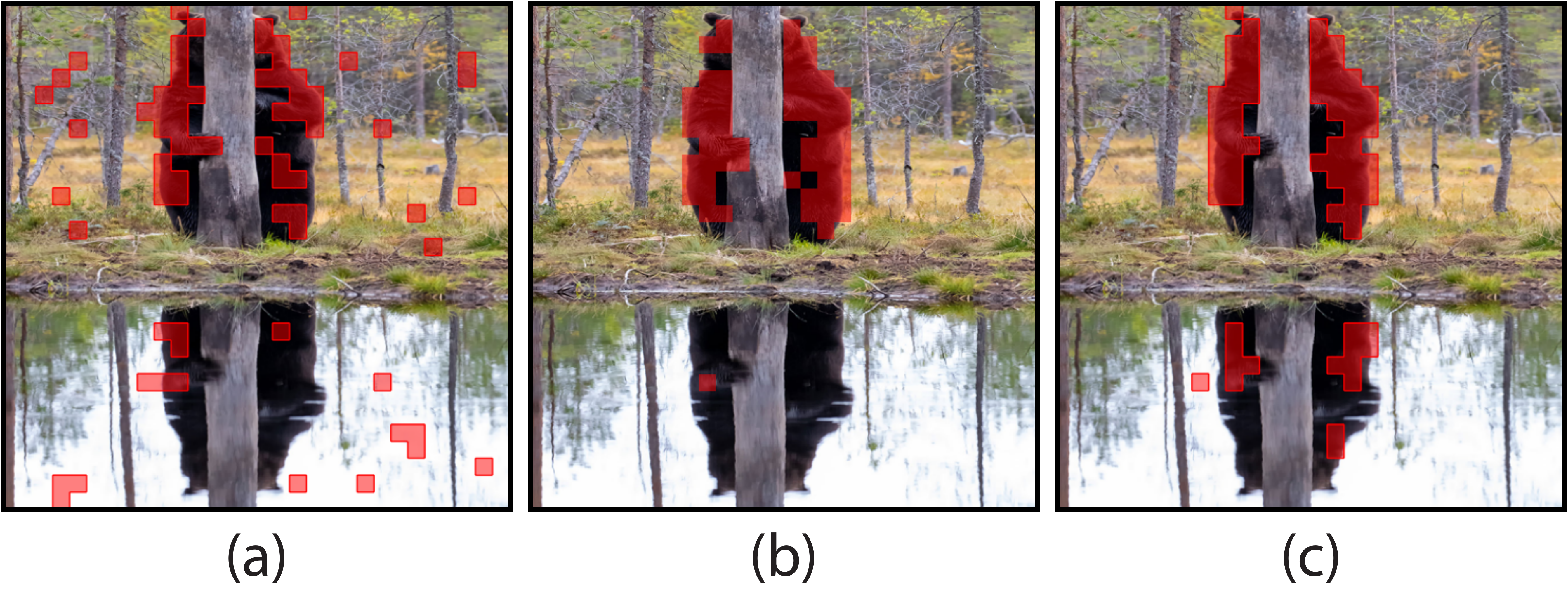}
\caption{ViT attention maps obtained using the \texttt{[CLS]} token query and thresholded  to keep $60\%$ of the mass following \citep{DINO}. 
     (a) Standard supervised learning. 
      (b) DINO. 
      (c) Training using  SAR. 
      In (c), the two largest connected components correspond to the main object (a bear, occluded by a tree), while the third and the fourth largest connected components correspond to the specular reflection of the bear into the river.
      }
    \label{fig:teaser}
\end{figure*}

Vision Transformers (VTs) are increasingly emerging as the dominant Computer Vision architecture, alternative to standard Convolutional Neural Networks (CNNs). 
VTs are inspired by the 
Transformer network \citep{attention-is-all-you-need}, which is the \emph{de facto} standard in Natural Language Processing (NLP) 
\citep{devlin-etal-2019-bert,Radford2018ImprovingLU} and it is based on 
 multi-head attention layers transforming the input {\em tokens} (e.g., language words) into a set of final embedding tokens.
 \cite{ViT}  proposed an analogous processing paradigm, called ViT\footnote{In this paper, we use VT to refer to generic Vision Transformer archiectures, and ViT to refer to the specific architecture proposed in \citep{ViT}.}, where word tokens are replaced by image patches, and self-attention layers are used to model global pairwise dependencies over all the input tokens. As a consequence, differently from CNNs, where the convolutional kernels have a spatially limited receptive field,
 ViT  has a dynamic receptive field, which is given by its attention maps \citep{IntriguingProperties-of-VTs}. 
However, ViT heavily
relies  on huge training datasets
(e.g., 
JFT-300M \citep{ViT}, a proprietary dataset of 303 million images),
and underperforms CNNs when trained on ImageNet-1K ($\sim$ 1.3 million images)~\citep{imagenet} or using smaller datasets \citep{ViT,do-VTs-see-like-CNNs}. To mitigate the need for a huge quantity of  training data, a recent line of research is exploring the possibility of reintroducing typical CNN mechanisms
in VTs \citep{T2T,Swin,CvT,CeiT,CoaT,LocalViT,GAT,DBLP:journals/corr/abs-2204-07143}. The main idea behind these ``hybrid'' VTs is that convolutional layers, mixed with the VT self-attention layers,  help   to embed a {\em local inductive bias} in the VT {\em architecture}, i.e., to encourage the network to focus on local properties of the image domain.
However, 
the disadvantage of this  paradigm is that
it requires
drastic architectural changes to the original ViT, where the latter has now become a \emph{de facto} standard in different vision and vision-language tasks (\cref{sec:RelatedWork}). 
 Moreover, as emphasised by \cite{ViTDet}, one of the main advantages of CNNs is the 
large independence of the pre-training from the downstream tasks, which allows to use a uniform backbone, pre-trained only once for different vision tasks (e.g., classification, object detection, etc.). Conversely, the adoption of specific VT architectures breaks this
independence and makes it difficult to use the same pre-trained backbone for different downstream tasks \citep{ViTDet}.

In this paper, we follow an orthogonal (and relatively simpler) direction:
rather than changing the ViT architecture, we propose to include a local inductive bias using an additional {\em pretext task} during training, 
which 
implicitly ``teaches'' the network 
the connectedness property
of  objects in an image.
Specifically, we maximize the probability of producing attention maps whose highest values are clustered in  few local regions (of {\em variable} size),
based on the idea that, most of the time, an object is represented by one or very few spatially connected  regions in the input image.
This  pretext task exploits a locality principle, characteristic of the natural images, and extracts additional (self-supervised) 
information from images without the need of architectural changes.

Our work is inspired by the findings presented in \citep{DINO,BEIT,IntriguingProperties-of-VTs}, in which the authors show that VTs, trained using self-supervision \citep{DINO,BEIT} or shape-distillation \citep{IntriguingProperties-of-VTs}, can spontaneously  develop attention maps with a semantic segmentation structure.  For instance, \cite{DINO} show  that the last-layer attention maps of ViT, when this is trained with their self-supervised DINO algorithm, 
 can be thresholded and used to segment the most important foreground objects of the input image  without any pixel-level annotation during training
 (see Fig.~\ref{fig:teaser} (b)).
 Similar findings are shown in \citep{BEIT,IntriguingProperties-of-VTs}.
Interestingly, however, \cite{DINO} show that the same ViT architectures, when trained with supervised methods,
produce much more spatially disordered attention maps (Fig.~\ref{fig:teaser} (a)). 
This  is confirmed by \cite{IntriguingProperties-of-VTs}, who  observed that the attention maps of ViT, trained with  supervised protocols, have a widely spread structure over the whole image.
The reason why ``blob''-like attention maps {\em spontaneously} emerge when VTs are trained with some 
algorithms but not with others, is still unclear.
However, in this paper we build on top of these findings and we propose 
a {\em spatial entropy} loss function  which {\em explicitly} encourages the emergence of  locally structured attention maps
(\cref{fig:teaser} (c)), independently of the
main algorithm used for training.
Note that our goal is {\em not} to use the so obtained attention maps to extract 
 segmentation structure from images or for other post-processing steps. Instead, we use the proposed 
spatial entropy loss
to introduce an {\em object-based local inductive bias} in VTs. Since real life objects usually correspond to one or very few connected image regions, then also the corresponding attention maps of a VT head should focus most of their highest values on spatially clustered regions. The possible discrepancy between this inductive bias (an object in a given image is a mostly connected structure) and the actual spatial entropy measured in each VT head, provides a {\em self-supervised} signal which 
 alleviates the need for huge supervised training datasets,
without changing the  ViT architecture.

The second contribution of this paper is based on the empirical results 
  presented by \cite{do-VTs-see-like-CNNs}, who showed that 
VTs are more influenced by the skip connections than CNNs and, specifically, that
in the last blocks of ViT,
the {\em patch tokens}
 (see \cref{sec:Preliminaries})
representations are mostly influenced by the skip connection path. This means that,
in the last blocks of ViT,
the self-attention layers have a relatively small influence on the final token embeddings. 
Since our 
spatial entropy is measured on the last-block  attention maps, we propose to remove the skip connections in  the last layer (only). We empirically show that this minor architectural change is beneficial for ViT, 
both when used jointly with our spatial entropy loss, and when used with a standard training procedure.

Our regularization method, 
which we call  
 SAR (Spatial Attention-based Regularization),
can be easily plugged into existing VTs 
(including hybrid VTs)
without drastic architectural changes of their architecture,  and it can be applied to different scenarios, jointly with a main-task loss function. For instance, when used in a supervised classification task, the main loss  is the (standard) cross entropy, used jointly with our 
spatial entropy loss. 
We empirically show that SAR is sample efficient: it can 
significantly 
 boost the accuracy of ViT in different downstream tasks such as classification, object detection or segmentation, and it is particularly useful when 
 training  with small-medium datasets.
Our experiments show also that SAR is beneficial when plugged on  hybrid VT architectures, especially  when the training data are scarce.

In summary, our main contributions are the following:
(1) We propose to embed a local inductive bias in ViT using 
 spatial entropy  
 as an alternative to re-introducing convolutional mechanisms in the VT architecture.
(2) We propose to remove the last-block skip connections, empirically showing that this is beneficial for the patch token representations.
(3) Using extensive experiments, we show that SAR improves the accuracy of different VT architectures, 
and it is particularly helpful when supervised training data are scarce.

\section{Related work}
\label{sec:RelatedWork}

{\bf Vision Transformers.}
One of the very first fully-Transformer architectures for Computer Vision is
 iGPT \citep{iGPT}, in which each image pixel is represented as a token. However, due to the quadratic computational complexity of Transformer networks \citep{attention-is-all-you-need}, iGPT can only operate with very small resolution images. This problem has been largely alleviated by ViT \citep{ViT}, where the input tokens are $p \times p$ image patches
 (\cref{sec:Preliminaries}). The success of ViT
 has  inspired 
 several similar 
 Vision Transformer (VT) 
architectures  in different application domains, such as
image classification \citep{ViT,DeiT,T2T,Swin,CvT,CeiT,LocalViT,CoaT,ConViT}, object detection \citep{ViTDet,DETR,Deformable-DETR,UP-DETR}, segmentation \citep{strudel2021segmenter,DenseCLIP},
human pose estimation \citep{zheng20213d},
object tracking \citep{TrackFormer},
video processing \citep{DBLP:journals/corr/abs-2102-00719,DBLP:journals/corr/abs-2112-01526},
image generation \citep{TransGAN,GAT,DALL-E-2,DBLP:journals/corr/abs-2202-04200}, point cloud processing \citep{DBLP:journals/cvm/GuoCLMMH21,PointTransformer}, 
vision-language foundation models \citep{DBLP:journals/corr/abs-2206-01127,DBLP:journals/corr/abs-2204-14198,DBLP:conf/iclr/WangYYDT022}
and many others.
However, the lack of the typical CNN local inductive biases makes VTs to need more  data for training
\citep{ViT,do-VTs-see-like-CNNs}.
For this reason, many recent works are addressing this problem by
 proposing hybrid architectures, which
reintroduce typical convolutional 
mechanisms into the VT design 
\citep{T2T,Swin,CvT,CeiT,CoaT,LocalViT,ConViT,GAT,DBLP:journals/corr/abs-2112-01526,DBLP:journals/corr/abs-2204-07143}. 
In contrast, we propose a different and simpler solution, in which, rather than changing the VT architecture, we introduce an  
object-based local inductive bias  (\cref{sec:intro}) by means of a pretext task based on the spatial entropy minimization. 

Note that our proposal is different from object-centric learning \citep{SlotAttention,DBLP:journals/corr/abs-2006-16225,DBLP:conf/nips/DidolkarGKBBHMB21,Genesisv2,ObjectSceneRepresentation,ObjectRegionVideoTransformer,ReMixer}, 
where the goal is to use discrete objects (usually obtained using a pre-trained object detector or a segmentation approach) for object-based reasoning and modular/causal inference. In fact, despite the attention maps produced by SAR can potentially be thresholded and used as discrete objects,
our goal is not to segment the image patches or to use the patch clusters for  further processing steps, but to exploit the clustering process as an additional self-supervised loss function which helps   to reduce the need of labeled training samples (\cref{sec:intro}).    


{\bf Self-supervised learning.}
Most of the self-supervised approaches with still images and ResNet backbones \citep{he2016deep} 
impose a semantic consistency between different views of the same image,
where the views are obtained with data-augmentation techniques.
This work can be roughly grouped
in  contrastive learning \citep{CPC,DIM,simclr,MoCo,tian2019contrastive,wang2020understanding,little-friends}, clustering methods 
\citep{NIPS2016_65fc52ed,DBLP:journals/corr/abs-1903-12355,IIC,DeepClustering,DBLP:conf/iclr/AsanoRV20a,DBLP:conf/eccv/GansbekeVGPG20,SwAV,DINO}, 
asymmetric networks   \citep{byol,simsiam} and feature-decorrelation methods \citep{w-mse,barlow,VICReg,hua2021feature}. 

Recently, different works  use VTs for self-supervised learning.
For instance,  \cite{MoCo-v3} have empirically tested different representatives of the above categories using VTs, and they also proposed MoCo-v3, a contrastive approach based on MoCo \citep{MoCo} but without the queue of the  past-samples.   
 DINO  \citep{DINO} is an on-line clustering method which is one of the current state-of-the-art self-supervised approaches using VTs.
BEiT \citep{BEIT} adopts the typical ``masked-word'' NLP pretext task 
\citep{devlin-etal-2019-bert}, 
but it needs to pre-extract a vocabulary of visual words
using the discrete VAE pre-trained in \citep{DALL-E}.
Other  recent works which use a ``masked-patch''  pretext task are \citet{DBLP:journals/corr/abs-2111-06377,DBLP:journals/corr/abs-2111-09886,DBLP:journals/corr/abs-2112-09133,DBLP:journals/corr/abs-2111-12710,RandSAC,DBLP:journals/corr/abs-2202-03026,MultiMAE,DBLP:journals/corr/abs-2112-10740,DBLP:journals/corr/abs-2111-07832,What-to-Hide}.

\cite{PatchLevelRepresentation} use the assumption that adjacent patches usually belong to the same object in order to collect positive patches for a contrastive learning approach. Our inductive bias shares a similar intuitive idea but, rather than   a contrastive method where positive pairs are compared with negative ones, our self-supervised loss is based on the proposed spatial entropy which groups together patches 
even when they are not pairwise adjacent.
Generally speaking, in this paper, we do not propose a fully self-supervised algorithm, but we rather use self-supervision (we extract information from samples without additional manual annotation) to speed-up the convergence in a supervised scenario and decrease the quantity of annotated information needed for training. In the Appendix, we also show that SAR can be plugged on top of both MoCo-v3 and DINO, boosting the accuracy of both of them. 

Similarly to this paper, \cite{OurNIPS} propose a VT  regularization approach based on predicting the geometric distance between patch tokens.  In contrast, we use the highest-value connected regions in the VT attention maps to extract additional unsupervised  information from images and the two regularization methods can potentially be used jointly.
\cite{DBLP:journals/pami/LiWPEF20}  compute the gradients of a ResNet  with respect to the image pixels to get an attention (saliency)  map. This map is thresholded 
and used to mask-out
the most salient pixels. Minimizing  the classification loss on this masked image  encourages the attention on the non-masked image to include most of the useful information. Our approach is radically different and much simpler, because we do not need to manually set the thresholding value  and we require only one forward and one backward pass per image.


\noindent{\bf Spatial entropy.}
There are many definitions of spatial entropy \citep{Razlighi2009ACS,altieri2018spatentropy}. For instance,  \cite{Batty-spatial-entropy} normalizes
 the probability of an event occurring in a given zone by the area of that zone, this way accounting for unequal space partitions. In 
\citep{901061}, spatial entropy is defined over a Markov Random Field describing the image content, but its computation is very expensive \citep{Razlighi2009ACS}. In contrast, our  spatial entropy loss  can be efficiently computed and it is differentiable, thus it can be easily used as an auxiliary regularization task in existing VTs.

\section{Background}
\label{sec:Preliminaries}

Given an input image $I$, ViT \citep{ViT}  splits $I$ in a grid of $K \times K$ non-overlapping patches, and each patch is linearly projected into a (learned) input embedding space. The input of ViT is this set of $n = K^2$ {\em patch tokens}, jointly with a special token, called  \texttt{[CLS]} token, which is used to represent the whole image. 
Following a standard Transformer network \citep{attention-is-all-you-need}, ViT  transforms these $n + 1$ tokens in corresponding 
final $n + 1$ token embeddings using a sequence of $L$ Transformer blocks. Each block is composed of LayerNorm (LN), Multiheaded Self Attention
(MSA) and MLP layers, plus skip connections. Specifically, if the token embedding sequence in  the $(l-1)$-th layer is
$\pmb{z}^{l-1} = [ \pmb{z}_{CLS}; \pmb{z}_1; ... \pmb{z}_n ]$, then:
\begin{align}
    \pmb{z}' = \mbox{MSA}(\mbox{LN}(\pmb{z}^{l-1})) + \pmb{z}^{l-1} , && l=1,\ldots, L \label{eq:transformer-standard-1}\\
    \pmb{z}^l = \mbox{MLP}(\mbox{LN}(\pmb{z}')) + \pmb{z}' , && l=1,\ldots, L \label{eq:transformer-standard-2}
\end{align}
\noindent
where the addition ($+$) denotes a skip (or ``identity'')
connection, which is used both in the MSA (Eq.~\ref{eq:transformer-standard-1})
and in the MLP (Eq.~\ref{eq:transformer-standard-2}) layer.
The MSA layer is composed of $H$ different heads, and, in the $h$-th head ($1 \leq h \leq H$), each token embedding
$\pmb{z}_i \in \mathbb{R}^d$ is projected into a query ($\pmb{q}_i^h$), a key  ($\pmb{k}_i^h$) and a value ($\pmb{v}_i^h$). 
Given query ($Q^h$), key ($K^h$) and value ($V^h$) matrices containing the corresponding elements, the $h$-th self-attention matrix ($A^h$) is given by:
\begin{equation}
\label{eq:attention-standard}
A^h = \mbox{softmax} \left( \frac{Q^h (K^h)^T}{\sqrt{d}} \right).
\end{equation}
\noindent
Using $A^h$, each head outputs a weighted sum of the values in $V^h$.
The final MSA layer output is obtained by concatenating all the head outputs and then projecting each token embedding into a $d$-dimensional space.
Finally, the last-layer ($L$) class token embedding $\pmb{z}_{CLS}^{L}$ is fed to
an MLP  head, which computes a posterior distribution over the set of the target classes and the whole network is trained using a standard cross-entropy loss
($\mathcal{L}_{ce}$). 
Some hybrid VTs (see \cref{sec:RelatedWork}) such as CvT \citep{CvT} and PVT \citep{PVT},  progressively subsample the number of patch tokens, leading to a final $k \times k$ patch token grid ($k \leq K$). In the rest of this paper, we generally refer to a spatially arranged grid of final patch token embeddings with a $k \times k$ resolution.

\section{Method}
\label{sec:Method}

Generally speaking, an object  usually corresponds to one or very few connected regions of a given image. For instance, the bear in \cref{fig:teaser}, despite being occluded by a tree, occupies only 2 distinct connected regions of the image.
Our goal is to exploit this natural image inductive bias and  penalize those attention maps which do not lead to a spatial clustering of their largest values.
Intuitively, if we compare \cref{fig:teaser} (a) with \cref{fig:teaser} (b), we observe that, in the latter case 
(in which DINO  was used for training), 
the  attention maps are more ``spatially ordered'', i.e. there are {\em less} and {\em bigger} 
``blobs'' (obtained after thresholding the  map values \citep{DINO}).
Since an image is usually composed of a few main objects, each of which typically corresponds to one or very few connected regions  of tokens,
during training we penalize those attention maps which produce a large number of small blobs.
We use this as an auxiliary pretext task which extracts information from  images without additional annotation, by exploiting the assumption that
spatially close tokens should preferably belong to the same cluster.

\subsection{Spatial entropy loss}
\label{sec:Spatial-entropy}

For each head  of the last Transformer block, we compute a similarity map $S^h$ ($1 \leq h \leq H$, see \cref{sec:Preliminaries}) by comparing the \texttt{[CLS]}  token query ($\pmb{q}_{CLS}^h$) with all the patch token keys ($\pmb{k}_{x,y}^h$, where $(x,y) \in \{1,...,k\}^2$):
\begin{equation}
\label{eq:similarity-map}
S_{x,y}^h = <\pmb{q}_{CLS}^h, \pmb{k}_{x,y}^h>/\sqrt{d}, \quad (x,y) \in \{1,...,k\}^2,
\end{equation}
\noindent
where $<\pmb{a}, \pmb{b}>$ is the dot product between  $\pmb{a}$ and $\pmb{b}$.
$S^h$ is extracted from  the self-attention map $A^h$ 
by selecting  the \texttt{[CLS]} token as the only query
and before applying the $\mbox{softmax}$ 
(see \cref{sec:Discussion} for a discussion about this choice).
$S^h$ is a $k \times k$ matrix corresponding to the final  $k \times k$ spatial grid of patches
(\cref{sec:Preliminaries}), and
 $(x,y)$ corresponds to the ``coordinates'' of a patch token in this grid.

\begin{figure*}[t]
\centering
\includegraphics[width=0.78\linewidth]{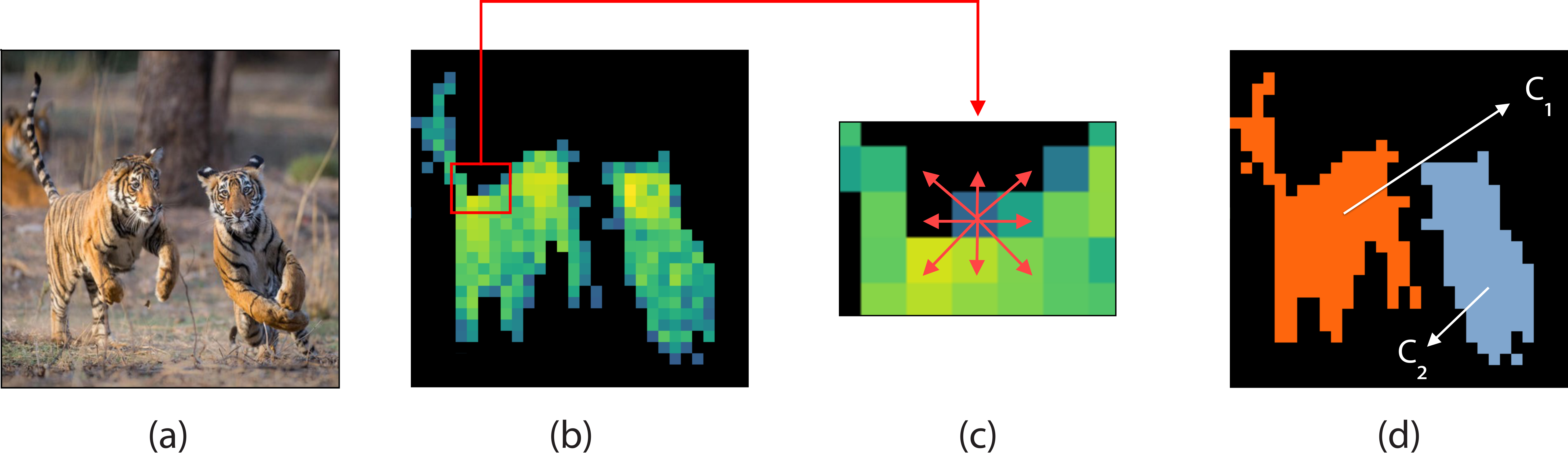}
\caption{A schematic illustration of the spatial entropy. (a) The original image. (b) The thresholded similarity map $B^h$
(zero values shown in black). 
(c) The 8-connectivity
relation used to group non-zero elements in $B^h$.
(d) The resulting two connected components ($C_1$ and $C_2$). 
}
\label{fig:scheme}
\end{figure*}

In order to extract  a set of connected components containing the largest values in $S^h$,  we zero-out those elements of $S^h$
which are smaller than the mean value $m = 1/n \sum_{(x,y) \in \{1,...,k\}^2} S^h_{x,y}$: 
\begin{equation}
\label{eq:binary-map}
B_{x,y}^h = \mbox{ReLU}(S_{x,y}^h - m), \quad (x,y) \in \{1,...,k\}^2,
\end{equation}
\noindent
where  thresholding using $m$
 corresponds to retain half of the  total ``mass'' of Eq.~\ref{eq:similarity-map}. 
 We can now use a standard algorithm \citep{DBLP:journals/tip/GranaBC10} 
to extract the connected components from $B^h$, obtained using an 8-connectivity
relation between non-zero elements in 
$B^h$ (see \cref{fig:scheme}): 
\begin{equation}
\label{eq:connected-components}
C^h = \{C_1, ..., C_{h_r} \} = \mbox{ConnectedComponents}(B^h).
\end{equation}
\noindent
$C_j$ ($1 \leq j \leq h_r$)
in $C^h$ is the set of coordinates ($C_j = \{ (x_1, y_1), ..., (x_{n_j}, y_{n_j})\}$) 
of the $j$-th connected component, whose cardinality ($n_j$) is variable, and such is the total number of components ($h_r$).
Given $C^h$, we  define the spatial entropy of $S^h$ ($\mathcal{H}(S^h)$) as follows:
\begin{equation}
\label{eq:spatial-entropy}
\mathcal{H}(S^h) = -  \sum_{j=1}^{h_r} P^h (C_j) \log P^h (C_j),
\end{equation}
\noindent
\begin{equation}
\label{eq:single-component-probability}
P^h (C_j) =  \frac{1}{\lvert B^h \rvert}  \sum_{(x,y) \in C_j} B_{x,y}^h,
\end{equation}
\noindent
where $\lvert B^h \rvert = \sum_{(x,y) \in \{1,...,k\}^2} B_{x,y}^h$.
Importantly, in Eq.~\ref{eq:single-component-probability}, the probability of each  region ($P^h (C_j)$)
is computed using all its elements, and this makes the difference with respect to a non-spatial entropy which is directly computed over all the individual elements in 
$S^h$, without considering the connectivity relation. Note that the less the number of components $h_r$ or the less uniformly distributed the probability values 
$P^h (C_1), ... P^h (C_{h_r})$, the lower $\mathcal{H}(S^h)$.
Using Eq.~\ref{eq:spatial-entropy}, the spatial entropy loss is defined as:
\begin{equation}
\label{eq:spatial-entropy-loss}
\mathcal{L}_{se} = \frac{1}{H} \sum_{h=1}^H \mathcal{H}(S^h) .
\end{equation}

$\mathcal{L}_{se}$ is used jointly with the main task loss. For instance, in case of supervised training, we use: 
$\mathcal{L}_{tot} = \mathcal{L}_{ce} + \lambda \mathcal{L}_{se}$, where $\lambda$ is a weight used to balance the two losses.

\subsection{Removing the skip connections}
\label{sec:skip-connections}

 \cite{do-VTs-see-like-CNNs} empirically showed that, {\em in the last blocks} of ViT, 
the patch token representations are 
mostly  propagated from the previous layers using 
the skip connections  (\cref{sec:intro}). 
We presume this is (partially) due to the fact that only the \texttt{[CLS]} token is used as input to the classification MLP head (\cref{sec:Preliminaries}), thus, during training, the last-block patch token embeddings are usually neglected. 
Moreover, \cite{do-VTs-see-like-CNNs}  show that the effective receptive field \citep{DBLP:journals/corr/LuoLUZ17}
of each block, when computed {\em after} the MSA skip connections, is much smaller than the 
effective receptive field computed {\em before} the MSA skip connections.
Both empirical observations lead to the conclusion that the MSA skip connections in the last blocks may be detrimental for the 
{\em representation} capacity of the final patch token embeddings. 
This problem is emphasized when using our spatial entropy loss, 
since it is computed using the attention maps of the 
  last-block MSA (\cref{sec:Spatial-entropy}).
  For these reasons,  we propose to remove the MSA skip connections in the last block ($L$).
Specifically, in the $L$-th block, we replace Eq.~\ref{eq:transformer-standard-1}-\ref{eq:transformer-standard-2} with:%
\begin{align}%
    \pmb{z}' = \mbox{MSA}(\mbox{LN}(\pmb{z}^{L-1})) , \label{eq:skip-connection-removal-1}\\
    \pmb{z}^L = \mbox{MLP}(\pmb{z}')  + \pmb{z}' . \label{eq:skip-connection-removal-2}%
\end{align}
\noindent
Note that, in addition to removing the MSA skip connections (Eq.~\ref{eq:skip-connection-removal-1}), we also remove the subsequent LN (Eq.~\ref{eq:skip-connection-removal-2}), 
because we empirically observed that this further improves the VT accuracy (see \cref{sec:Ablation}).

\subsection{Discussion}
\label{sec:Discussion}

In this section, we discuss and motivate  the choices made in \cref{sec:Spatial-entropy} and \cref{sec:skip-connections}.
First, we use  $S^h$, extracted before the $\mbox{softmax}$ (Eq.~\ref{eq:attention-standard}) because, using the $\mbox{softmax}$,
the network can ``cheat'', by increasing the norm of the vectors 
$\pmb{q}_{CLS}$ and $\pmb{k}_{x,y}$ ($(x,y) \in \{1,...,k\}^2$). As a result, the dot product $<\pmb{q}_{CLS}, \pmb{k}_{x,y}>$ also largely increases, and the 
$\mbox{softmax}$ operation (based on the exponential function)
enormously exaggerates the difference between the elements in $S^h$,
generating a very peaked distribution, which zeros-out non-maxima $(x,y)$ elements. 
We 
observed that, when using the $\mbox{softmax}$, the VT is able to minimize Eq.~\ref{eq:spatial-entropy-loss} by producing single-peak similarity maps which have a 0 entropy, each being composed of only one connected component with only one single token (i.e., $h_r = 1$ and $n_j = 1$). 

Second, the spatial entropy (Eq.~\ref{eq:spatial-entropy}) is
computed for each head separately and then averaged (Eq.~\ref{eq:spatial-entropy-loss}) to
allow each head to focus on {\em different} image regions. 
Note that, although computing the connected components (Eq.~\ref{eq:connected-components}) is a non-differentiable operation, $C^h$ is only used to ``pool'' the values of $B^h$ (Eq.~\ref{eq:single-component-probability}), and each $C_j$ can be implemented as a binary mask (more details in the  Appendix, where we also compare $\mathcal{L}_{se}$ with other solutions). 
It is also important to note that, although a smaller number of connected components ($h_r$) can decrease $\mathcal{L}_{se}$,
this does not force the VT to always produce a {\em single} connected component (i.e., $h_r = 1$) because of the contribution of the main task loss (e.g., $\mathcal{L}_{ce}$). For instance, \cref{fig:teaser} (c) shows 4 big connected components which correctly correspond to the non-occluded parts of the bear and their reflections into the river, respectively.

Finally, we
remove the MSA skip connections only in the last block 
(Eq.~\ref{eq:skip-connection-removal-1}-\ref{eq:skip-connection-removal-2})
because, according to the  results reported in \citep{do-VTs-see-like-CNNs},  removing the skip connections in the ViT intermediate blocks brings to an accuracy drop.  
In contrast, in \cref{sec:Ablation} we  show that our strategy, which keeps the ViT architecture unchanged apart from the last block, is beneficial even when used {\em without} our spatial entropy loss.
Similarly, in preliminary experiments in which we used the spatial entropy loss also in other intermediate layers ($l < L$), we did not observe  any significant improvement.
In the rest of this paper, we refer to our full method
SAR as composed of the spatial entropy loss (\cref{sec:Spatial-entropy})
and the last-block MSA skip connection and LN removal (\cref{sec:skip-connections}).

\section{Experiments}
\label{sec:Experiments}

In \cref{sec:Ablation} we 
 analyse the contribution of the spatial entropy loss and the skip connection removal.
In \cref{sec:Main-results} we show that   SAR  improves  ViT   in different training-testing scenarios and with different downstream tasks.
In \cref{sec:Attention-map-analysis} 
we analyse the properties of the attention maps generated using SAR. In the Appendix,  
 we provide additional experiments  and we use SAR jointly with fully self-supervised learning approaches. 
Each model was trained using 8  NVIDIA V100 32GB GPUs.

\subsection{Ablation study}
\label{sec:Ablation}

In this section, we  analyse the influence of 
the $\lambda$ value (\cref{sec:Spatial-entropy}), the removal of
  the skip connections and the LN in the last ViT block (\cref{sec:skip-connections}), and the use of the spatial entropy loss (\cref{sec:Spatial-entropy}).
In all the ablation  experiments,  we use ImageNet-100 (IN-100) \citep{tian2019contrastive, wang2020understanding}, which is a subset of 100 classes of ImageNet, and ViT-S/16, a 22 million parameter ViT \citep{ViT} trained with $224 \times 224$ resolution images and 
$14 \times 14$ patches tokens ($k = 14$) with a patch resolution of $16 \times 16$ \citep{DeiT}.
Moreover, in all the experiments in this section,  we adopt the training protocol and the data-augmentations described in 
\citep{Swin}. 
Note that these data-augmentations include, among other things, the use of Mixup \citep{mixup} and CutMix \citep{CutMix}
(which are also used in  all the supervised classification experiments of \cref{sec:Main-results}), and this shows that our entropy loss can be used jointly with ``image-mixing'' techniques.

In \cref{tab:ablation} (a), 
 we train from scratch all the models 
using 100 epochs and we show the impact on the test set accuracy using different values of $\lambda$.
In  the experiments of this table, we use
our loss function ($\mathcal{L}_{tot} = \mathcal{L}_{ce} + \lambda \mathcal{L}_{se}$)
and we remove both the skip connections and the 
LN in the last block (Eq.~\ref{eq:skip-connection-removal-1}-\ref{eq:skip-connection-removal-2}),
thus the column $\lambda = 0$ corresponds to the result reported in \cref{tab:ablation} (b),  Row ``C"
(see below).
In the rest of the paper, we use the results obtained with this setting (IN-100, 100 epochs,  etc.) and the best $\lambda$ value ($\lambda = 0.01$) for {\bf all} the other datasets,
 training scenarios (e.g., training from scratch, fine-tuning, fully self-supervised learning) and VT architectures (e.g., ViT, CvT, PVT, etc.).
 In fact, although a higher accuracy can very likely be obtained by tuning  $\lambda$, our goal is to show that SAR is an easy-to-use regularization approach, even 
without tuning its only hyperparameter.

\begin{table*}[t]
\setcounter{table}{0}
\caption{IN-100. (a) Influence of the spatial entropy loss weight $\lambda$ (100 training epochs). (b) Analysis of the different components of SAR (100 epochs). (c) Influence of the number of epochs.}
\setlength{\tabcolsep}{1.0pt}
\small
    \begin{minipage}[t]{.15\textwidth}
    \captionsetup{labelformat=empty} 
    \centering
        \begin{tabular}{lc}
        \toprule
        \boldmath{$\lambda$} & \textbf{Top-1 Acc.} \\
         \midrule
         0 & 75.72 \\
         0.001 & 75.82 \\
         0.005 & 76.16 \\
         0.01 & \textbf{76.72} \\
         0.05 & 76.22 \\
         0.1 & 75.88 \\
        \bottomrule
        \end{tabular}
        \caption{(a)}
    \end{minipage}
    \hfill
    \begin{minipage}[t]{.45\textwidth}
    \captionsetup{labelformat=empty} 
        \centering
        \begin{tabularx}{\columnwidth}{Xl}
        \toprule
        \textbf{Model} & \textbf{Top-1 Acc.}  \\
        \midrule
        A: Baseline  & 74.22 \\
        B: A + no MSA skip connections & 74.64   \posimprov{0.42} \\
        C: B +  no LN & 75.72    \posimprov{1.5} \\
        D: C +  no MLP skip connections & 73.76    \negimprov{0.46} \\
        E: A + spatial entropy loss & 74.78 \posimprov{0.56} \\
        F: A + SAR  & 76.72     \posimprov{2.5} \\
        \bottomrule
    \end{tabularx}
    \caption{(b)}
    \end{minipage}
    \hfill
    \begin{minipage}[t]{0.33\textwidth}
    \captionsetup{labelformat=empty}
    \centering
    \begin{tabularx}{\columnwidth}{Xll}
    \toprule
    \textbf{Model} & \multicolumn{2}{c}{\textbf{Top-1 Acc.}} \\
    \cmidrule{2-3} 
         & 100 ep. & 300 ep.\\
        \midrule
        ViT-S/16 & 74.22 & 80.82\\
        
        \rowcolor{Light}ViT-S/16+SAR & \makecell{\textbf{76.72} \\ \posimprov{2.5}} & \makecell{\textbf{85.24}\\  \posimprov{4.42}} \\
    \bottomrule
    \end{tabularx}
    \caption{(c)}
    \end{minipage}
\label{tab:ablation}
\end{table*}

In \cref{tab:ablation} (b), we  train from scratch all the models 
using 100 epochs and Row ``A'' corresponds to our run of the original ViT-S/16
(Eq.~\ref{eq:transformer-standard-1}-\ref{eq:transformer-standard-2}).
When we
remove the MSA skip connections (Row ``B''), we observe  a $+0.42$ points improvement, which becomes $+1.5$ if we also remove the LN (Row ``C'').
This experiment confirms that the last block patch tokens can learn more useful representations if we inhibit the MSA identity path
(Eq.~\ref{eq:skip-connection-removal-1}-\ref{eq:skip-connection-removal-2}).
However, if we also remove the skip connections in the subsequent MLP layer (Row ``D''), the results are inferior to the baseline.
Finally, when we use the spatial entropy loss with the original architecture 
(Row ``E''), the improvement is marginal,  but using 
$\mathcal{L}_{se}$ jointly with Eq.~\ref{eq:skip-connection-removal-1}-\ref{eq:skip-connection-removal-2} (full model, Row ``F''),
the  accuracy boost with respect to the baseline is much stronger. \cref{tab:ablation} (c) compares training with 100 and  300 epochs and shows that, in the latter case, SAR can reach a much higher relative improvement with respect to the baseline ($+4.42$).

\subsection{Main results}
\label{sec:Main-results}

\noindent{\bf Sample efficiency.} 
In order to show that SAR can alleviate the need of large labeled datasets (\cref{sec:intro}),
 we  follow a recent trend of works \citep{OurNIPS,DBLP:journals/corr/abs-2112-10740,DBLP:journals/corr/abs-2103-13559}
where VTs are trained from scratch on small-medium datasets (without pre-training on ImageNet). Specifically, we 
  strictly follow the training protocol proposed by \cite{DBLP:journals/corr/abs-2112-10740}, where 5,000 epochs are used to train  ViT-S/16 directly on 
each target dataset. 
The results are shown in \cref{tab:medium_size_dataset-new}, 
which also provides the number of training and testing samples of each dataset, jointly with
 the accuracy values of the baseline (ViT-S/16, trained in a standard way,  without SAR), both reported from \cite{DBLP:journals/corr/abs-2112-10740}. \cref{tab:medium_size_dataset-new} shows that SAR can {\em drastically} improve the ViT-S/16 accuracy on these small-medium datasets, with an  improvement ranging from $+18.17$ to $+30.78$ points.
These results, jointly with the results obtained on IN-100 (\cref{tab:ablation} (c)), show that SAR is particularly effective in boosting the performance of ViT when labeled training data are scarce.

We further analyze the impact of the amount of training data  using different subsets of IN-100 with different sampling ratios (ranging from $25\%$ to $75\%$, with images randomly selected). We use the same training protocol of \cref{tab:ablation}  (c) (e.g., 100 training epochs, etc.) and we test on the {\em whole} IN-100 validation set. \cref{tab:sampling_ratio_in100-new}  shows the results, confirming  that, with less data, the accuracy boost obtained using SAR can significantly increase (e.g., with $75\%$ of the data we have a 10.5 points improvement).

\setcounter{table}{1}

\begin{table*}[t]
\small
\caption{Training from scratch on small-medium datasets. The baseline (ViT-S/16) results are reported from \citep{DBLP:journals/corr/abs-2112-10740}.}
\label{tab:medium_size_dataset-new}
    \centering
    \begin{tabular}{lccccc}
        \toprule
        \textbf{Dataset} & {$\#$ Train samples} & {$\#$ Test samples} & {$\#$ Classes} &  \textbf{ViT-S/16} & \textbf{ViT-S/16+SAR} \\
        \midrule
        Stanford Cars & 8,144 & 8,041 & 196 & 35.3 & \cellcolor{Light}\textbf{64.65} \posimprov{29.35} \\
        Clipart & 34,019 & 14,818 & 345 & 41.0 & \cellcolor{Light}\textbf{64.95} \posimprov{23.95} \\
        Painting & 52,867 & 22,892 & 345 & 38.4 & \cellcolor{Light}\textbf{57.11} \posimprov{18.17} \\
        Sketch & 49,115 & 21,271 & 345 & 37.2 & \cellcolor{Light}\textbf{62.98} \posimprov{30.78}\\
        \bottomrule
    \end{tabular}
\end{table*}

\begin{table}[t]
\small
\caption{IN-100 experiments with different sampling ratios (100 epochs).}
\label{tab:sampling_ratio_in100-new}
\setlength{\tabcolsep}{3.5pt}
\centering
    \begin{tabularx}{\columnwidth}{Xllll}
    \toprule
    \textbf{Model} & \multicolumn{4}{c}{\textbf{Top-1 Acc.}} \\
    \cmidrule{2-5} 
         & 0.25 & 0.50 & 0.75 & 1.00\\
    \midrule
    ViT-S/16 & 21.66 & 29.86 & 35.62 & 74.22 \\
    \rowcolor{Light}ViT-S/16+SAR &\makecell{\textbf{29.06}\\ \posimprov{7.4}} & \makecell{\textbf{38.02}\\ \posimprov{8.16}} & \makecell{\textbf{46.12}\\ \posimprov{10.5}} & \makecell{\textbf{76.72}\\ \posimprov{2.5}}\\ 
    \bottomrule
    \end{tabularx}
\end{table}

\begin{figure}[ht]
    \centering
    \includegraphics[width=0.8\columnwidth]{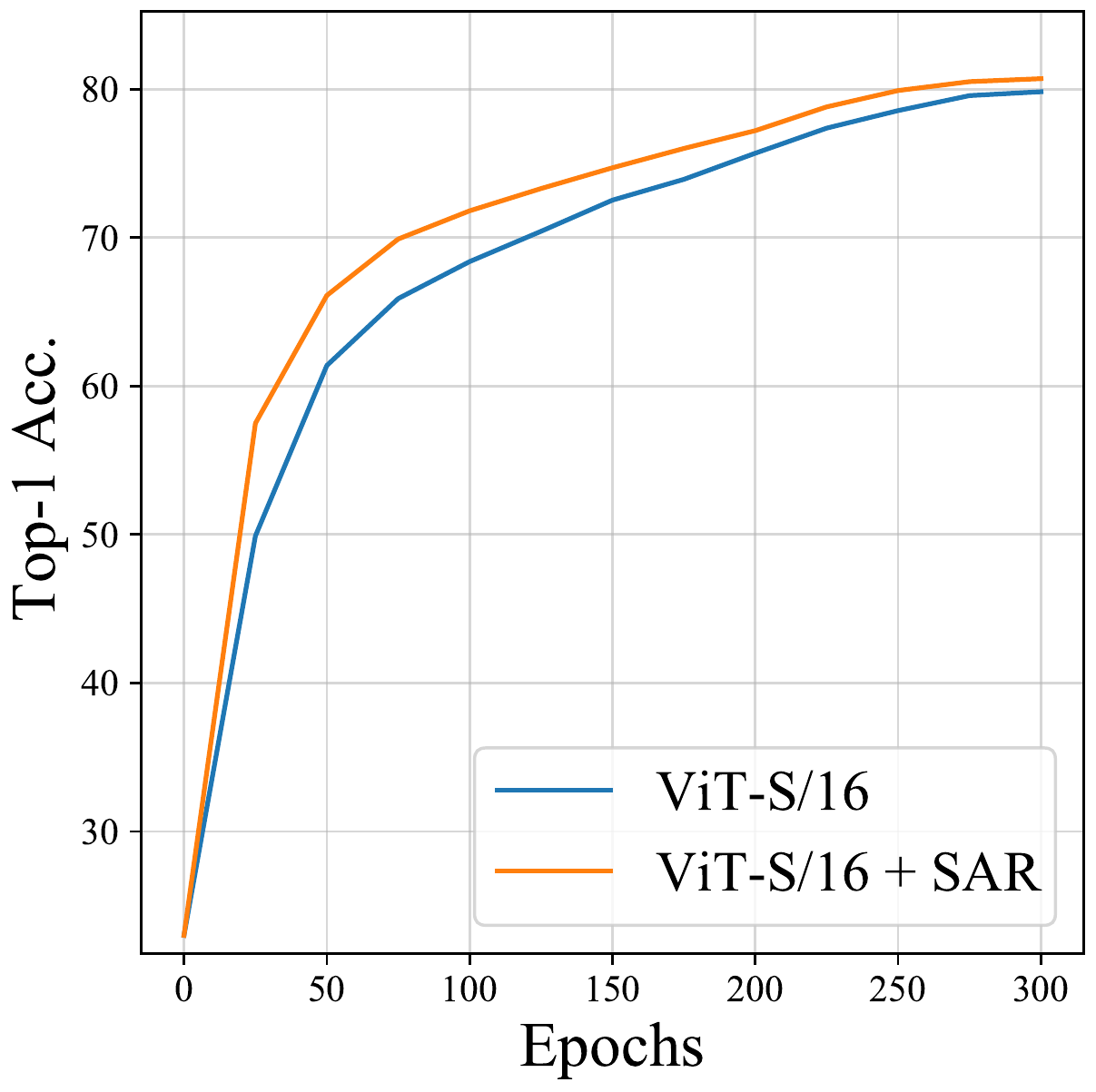}
    \caption{IN-1K,  validation set accuracy with respect to the number of training epochs.}
    \label{fig:curves}
\end{figure}

\noindent
{\bf Training  on ImageNet-1K.}
We extend the previous results training ViT 
 on ImageNet-1K (IN-1K), 
 and comparing SAR with the baseline 
 (ViT-S/16, trained in a standard way,  without SAR), and with 
the Dense Relative Localization (DRLoc) loss \citep{OurNIPS}, which, similarly to SAR, is based on an auxiliary self-supervised task used to regularize VTs training (\cref{sec:RelatedWork}).
DRLoc  encourages the VT to learn spatial relations within an image  by  predicting the relative distance between the $(x,y)$ positions of randomly sampled output embeddings from the $k \times k$ grid of the last layer $L$. 
\cref{tab:ourneuipsvsthis} shows that SAR can boost the accuracy of ViT of almost 1 point
{\em without any additional learnable parameters} or drastic architectural changes, 
and this gain is higher than DRLoc.
The reason why the relative improvement is smaller with respect to what was obtained with smaller datasets 
is likely due to the fact that, usually, regularization techniques are mostly effective with small(er) datasets 
\citep{OurNIPS,https://doi.org/10.48550/arxiv.2204.03632}.
Nevertheless,  Fig.~\ref{fig:curves} shows that  SAR can be used jointly with large datasets to significantly speed-up training. For instance, ViT-S/16 + SAR, with 100 epochs, achieves almost the same accuracy as the baseline trained with 150 epochs, while we surpass the final baseline accuracy  ($79.8 \%$ at epoch 300) with only 250 training epochs ($79.9 \%$ at epoch 250). 
Finally,  the two regularization approaches 
 (SAR and DRLoc) can potentially be combined, but we leave this for future work.

\begin{table}[t]
\small
    \caption{IN-1K experiments (300 training epochs).
    \textsuperscript{\dag}~This result has been obtained by us using the publicly available code taken from \citep{OurNIPS}.
    }
    \label{tab:ourneuipsvsthis}
    \centering
    \begin{tabularx}{\columnwidth}{Xl}
        \toprule
        \textbf{Method} & \textbf{Top-1 Acc.}\\ 
        \midrule
         ViT-S/16~\citep{DeiT}  & 79.8 \\
         ViT-S/16 + DRLoc\textsuperscript{\dag}  & 80.2 \posimprov{0.4} \\
        \rowcolor{Light}ViT-S/16 + SAR & \textbf{80.7}   \posimprov{0.9} \\
        \bottomrule
    \end{tabularx}
\end{table}

\noindent
{\bf Transfer learning with object detection and image segmentation tasks.}
We further analyze the quality of the models pre-trained on IN-1K using object detection and semantic segmentation downstream tasks. Specifically, we use ViTDet \citep{ViTDet}, a recently proposed object detection/segmentation framework in which a (standard) pre-trained ViT backbone is adapted {\em only at fine-tuning time} in order to generate a feature pyramid to be used for multi-scale object detection or image segmentation. Note that, as mentioned in \cref{sec:intro}, hybrid  approaches which are based on ad hoc architectures 
are not suitable from this framework, because
they need to redesign their   backbone 
and introduce a feature pyramid also in the
 pre-training stage \citep{ViTDet,PVT}.
Conversely, we use the pre-trained networks whose results are reported in \cref{tab:ourneuipsvsthis}, where the baseline is ViT-S/16 and our approach corresponds to ViT-S/16 + SAR. 
For the object detection task,
following  \citep{DBLP:conf/iccv/Girshick15}, we use the
trainval set of PASCAL VOC 2007 and 2012 \citep{Pascal-voc-dataset} (16.5K training images) to fine-tune  the two models using ViTDet, and the test set of PASCAL VOC 2007 for evaluation.
The results,  reported in \cref{tab:obj-det-with-ViTDet},  show that the model pre-trained using SAR outperforms the baseline of more than 2 points, which is an increment even larger that the boost obtained in the classification task used during pre-training (\cref{tab:ourneuipsvsthis}). 
Similarly, for the segmentation task, we use PASCAL VOC-12 trainval for fine-tuning and  PASCAL VOC-12 test for evaluation.
\cref{tab:segmentation-with-ViTDet} shows that the model pre-trained with  
SAR achieves an improvement of more than 2.5 mIoU points compared to the baseline.
These detection and segmentation improvements confirm that the local inductive bias introduced in ViT using SAR can be very useful for localization tasks, especially when the fine-tuning data are scarce like in PASCAL VOC.

\begin{table}[t]
\small
    \caption{Object detection on PASCAL VOC 2007, evaluated using mean Average Precision (mAP).}
    \label{tab:obj-det-with-ViTDet}
    \centering
    \begin{tabularx}{\columnwidth}{Xll}
        \toprule
        \textbf{Model} & \textbf{mAP}  \\ 
        \midrule
        ViT-S/16 & 77.1  \\
        \rowcolor{Light} ViT-S/16 + SAR & \textbf{79.2} \posimprov{2.1}  \\
        \bottomrule
    \end{tabularx}
\end{table}

\begin{table}[t]
\small
    \caption{Semantic segmentation on PASCAL VOC 2012, evaluated using mean Intersection over Union (mIoU).}
    \label{tab:segmentation-with-ViTDet}
    \centering
    \begin{tabularx}{\columnwidth}{Xll}
        \toprule
        \textbf{Model}  & \textbf{mIoU} \\ 
        \midrule
        ViT-S/16 &  66.17 \\
        \rowcolor{Light} ViT-S/16 + SAR  & \textbf{68.68} \posimprov{2.51} \\
        \bottomrule
    \end{tabularx}
\end{table}

\noindent 
{\bf Transfer learning with different fine-tuning protocols.}  In this battery of experiments,  we evaluate SAR in a transfer learning scenario with classification tasks.
We adopt the four datasets used in \citep{ViT,DeiT,MoCo-v3,DINO}:
CIFAR-10 and CIFAR-100~\citep{cifar}, Oxford Flowers102~\citep{flower}, and 
Oxford-IIIT-Pets \citep{Pets-dataset}.
The standard
transfer learning protocol  consists in  pre-training on 
IN-1K, and then  fine-tuning on each  dataset. 
This corresponds to the first row 
in  \cref{tab:fine-tuning-new},
where the IN-1K pre-trained model is ViT-S/16 in \cref{tab:ourneuipsvsthis}.
The next three rows show different pre-training/fine-tuning configurations, in which we use SAR 
in one of the two phases or in both
(see the  Appendix  for more details).
All the  configurations lead to an overall  improvement of the accuracy with respect to the baseline, and show that SAR can be used flexibly. For instance, SAR can be used when fine-tuning a VT trained in a standard way, without the need to re-train it on ImageNet.

\begin{table*}[t]
\small
    \caption{Transfer learning results (100 epochs fine-tuning). 
    The first row corresponds to a standard fine-tuning protocol, while  the other configurations include SAR either in the pre-training or in the fine-tuning stage.}
    \label{tab:fine-tuning-new}
    \centering
    \begin{tabular}{ccllll}
    \toprule
    \textbf{SAR} & \textbf{SAR} & \textbf{CIFAR-10} & \textbf{CIFAR-100} & \textbf{Flowers} & \textbf{Pets} \\
    \textbf{pre-training} & \textbf{fine-tuning} &  &  &  &  \\
    \midrule
    \xmark & \xmark & 98.59 & 88.95 & 95.07 & 92.21 \\
    \rowcolor{Light}     \cmark   & \xmark & 98.69 \posimprov{0.1} & 89.19 \posimprov{0.24} & 96.05 \posimprov{0.98} & 92.7 \posimprov{0.49} \\
    \rowcolor{Light} \xmark & \cmark & 98.72 \posimprov{0.13} & 88.95 & 95.1 \posimprov{0.03}  & 92.34 \posimprov{0.13} \\
    \rowcolor{Light} \cmark & \cmark & 98.65 \posimprov{0.06} & 89.21 \posimprov{0.26} & 96.1 \posimprov{1.03} & 92.7 \posimprov{0.49} \\
    \bottomrule
    \end{tabular}
\end{table*}    

\noindent{\bf Out-of-distribution testing.} We  test the robustness of our ViT trained with SAR when the testing distribution is different from the training distribution. Specifically, following \cite{DBLP:conf/nips/BaiMYX21}, we use two different testing sets: (1) ImageNet-A \citep{DBLP:journals/corr/abs-1907-07174}, which are real-world images but 
collected from challenging  scenarios (e.g., occlusions, fog scenes, etc.), and (2) ImageNet-C \citep{DBLP:journals/corr/abs-1807-01697},
which is designed to measure the model robustness against  common image corruptions. 
Note that training is done only on IN-1K. Thus, in \cref{tab:robustness_results},
ViT-S/16 and  ViT-S/16 + SAR  correspond to the models we trained on IN-1K whose results on the IN-1K standard validation set are reported in \cref{tab:ourneuipsvsthis}. ImageNet-A and ImageNet-C are used {\em only for testing}, hence they are useful to assess the behaviour of a model when evaluated on a distribution different from the training distribution \citep{DBLP:conf/nips/BaiMYX21}. 
The results reported in \cref{tab:robustness_results} show that SAR can significantly  improve the robustness of ViT (note that, with the mCE metric, the lower the better \citep{DBLP:conf/nips/BaiMYX21}). We presume that this is a side-effect of our spatial entropy loss minimization, which leads to heads usually focusing on the foreground objects and, therefore, reducing the dependence with respect to  the background appearance variability distribution.

\begin{table}[t]
\setlength{\tabcolsep}{2.0pt}
\small
    \caption{Out-of-distribution testing  on ImageNet-A (IN-A) and ImageNet-C (IN-C).}
    \label{tab:robustness_results}
    \centering
    \begin{tabularx}{\linewidth}{Xll}
    \toprule
    \textbf{Model} &  \textbf{IN-A} (Acc. $\uparrow$)  & \textbf{IN-C} (mCE $\downarrow$)\\ 
    \midrule
    ViT-S/16  & 19.2  & 52.8  \\
    \rowcolor{Light}ViT-S/16 + SAR & \textbf{22.39} \posimprov{3.19} & \textbf{51.6} \negimprovgreen{1.2} \\
    \bottomrule 
    \end{tabularx}
\end{table}

\noindent
{\bf Different VT architectures.}
Finally, we show that SAR can be  used with VTs of different capacities and with architectures different from ViT. For this purpose,
 we plug SAR into the following VT architectures: ViT-S/16~\citep{DeiT}, T2T \citep{T2T}, PVT \citep{PVT} and CvT \citep{CvT}. 
 Specifically, T2T, PVT  and CvT are hybrid architectures, which use typical CNN mechanisms to introduce a local inductive bias into the VT training (\cref{sec:intro,sec:RelatedWork}).
  We omit other common frameworks such as, for instance, Swin \citep{Swin} because of the lack of a \texttt{[CLS]} token in their architecture. Although the \texttt{[CLS]} token used, e.g, in \cref{sec:Spatial-entropy} to compute $S^h$, can potentially be replaced by a vector obtained by average-pooling all the patch embeddings, we leave this for future investigations. 
Moreover, for computational reasons, we focus on small-medium capacity VTs (see \cref{tab:ImageNet-1K-new} for details on the number of parameters of each VT). Importantly, for each tested method, we use the original  training protocol developed by the corresponding authors, including, e.g., the learning rate schedule, the batch size, the VT-specific hyperparameter values and the data-augmentation type
used to obtain the corresponding published results, 
both when we train the baseline and when we train using SAR.
Moreover, as usual (\cref{sec:Ablation}), we keep fixed the only SAR hyperparameter ($\lambda = 0.01$). Although better results can likely be obtained by 
adopting the common practice of
hyperparameter tuning (including the VT-specific hyperparameters), 
 our goal is to show that SAR can be easily used in different VTs, increasing their final testing accuracy.
  The results  reported in \cref{tab:imagenet-100} and \cref{tab:ImageNet-1K-new} show that SAR improves {\em all} the tested VTs, independently of their specific architecture, model capacity or training protocol. 
  Note that both PVT and CvT have a final  grid resolution of $7 \times 7$, which is smaller than the $14 \times 14$ grid used in ViT and T2T, and this probably has a negative impact on our spatial based entropy loss. 

Overall, the results reported in   \cref{tab:imagenet-100} and \cref{tab:ImageNet-1K-new}: (1) Confirm that SAR is mostly useful with smaller datasets (being the relative improvements on IN-100 significantly larger than those obtained on IN-1K). (2) Show that the object-based inductive bias introduced when training with SAR is (partially) complementary with respect to the local bias embedded in the hybrid VT architectures, as witnessed by the positive boost obtained when these VTs are used jointly with SAR. (3) Show that, on IN-1K, the accuracy of ViT-S/16 + SAR is comparable with the hybrid VTs (without SAR). However, the advantage of ViT-S/16 + SAR is its simplicity, which does not need drastic architectural changes to the original ViT architecture, where the latter is quickly becoming a \emph{de facto} standard in many vision and vision-language tasks (\cref{sec:RelatedWork}).
  
\begin{table}[h]
  \centering
  \caption{IN-100 experiments with different VTs (100 training epochs).  The results of the baselines have been obtained by us using the corresponding publicly available code.}
  \small
\begin{tabularx}{\columnwidth}{X l }
    \toprule
     \textbf{Model} & \textbf{Top-1 Acc.}  \\
    \midrule
        ViT-S/16  & 74.22 \\
        \rowcolor{Light}ViT-S/16 + SAR & \textbf{76.72}   \posimprov{2.5} \\
        \midrule
        T2T-ViT-14 & 82.42   \\
        \rowcolor{Light}T2T-ViT-14 + SAR & \textbf{83.96} \posimprov{1.54} \\
        \midrule
        PVT-Small & 76.57 \\
        \rowcolor{Light}PVT-Small + SAR  & \textbf{77.78} \posimprov{1.21} \\
        \midrule
        CvT-13 & 83.38 \\
        \rowcolor{Light}CvT-13 + SAR  & \textbf{85.20} \posimprov{1.82} \\
        \bottomrule
    \end{tabularx}
  \label{tab:imagenet-100}
\end{table}

\begin{table}[t]
\small
\caption{
IN-1K experiments with different VTs (300 training epochs).  All results but ours are reported from the corresponding paper. The number of parameters are in millions.}
\label{tab:ImageNet-1K-new}
\centering
    \begin{tabularx}{\columnwidth}{lcl}
    \toprule
   \textbf{Module} & \textbf{Param.}  & \textbf{Acc.} \\
   \midrule
   ViT-S/16  & 22 & 79.8 \\
   \rowcolor{Light}ViT-S/16 + SAR    &  22  & \textbf{80.7} \posimprov{0.9} \\
   \midrule
   T2T-ViT-14  & 21.5  & 81.5 \\
   \rowcolor{Light}T2T-ViT-14 + SAR   & 21.5  & \textbf{81.9} \posimprov{0.4} \\
   \midrule
   PVT-Small  & 24.5 & 79.8 \\
   \rowcolor{Light}PVT-Small + SAR   & 24.5  & 79.84 \posimprov{0.04} \\
   \midrule
    CvT-13  & 20 & 81.6 \\
   \rowcolor{Light}CvT-13 + SAR   & 20  & \textbf{81.8} \posimprov{0.2}  \\
   \bottomrule
    \end{tabularx}
\end{table}


\begin{figure*}[!h]
    \vspace{-1mm}
    \centering
    \def\arraystretch{0.0}
    \setlength\tabcolsep{0.25pt}
    \footnotesize
    \renewcommand{\arraystretch}{0.0}
    \begin{tabular}{cccccccccccccc}
        \multicolumn{7}{c}{\scalebox{1}{ViT-S/16}} & \multicolumn{7}{c}{\scalebox{1}{ViT-S/16+SAR}} \vspace{0.2em}\\
    
         
           \includegraphics[trim=2 0 0 2,clip,width=0.15\columnwidth]{figures/attention_maps/deit/id06_img.png} &
          \includegraphics[trim=2 0 0 2,clip,width=0.15\columnwidth]{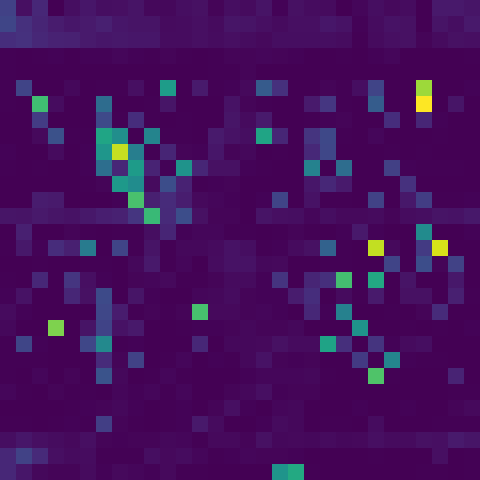} &
          \includegraphics[trim=2 0 0 2,clip,width=0.15\columnwidth]{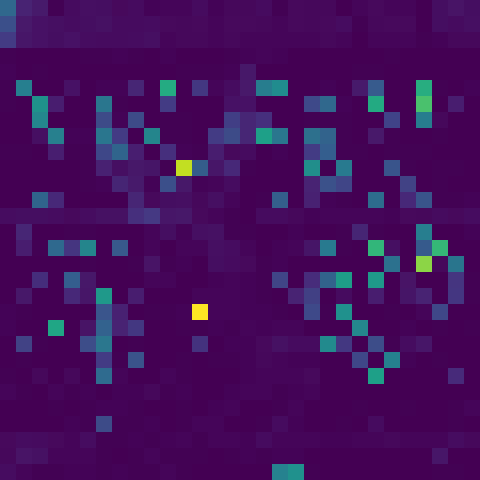} &
          \includegraphics[trim=2 0 0 2,clip,width=0.15\columnwidth]{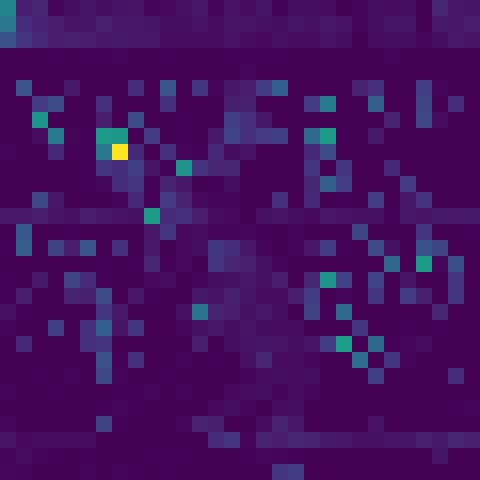} &
          \includegraphics[trim=2 0 0 2,clip,width=0.15\columnwidth]{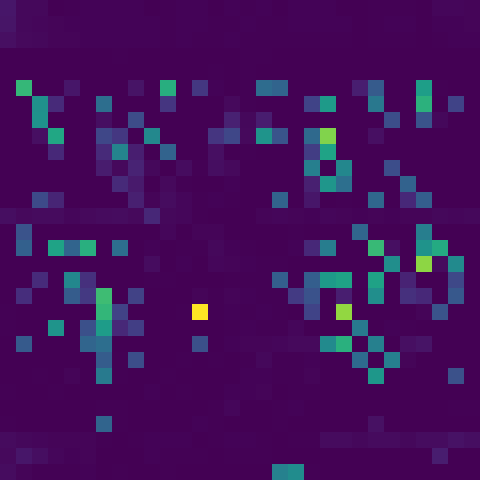} &
          \includegraphics[trim=2 0 0 2,clip,width=0.15\columnwidth]{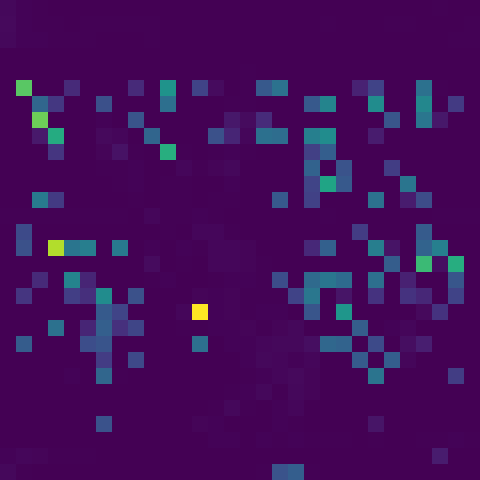} &
          \includegraphics[trim=2 0 0 2,clip,width=0.15\columnwidth]{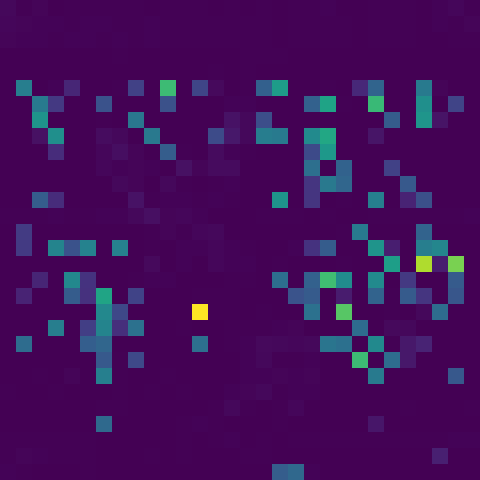} &
          \includegraphics[trim=2 0 0 2,clip,width=0.15\columnwidth]{figures/attention_maps/sar/id06_img.png} &
          \includegraphics[trim=2 0 0 2,clip,width=0.15\columnwidth]{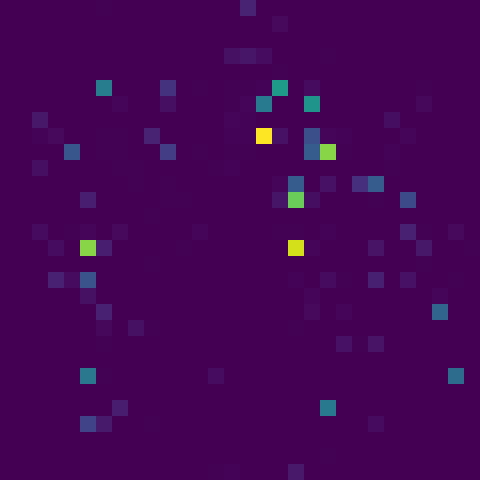} &
          \includegraphics[trim=2 0 0 2,clip,width=0.15\columnwidth]{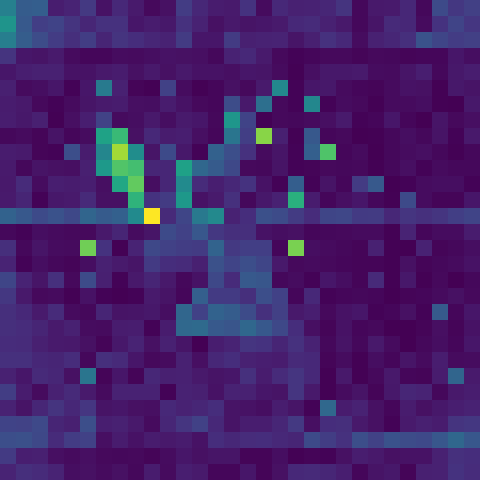} &
          \includegraphics[trim=2 0 0 2,clip,width=0.15\columnwidth]{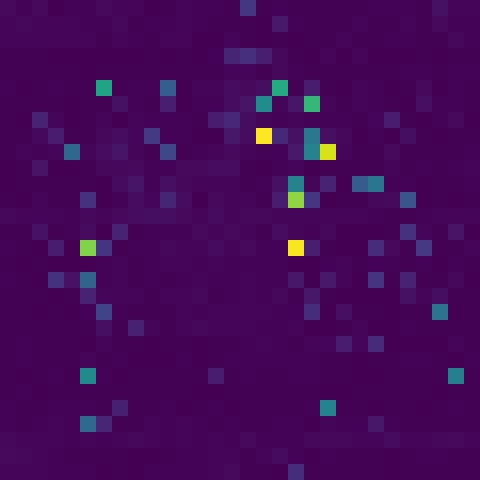} &
          \includegraphics[trim=2 0 0 2,clip,width=0.15\columnwidth]{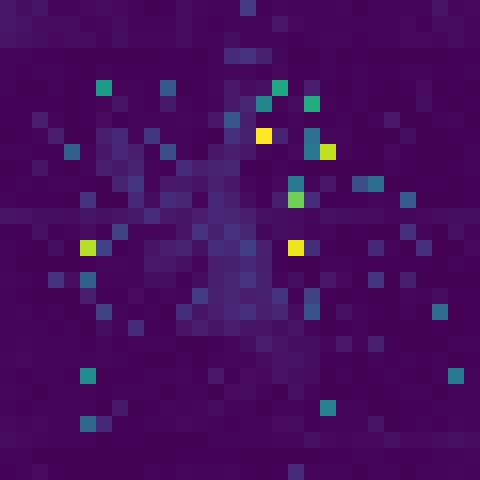} &
          \includegraphics[trim=2 0 0 2,clip,width=0.15\columnwidth]{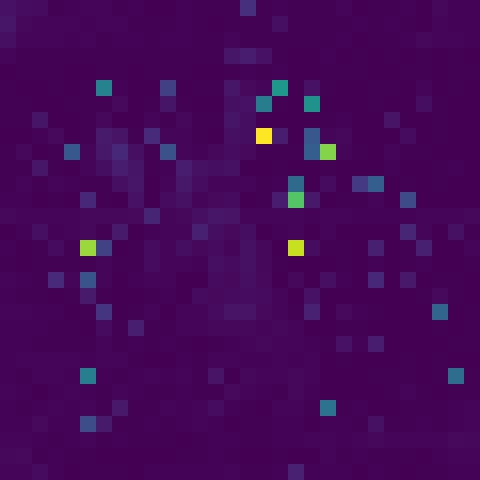} &
          \includegraphics[trim=2 0 0 2,clip,width=0.15\columnwidth]{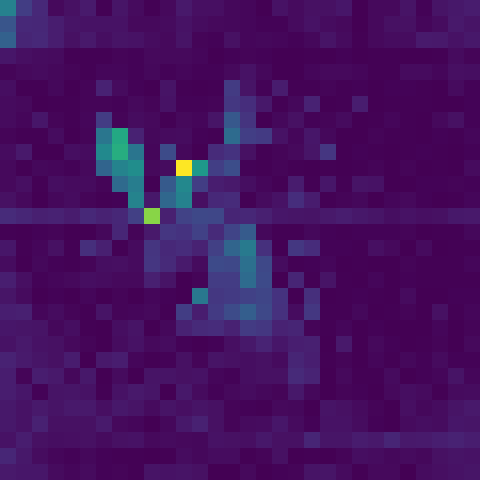} \\
         

         \includegraphics[trim=2 0 0 2,clip,width=0.15\columnwidth]{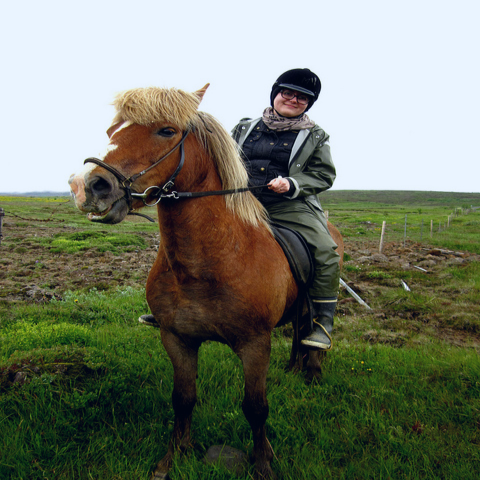} &
         \includegraphics[trim=2 0 0 2,clip,width=0.15\columnwidth]{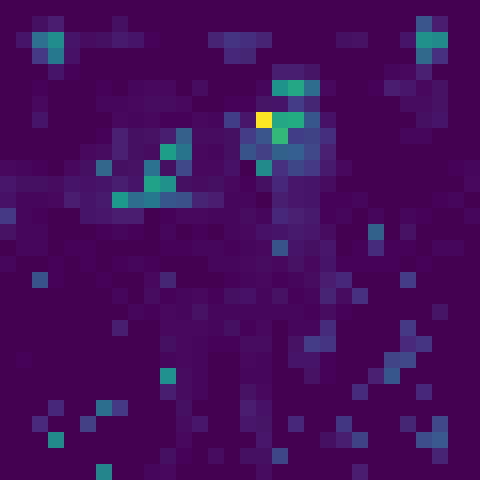} &
         \includegraphics[trim=2 0 0 2,clip,width=0.15\columnwidth]{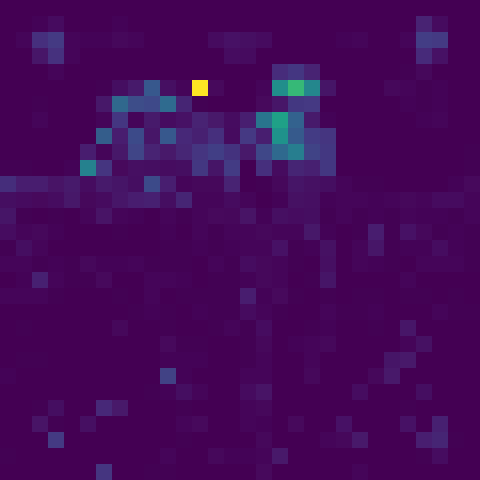} &
         \includegraphics[trim=2 0 0 2,clip,width=0.15\columnwidth]{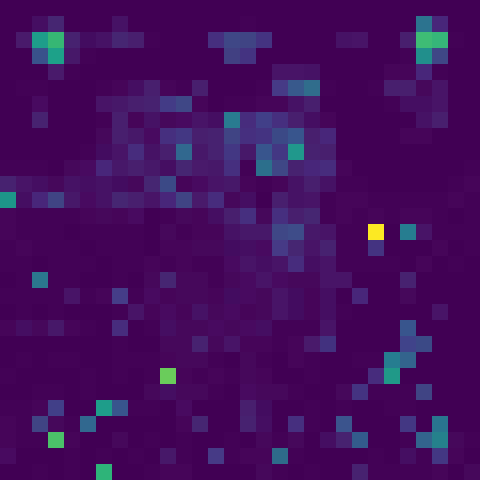} &
         \includegraphics[trim=2 0 0 2,clip,width=0.15\columnwidth]{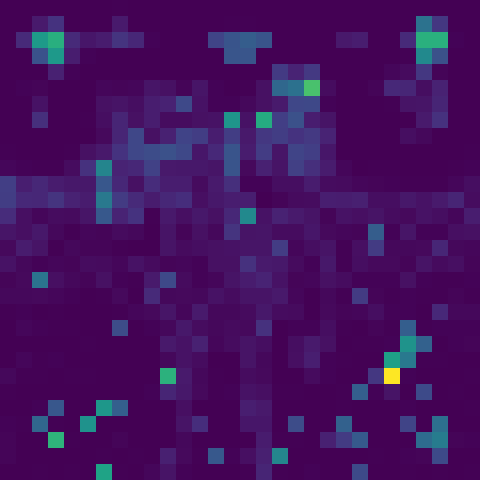} &
         \includegraphics[trim=2 0 0 2,clip,width=0.15\columnwidth]{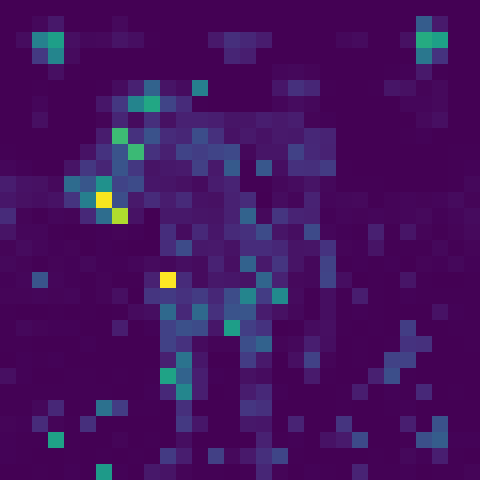} &
         \includegraphics[trim=2 0 0 2,clip,width=0.15\columnwidth]{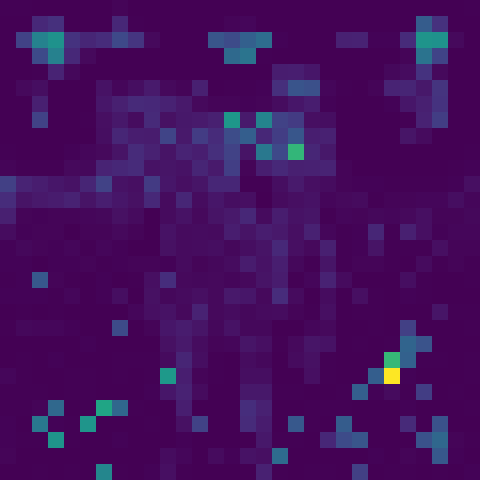} &
         \includegraphics[trim=2 0 0 2,clip,width=0.15\columnwidth]{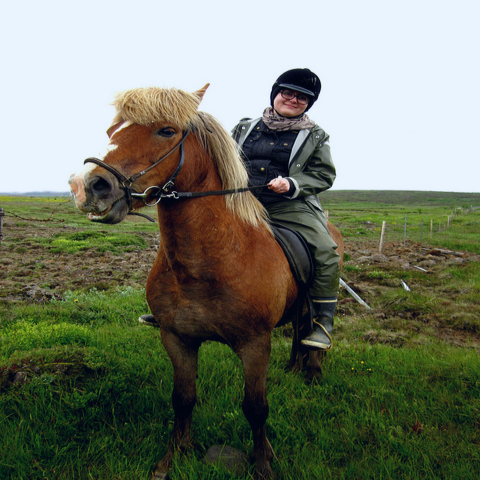} &
         \includegraphics[trim=2 0 0 2,clip,width=0.15\columnwidth]{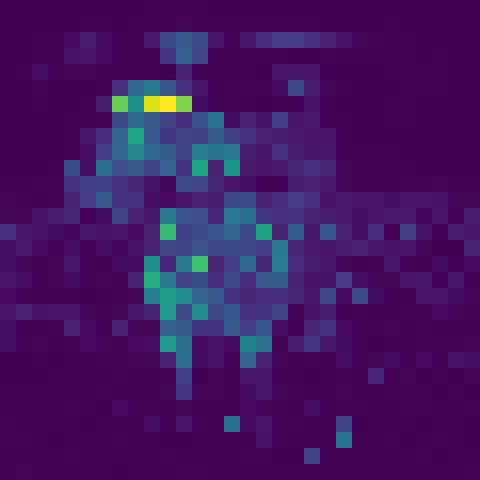} &
         \includegraphics[trim=2 0 0 2,clip,width=0.15\columnwidth]{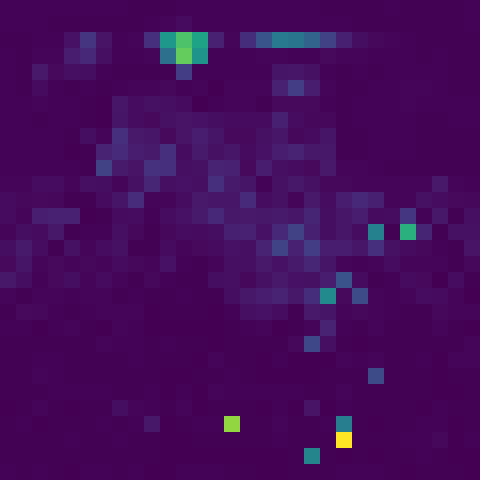} &
         \includegraphics[trim=2 0 0 2,clip,width=0.15\columnwidth]{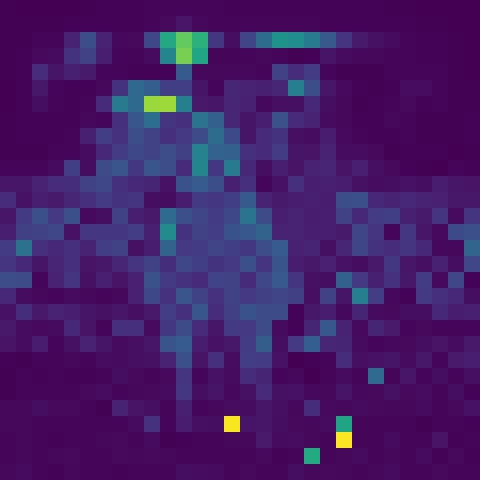} &
         \includegraphics[trim=2 0 0 2,clip,width=0.15\columnwidth]{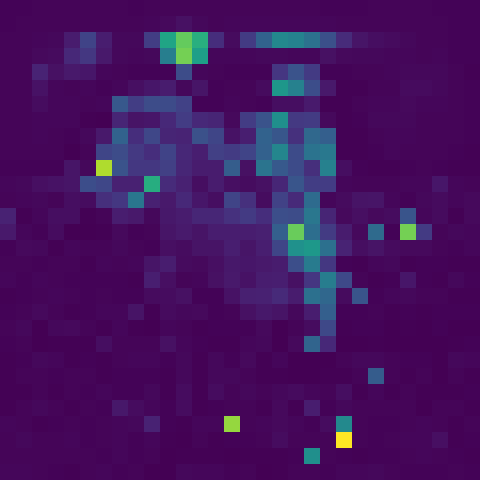} &
         \includegraphics[trim=2 0 0 2,clip,width=0.15\columnwidth]{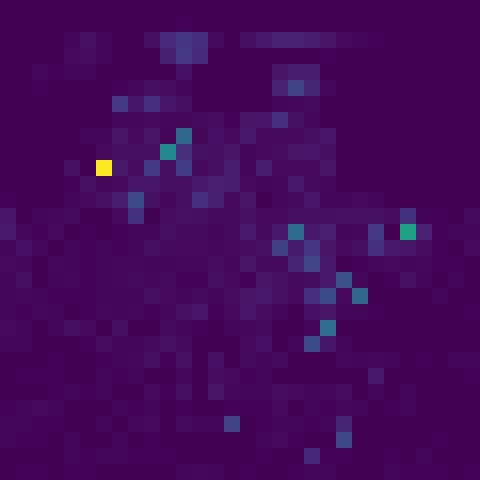} &
         \includegraphics[trim=2 0 0 2,clip,width=0.15\columnwidth]{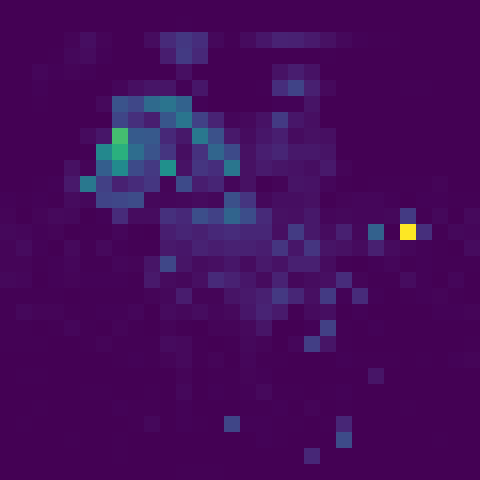} \\
         
         \includegraphics[trim=2 0 0 2,clip,width=0.15\columnwidth]{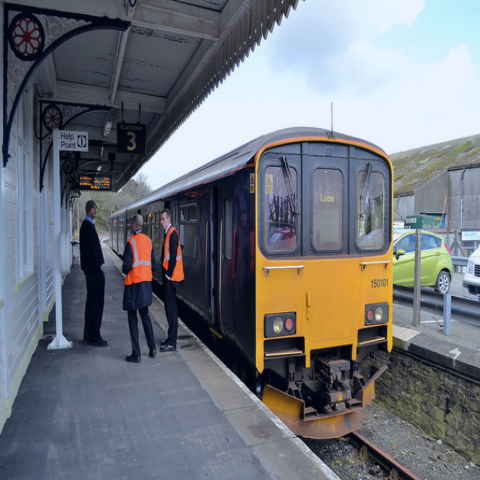} &
         \includegraphics[trim=2 0 0 2,clip,width=0.15\columnwidth]{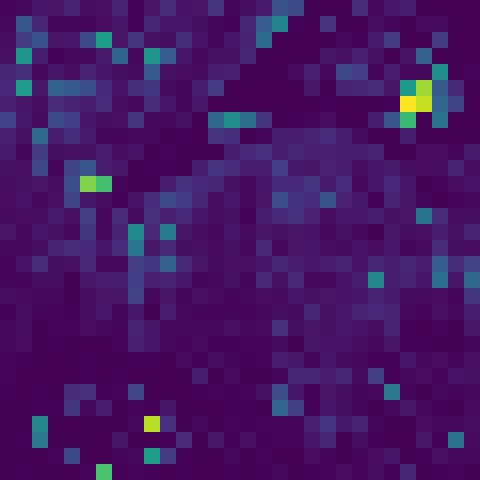} &
         \includegraphics[trim=2 0 0 2,clip,width=0.15\columnwidth]{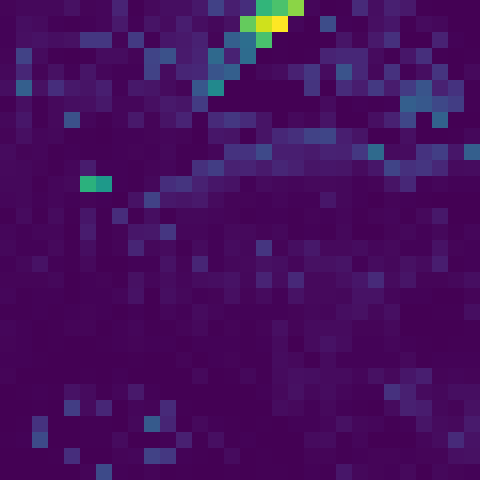} &
         \includegraphics[trim=2 0 0 2,clip,width=0.15\columnwidth]{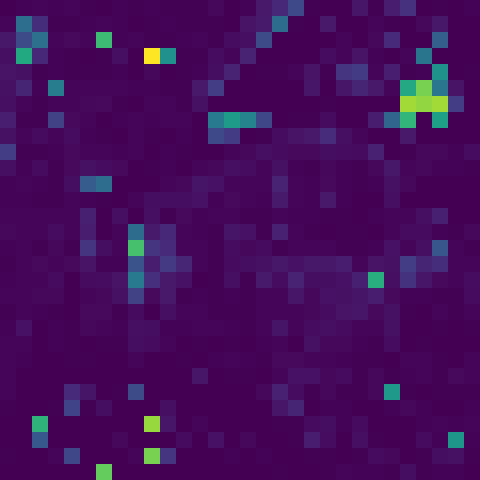} &
         \includegraphics[trim=2 0 0 2,clip,width=0.15\columnwidth]{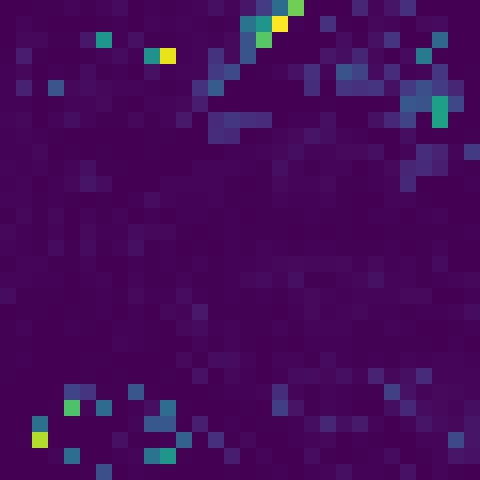} &
         \includegraphics[trim=2 0 0 2,clip,width=0.15\columnwidth]{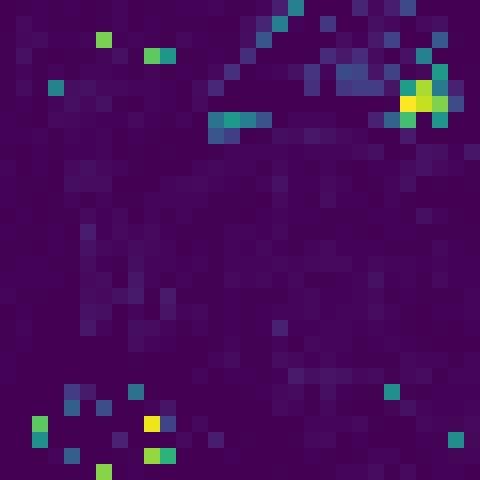} &
         \includegraphics[trim=2 0 0 2,clip,width=0.15\columnwidth]{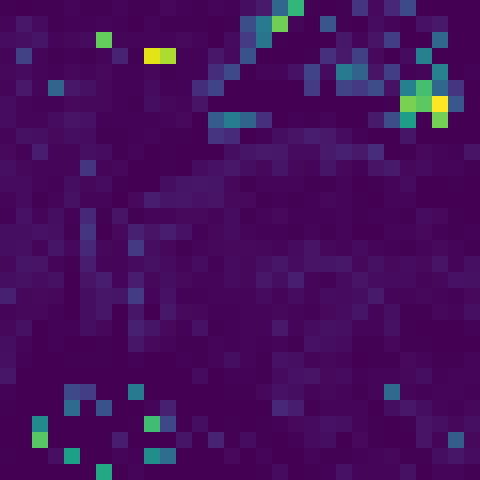} &
         \includegraphics[trim=2 0 0 2,clip,width=0.15\columnwidth]{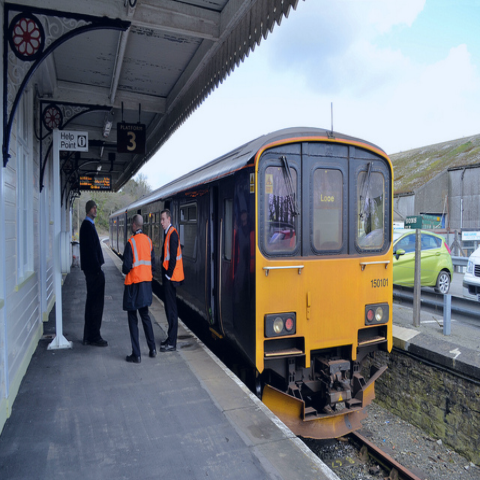} &
         \includegraphics[trim=2 0 0 2,clip,width=0.15\columnwidth]{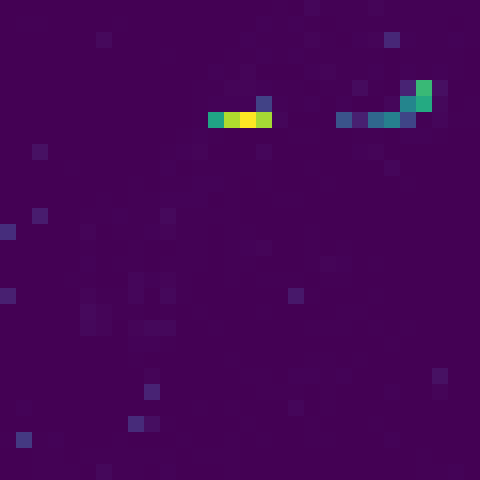} &
         \includegraphics[trim=2 0 0 2,clip,width=0.15\columnwidth]{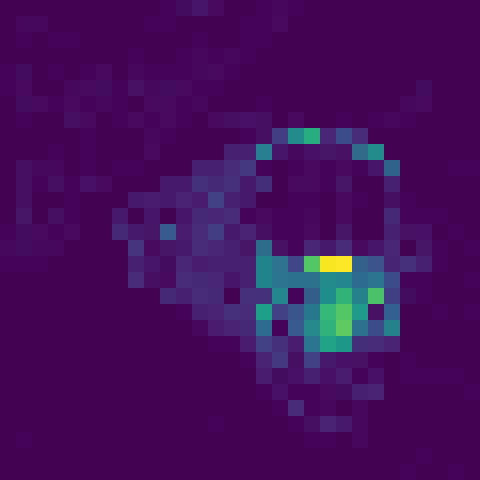} &
         \includegraphics[trim=2 0 0 2,clip,width=0.15\columnwidth]{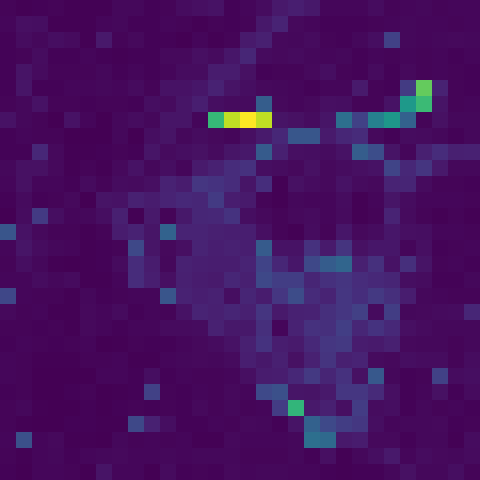} &
         \includegraphics[trim=2 0 0 2,clip,width=0.15\columnwidth]{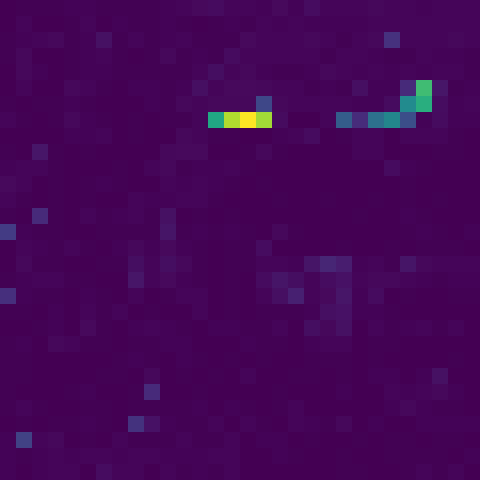} &
         \includegraphics[trim=2 0 0 2,clip,width=0.15\columnwidth]{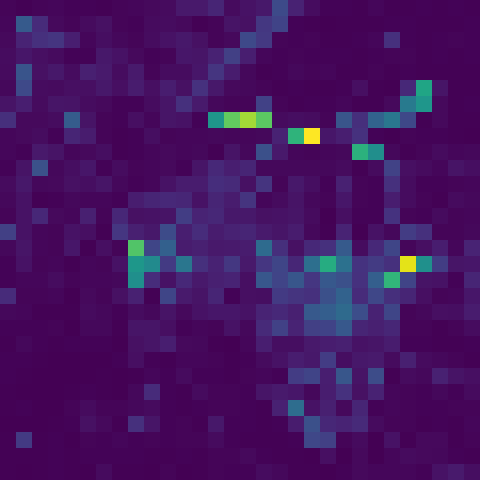} &
         \includegraphics[trim=2 0 0 2,clip,width=0.15\columnwidth]{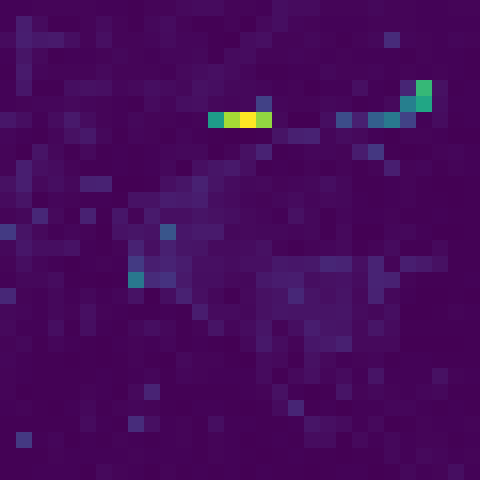} \\

    \end{tabular}
    \captionof{figure}{A qualitative comparison between the attention maps generated by ViT-S/16 (left) and ViT-S/16 + SAR (right). For each image, we show all the 6 attention maps ($A^h$) corresponding to the 6 last-block heads, computed using only the \texttt{[CLS]} token query.}
    \label{fig:attention-maps-supervised}
\end{figure*}

\subsection{Attention map analysis}
\label{sec:Attention-map-analysis}

This section qualitatively and quantitatively analyses the attention maps obtained using SAR.
Note that, as mentioned in \cref{sec:intro,sec:RelatedWork}, we do not directly use the attention map clusters for segmentation tasks or as input to a post-processing step. Thus, the goal of this analysis is to show that  the spatial entropy loss minimization effectively results in  attention maps with spatial clusters, leaving their potential use  for a segmentation-based post-processing as a future work.

\cref{fig:attention-maps-supervised} visually compares the attention maps obtained with ViT-S/16 and ViT-S/16 + SAR.
As expected,  standard training generates attention maps with a widely spread structure. Conversely, using SAR, a semantic segmentation structure clearly emerges.  In the Appendix, we show additional results.
For a quantitative analysis, we follow the protocol used in  \citep{DINO,IntriguingProperties-of-VTs}, where the Jaccard similarity is used to compare the ground-truth segmentation masks of the objects in PASCAL VOC-12  with the thresholded attention masks of the last ViT block. Specifically, the attention maps of all the heads are thresholded to keep $60\%$ of the mass, and the head with the highest Jaccard similarity with the ground-truth is selected \citep{DINO,IntriguingProperties-of-VTs}. \cref{tab:PASCAL-VOC} shows that  SAR significantly improves the segmentation results, quantitatively confirming the qualitative analysis in \cref{fig:attention-maps-supervised}.

\begin{table}[t]
\vspace{2mm}
\setlength{\tabcolsep}{1.0pt}
\small
    \caption{A  comparison of the segmentation properties of the attention maps on PASCAL VOC-12.}
    \label{tab:PASCAL-VOC}
    \centering
        \begin{tabularx}{\linewidth}{Xll}
        \toprule
        \textbf{Model} & \textbf{Jaccard similarity} ($\uparrow$)  \\
        \midrule
        ViT-S/16  & 19.18 \\
        \rowcolor{Light}ViT-S/16 + SAR  & \textbf{31.19}  \posimprov{12.01} \\
        \bottomrule
    \end{tabularx}
\end{table}

\section{Conclusions}
\label{sec:Conclusions}

In this paper we proposed 
SAR, a  regularization method which 
 exploits the connectedness property of the objects to introduce a local inductive bias 
into the VT training.  
By penalizing spatially disordered attention maps, an additional self-supervised signal can be extracted from the sample images, thereby reducing the reliance on large numbers of labeled training samples.
Using different downstream tasks and training-testing protocols (including fine-tuning and out-of-distribution testing), we showed tha SAR can significantly boost the accuracy of a ViT backbone, especially when the training data are scarce.
Despite SAR can also be used jointly with hybrid VTs, its main advantage 
over the latter
is the possibility to be easily plugged  into the original ViT backbone, whose architecture is widely adopted in many vision and vision language tasks.

\noindent{\bf Limitations.} Since training VTs is very computationally expensive, in our experiments we used only small/medium capacity VTs.
We leave the extension of our empirical analysis to larger capacity VTs for the future. 
For the same computational reasons, we have not tuned hyperparameters on the datasets. However, we believe that the SAR accuracy improvement, obtained in all the tested scenarios without hyperparameter tuning, further shows its robustness and ease to use.



\bibliography{egbib}

\begin{thebibliography}{}
\providecommand{\doi}[1]{\url{https://doi.org/#1}}
\bibcommenthead

\bibitem [\protect \citeauthoryear {%
Alayrac%
\ \protect \BOthers {.}}{%
Alayrac%
\ \protect \BOthers {.}}{%
{\protect \APACyear {2022}}%
}]{%
DBLP:journals/corr/abs-2204-14198}
\APACinsertmetastar {%
DBLP:journals/corr/abs-2204-14198}%
\begin{APACrefauthors}%
Alayrac, J.%
, Donahue, J.%
, Luc, P.%
, Miech, A.%
, Barr, I.%
, Hasson, Y.%
\BDBL {}Simonyan, K.%
\end{APACrefauthors}%
\unskip\
\newblock
\APACrefYearMonthDay{2022}{}{}.
\newblock
{\BBOQ}\APACrefatitle {Flamingo: a Visual Language Model for Few-Shot Learning}
  {Flamingo: a visual language model for few-shot learning}.{\BBCQ}
\newblock
\APACjournalVolNumPages{arXiv:2204.14198}{}{}{}.
\newblock

\newblock

\PrintBackRefs{\CurrentBib}

\bibitem [\protect \citeauthoryear {%
Altieri%
, Cocchi%
\BCBL {}\ \BBA {} Roli%
}{%
Altieri%
\ \protect \BOthers {.}}{%
{\protect \APACyear {2018}}%
}]{%
altieri2018spatentropy}
\APACinsertmetastar {%
altieri2018spatentropy}%
\begin{APACrefauthors}%
Altieri, L.%
, Cocchi, D.%
\BCBL {} Roli, G.%
\end{APACrefauthors}%
\unskip\
\newblock
\APACrefYearMonthDay{2018}{}{}.
\newblock
{\BBOQ}\APACrefatitle {{SpatEntropy: Spatial Entropy Measures in R}}
  {{SpatEntropy: Spatial Entropy Measures in R}}.{\BBCQ}
\newblock
\APACjournalVolNumPages{arXiv:1804.05521}{}{}{}.
\newblock

\newblock

\PrintBackRefs{\CurrentBib}

\bibitem [\protect \citeauthoryear {%
Asano%
, Rupprecht%
\BCBL {}\ \BBA {} Vedaldi%
}{%
Asano%
\ \protect \BOthers {.}}{%
{\protect \APACyear {2020}}%
}]{%
DBLP:conf/iclr/AsanoRV20a}
\APACinsertmetastar {%
DBLP:conf/iclr/AsanoRV20a}%
\begin{APACrefauthors}%
Asano, Y.M.%
, Rupprecht, C.%
\BCBL {} Vedaldi, A.%
\end{APACrefauthors}%
\unskip\
\newblock
\APACrefYearMonthDay{2020}{}{}.
\newblock
{\BBOQ}\APACrefatitle {Self-labelling via simultaneous clustering and
  representation learning} {Self-labelling via simultaneous clustering and
  representation learning}.{\BBCQ}
\newblock
 \APACrefbtitle {ICLR.} {Iclr.}
\PrintBackRefs{\CurrentBib}

\bibitem [\protect \citeauthoryear {%
Bachmann%
, Mizrahi%
, Atanov%
\BCBL {}\ \BBA {} Zamir%
}{%
Bachmann%
\ \protect \BOthers {.}}{%
{\protect \APACyear {2022}}%
}]{%
MultiMAE}
\APACinsertmetastar {%
MultiMAE}%
\begin{APACrefauthors}%
Bachmann, R.%
, Mizrahi, D.%
, Atanov, A.%
\BCBL {} Zamir, A.%
\end{APACrefauthors}%
\unskip\
\newblock
\APACrefYearMonthDay{2022}{}{}.
\newblock
{\BBOQ}\APACrefatitle {{MultiMAE: Multi-modal Multi-task Masked Autoencoders}}
  {{MultiMAE: Multi-modal Multi-task Masked Autoencoders}}.{\BBCQ}
\newblock
\APACjournalVolNumPages{arXiv:2204.01678}{}{}{}.
\newblock

\newblock

\PrintBackRefs{\CurrentBib}

\bibitem [\protect \citeauthoryear {%
Bai%
, Mei%
, Yuille%
\BCBL {}\ \BBA {} Xie%
}{%
Bai%
\ \protect \BOthers {.}}{%
{\protect \APACyear {2021}}%
}]{%
DBLP:conf/nips/BaiMYX21}
\APACinsertmetastar {%
DBLP:conf/nips/BaiMYX21}%
\begin{APACrefauthors}%
Bai, Y.%
, Mei, J.%
, Yuille, A.L.%
\BCBL {} Xie, C.%
\end{APACrefauthors}%
\unskip\
\newblock
\APACrefYearMonthDay{2021}{}{}.
\newblock
{\BBOQ}\APACrefatitle {Are Transformers more robust than CNNs?} {Are
  transformers more robust than cnns?}{\BBCQ}
\newblock
 \APACrefbtitle {NeurIPS.} {Neurips.}
\PrintBackRefs{\CurrentBib}

\bibitem [\protect \citeauthoryear {%
Balestriero%
, Bottou%
\BCBL {}\ \BBA {} LeCun%
}{%
Balestriero%
\ \protect \BOthers {.}}{%
{\protect \APACyear {2022}}%
}]{%
https://doi.org/10.48550/arxiv.2204.03632}
\APACinsertmetastar {%
https://doi.org/10.48550/arxiv.2204.03632}%
\begin{APACrefauthors}%
Balestriero, R.%
, Bottou, L.%
\BCBL {} LeCun, Y.%
\end{APACrefauthors}%
\unskip\
\newblock
\APACrefYearMonthDay{2022}{}{}.
\newblock
{\BBOQ}\APACrefatitle {The Effects of Regularization and Data Augmentation are
  Class Dependent} {The effects of regularization and data augmentation are
  class dependent}.{\BBCQ}
\newblock
\APACjournalVolNumPages{arXiv:2204.03632}{}{}{}.
\newblock

\newblock

\PrintBackRefs{\CurrentBib}

\bibitem [\protect \citeauthoryear {%
Bao%
, Dong%
\BCBL {}\ \BBA {} Wei%
}{%
Bao%
, Dong%
\BCBL {}\ \BBA {} Wei%
}{%
{\protect \APACyear {2022}}%
}]{%
BEIT}
\APACinsertmetastar {%
BEIT}%
\begin{APACrefauthors}%
Bao, H.%
, Dong, L.%
\BCBL {} Wei, F.%
\end{APACrefauthors}%
\unskip\
\newblock
\APACrefYearMonthDay{2022}{}{}.
\newblock
{\BBOQ}\APACrefatitle {{BEiT: BERT} Pre-Training of Image Transformers} {{BEiT:
  BERT} pre-training of image transformers}.{\BBCQ}
\newblock
\APACjournalVolNumPages{ICLR}{}{}{}.
\newblock

\newblock

\PrintBackRefs{\CurrentBib}

\bibitem [\protect \citeauthoryear {%
Bao%
, Wang%
, Dong%
\BCBL {}\ \BBA {} Wei%
}{%
Bao%
, Wang%
\BCBL {}\ \protect \BOthers {.}}{%
{\protect \APACyear {2022}}%
}]{%
DBLP:journals/corr/abs-2206-01127}
\APACinsertmetastar {%
DBLP:journals/corr/abs-2206-01127}%
\begin{APACrefauthors}%
Bao, H.%
, Wang, W.%
, Dong, L.%
\BCBL {} Wei, F.%
\end{APACrefauthors}%
\unskip\
\newblock
\APACrefYearMonthDay{2022}{}{}.
\newblock
{\BBOQ}\APACrefatitle {{VL-BEiT:} Generative Vision-Language Pretraining}
  {{VL-BEiT:} generative vision-language pretraining}.{\BBCQ}
\newblock
\APACjournalVolNumPages{arXiv:2206.01127}{}{}{}.
\newblock

\newblock

\PrintBackRefs{\CurrentBib}

\bibitem [\protect \citeauthoryear {%
Bardes%
, Ponce%
\BCBL {}\ \BBA {} LeCun%
}{%
Bardes%
\ \protect \BOthers {.}}{%
{\protect \APACyear {2022}}%
}]{%
VICReg}
\APACinsertmetastar {%
VICReg}%
\begin{APACrefauthors}%
Bardes, A.%
, Ponce, J.%
\BCBL {} LeCun, Y.%
\end{APACrefauthors}%
\unskip\
\newblock
\APACrefYearMonthDay{2022}{}{}.
\newblock
{\BBOQ}\APACrefatitle {{VICReg}: Variance-Invariance-Covariance Regularization
  for Self-Supervised Learning} {{VICReg}: Variance-invariance-covariance
  regularization for self-supervised learning}.{\BBCQ}
\newblock
\APACjournalVolNumPages{ICLR}{}{}{}.
\newblock

\newblock

\PrintBackRefs{\CurrentBib}

\bibitem [\protect \citeauthoryear {%
Batty%
}{%
Batty%
}{%
{\protect \APACyear {1974}}%
}]{%
Batty-spatial-entropy}
\APACinsertmetastar {%
Batty-spatial-entropy}%
\begin{APACrefauthors}%
Batty, M.%
\end{APACrefauthors}%
\unskip\
\newblock
\APACrefYearMonthDay{1974}{}{}.
\newblock
{\BBOQ}\APACrefatitle {Spatial entropy} {Spatial entropy}.{\BBCQ}
\newblock
\APACjournalVolNumPages{Geographical analysis}{6}{1}{1--31}.
\newblock

\newblock

\PrintBackRefs{\CurrentBib}

\bibitem [\protect \citeauthoryear {%
Bautista%
, Sanakoyeu%
, Tikhoncheva%
\BCBL {}\ \BBA {} Ommer%
}{%
Bautista%
\ \protect \BOthers {.}}{%
{\protect \APACyear {2016}}%
}]{%
NIPS2016_65fc52ed}
\APACinsertmetastar {%
NIPS2016_65fc52ed}%
\begin{APACrefauthors}%
Bautista, M.A.%
, Sanakoyeu, A.%
, Tikhoncheva, E.%
\BCBL {} Ommer, B.%
\end{APACrefauthors}%
\unskip\
\newblock
\APACrefYearMonthDay{2016}{}{}.
\newblock
{\BBOQ}\APACrefatitle {{CliqueCNN:} Deep Unsupervised Exemplar Learning}
  {{CliqueCNN:} deep unsupervised exemplar learning}.{\BBCQ}
\newblock
 \APACrefbtitle {NeurIPS.} {Neurips.}
\PrintBackRefs{\CurrentBib}

\bibitem [\protect \citeauthoryear {%
Cao%
\ \BBA {} Wu%
}{%
Cao%
\ \BBA {} Wu%
}{%
{\protect \APACyear {2021}}%
}]{%
DBLP:journals/corr/abs-2103-13559}
\APACinsertmetastar {%
DBLP:journals/corr/abs-2103-13559}%
\begin{APACrefauthors}%
Cao, Y.%
\BCBT {}\ \BBA {} Wu, J.%
\end{APACrefauthors}%
\unskip\
\newblock
\APACrefYearMonthDay{2021}{}{}.
\newblock
{\BBOQ}\APACrefatitle {Rethinking Self-Supervised Learning: Small is Beautiful}
  {Rethinking self-supervised learning: Small is beautiful}.{\BBCQ}
\newblock
\APACjournalVolNumPages{arXiv:2103.13559}{}{}{}.
\newblock

\newblock

\PrintBackRefs{\CurrentBib}

\bibitem [\protect \citeauthoryear {%
Carion%
\ \protect \BOthers {.}}{%
Carion%
\ \protect \BOthers {.}}{%
{\protect \APACyear {2020}}%
}]{%
DETR}
\APACinsertmetastar {%
DETR}%
\begin{APACrefauthors}%
Carion, N.%
, Massa, F.%
, Synnaeve, G.%
, Usunier, N.%
, Kirillov, A.%
\BCBL {} Zagoruyko, S.%
\end{APACrefauthors}%
\unskip\
\newblock
\APACrefYearMonthDay{2020}{}{}.
\newblock
{\BBOQ}\APACrefatitle {End-to-End Object Detection with Transformers}
  {End-to-end object detection with transformers}.{\BBCQ}
\newblock
 \APACrefbtitle {ECCV.} {Eccv.}
\PrintBackRefs{\CurrentBib}

\bibitem [\protect \citeauthoryear {%
Caron%
, Bojanowski%
, Joulin%
\BCBL {}\ \BBA {} Douze%
}{%
Caron%
\ \protect \BOthers {.}}{%
{\protect \APACyear {2018}}%
}]{%
DeepClustering}
\APACinsertmetastar {%
DeepClustering}%
\begin{APACrefauthors}%
Caron, M.%
, Bojanowski, P.%
, Joulin, A.%
\BCBL {} Douze, M.%
\end{APACrefauthors}%
\unskip\
\newblock
\APACrefYearMonthDay{2018}{}{}.
\newblock
{\BBOQ}\APACrefatitle {Deep Clustering for Unsupervised Learning of Visual
  Features} {Deep clustering for unsupervised learning of visual
  features}.{\BBCQ}
\newblock
 \APACrefbtitle {ECCV.} {Eccv.}
\PrintBackRefs{\CurrentBib}

\bibitem [\protect \citeauthoryear {%
Caron%
\ \protect \BOthers {.}}{%
Caron%
\ \protect \BOthers {.}}{%
{\protect \APACyear {2020}}%
}]{%
SwAV}
\APACinsertmetastar {%
SwAV}%
\begin{APACrefauthors}%
Caron, M.%
, Misra, I.%
, Mairal, J.%
, Goyal, P.%
, Bojanowski, P.%
\BCBL {} Joulin, A.%
\end{APACrefauthors}%
\unskip\
\newblock
\APACrefYearMonthDay{2020}{}{}.
\newblock
{\BBOQ}\APACrefatitle {Unsupervised Learning of Visual Features by Contrasting
  Cluster Assignments} {Unsupervised learning of visual features by contrasting
  cluster assignments}.{\BBCQ}
\newblock
 \APACrefbtitle {NeurIPS.} {Neurips.}
\PrintBackRefs{\CurrentBib}

\bibitem [\protect \citeauthoryear {%
Caron%
\ \protect \BOthers {.}}{%
Caron%
\ \protect \BOthers {.}}{%
{\protect \APACyear {2021}}%
}]{%
DINO}
\APACinsertmetastar {%
DINO}%
\begin{APACrefauthors}%
Caron, M.%
, Touvron, H.%
, Misra, I.%
, J{\'{e}}gou, H.%
, Mairal, J.%
, Bojanowski, P.%
\BCBL {} Joulin, A.%
\end{APACrefauthors}%
\unskip\
\newblock
\APACrefYearMonthDay{2021}{}{}.
\newblock
{\BBOQ}\APACrefatitle {Emerging Properties in Self-Supervised Vision
  Transformers} {Emerging properties in self-supervised vision
  transformers}.{\BBCQ}
\newblock
\APACjournalVolNumPages{ICCV}{}{}{}.
\newblock

\newblock

\PrintBackRefs{\CurrentBib}

\bibitem [\protect \citeauthoryear {%
Chang%
, Zhang%
, Jiang%
, Liu%
\BCBL {}\ \BBA {} Freeman%
}{%
Chang%
\ \protect \BOthers {.}}{%
{\protect \APACyear {2022}}%
}]{%
DBLP:journals/corr/abs-2202-04200}
\APACinsertmetastar {%
DBLP:journals/corr/abs-2202-04200}%
\begin{APACrefauthors}%
Chang, H.%
, Zhang, H.%
, Jiang, L.%
, Liu, C.%
\BCBL {} Freeman, W.T.%
\end{APACrefauthors}%
\unskip\
\newblock
\APACrefYearMonthDay{2022}{}{}.
\newblock
{\BBOQ}\APACrefatitle {{MaskGIT: Masked} Generative Image Transformer}
  {{MaskGIT: Masked} generative image transformer}.{\BBCQ}
\newblock
\APACjournalVolNumPages{arXiv:2202.04200}{}{}{}.
\newblock

\newblock

\PrintBackRefs{\CurrentBib}

\bibitem [\protect \citeauthoryear {%
M.~Chen%
\ \protect \BOthers {.}}{%
M.~Chen%
\ \protect \BOthers {.}}{%
{\protect \APACyear {2020}}%
}]{%
iGPT}
\APACinsertmetastar {%
iGPT}%
\begin{APACrefauthors}%
Chen, M.%
, Radford, A.%
, Child, R.%
, Wu, J.%
, Jun, H.%
, Luan, D.%
\BCBL {} Sutskever, I.%
\end{APACrefauthors}%
\unskip\
\newblock
\APACrefYearMonthDay{2020}{}{}.
\newblock
{\BBOQ}\APACrefatitle {Generative Pretraining From Pixels} {Generative
  pretraining from pixels}.{\BBCQ}
\newblock
 \APACrefbtitle {ICML.} {Icml.}
\PrintBackRefs{\CurrentBib}

\bibitem [\protect \citeauthoryear {%
T.~Chen%
, Kornblith%
, Norouzi%
\BCBL {}\ \BBA {} Hinton%
}{%
T.~Chen%
\ \protect \BOthers {.}}{%
{\protect \APACyear {2020}}%
}]{%
simclr}
\APACinsertmetastar {%
simclr}%
\begin{APACrefauthors}%
Chen, T.%
, Kornblith, S.%
, Norouzi, M.%
\BCBL {} Hinton, G.E.%
\end{APACrefauthors}%
\unskip\
\newblock
\APACrefYearMonthDay{2020}{}{}.
\newblock
{\BBOQ}\APACrefatitle {A Simple Framework for Contrastive Learning of Visual
  Representations} {A simple framework for contrastive learning of visual
  representations}.{\BBCQ}
\newblock
 \APACrefbtitle {ICML.} {Icml.}
\PrintBackRefs{\CurrentBib}

\bibitem [\protect \citeauthoryear {%
X.~Chen%
\ \protect \BOthers {.}}{%
X.~Chen%
\ \protect \BOthers {.}}{%
{\protect \APACyear {2022}}%
}]{%
DBLP:journals/corr/abs-2202-03026}
\APACinsertmetastar {%
DBLP:journals/corr/abs-2202-03026}%
\begin{APACrefauthors}%
Chen, X.%
, Ding, M.%
, Wang, X.%
, Xin, Y.%
, Mo, S.%
, Wang, Y.%
\BDBL {}Wang, J.%
\end{APACrefauthors}%
\unskip\
\newblock
\APACrefYearMonthDay{2022}{}{}.
\newblock
{\BBOQ}\APACrefatitle {Context Autoencoder for Self-Supervised Representation
  Learning} {Context autoencoder for self-supervised representation
  learning}.{\BBCQ}
\newblock
\APACjournalVolNumPages{arXiv:2202.03026}{}{}{}.
\newblock

\newblock

\PrintBackRefs{\CurrentBib}

\bibitem [\protect \citeauthoryear {%
X.~Chen%
\ \BBA {} He%
}{%
X.~Chen%
\ \BBA {} He%
}{%
{\protect \APACyear {2021}}%
}]{%
simsiam}
\APACinsertmetastar {%
simsiam}%
\begin{APACrefauthors}%
Chen, X.%
\BCBT {}\ \BBA {} He, K.%
\end{APACrefauthors}%
\unskip\
\newblock
\APACrefYearMonthDay{2021}{}{}.
\newblock
{\BBOQ}\APACrefatitle {Exploring Simple Siamese Representation Learning}
  {Exploring simple siamese representation learning}.{\BBCQ}
\newblock
 \APACrefbtitle {CVPR.} {Cvpr.}
\PrintBackRefs{\CurrentBib}

\bibitem [\protect \citeauthoryear {%
X.~Chen%
, Xie%
\BCBL {}\ \BBA {} He%
}{%
X.~Chen%
\ \protect \BOthers {.}}{%
{\protect \APACyear {2021}}%
}]{%
MoCo-v3}
\APACinsertmetastar {%
MoCo-v3}%
\begin{APACrefauthors}%
Chen, X.%
, Xie, S.%
\BCBL {} He, K.%
\end{APACrefauthors}%
\unskip\
\newblock
\APACrefYearMonthDay{2021}{}{}.
\newblock
{\BBOQ}\APACrefatitle {An Empirical Study of Training Self-Supervised Vision
  Transformers} {An empirical study of training self-supervised vision
  transformers}.{\BBCQ}
\newblock
\APACjournalVolNumPages{ICCV}{}{}{}.
\newblock

\newblock

\PrintBackRefs{\CurrentBib}

\bibitem [\protect \citeauthoryear {%
Dai%
, Cai%
, Lin%
\BCBL {}\ \BBA {} Chen%
}{%
Dai%
\ \protect \BOthers {.}}{%
{\protect \APACyear {2021}}%
}]{%
UP-DETR}
\APACinsertmetastar {%
UP-DETR}%
\begin{APACrefauthors}%
Dai, Z.%
, Cai, B.%
, Lin, Y.%
\BCBL {} Chen, J.%
\end{APACrefauthors}%
\unskip\
\newblock
\APACrefYearMonthDay{2021}{}{}.
\newblock
{\BBOQ}\APACrefatitle {{UP-DETR:} Unsupervised Pre-training for Object
  Detection with Transformers} {{UP-DETR:} unsupervised pre-training for object
  detection with transformers}.{\BBCQ}
\newblock
 \APACrefbtitle {CVPR.} {Cvpr.}
\PrintBackRefs{\CurrentBib}

\bibitem [\protect \citeauthoryear {%
d'Ascoli%
\ \protect \BOthers {.}}{%
d'Ascoli%
\ \protect \BOthers {.}}{%
{\protect \APACyear {2021}}%
}]{%
ConViT}
\APACinsertmetastar {%
ConViT}%
\begin{APACrefauthors}%
d'Ascoli, S.%
, Touvron, H.%
, Leavitt, M.L.%
, Morcos, A.S.%
, Biroli, G.%
\BCBL {} Sagun, L.%
\end{APACrefauthors}%
\unskip\
\newblock
\APACrefYearMonthDay{2021}{}{}.
\newblock
{\BBOQ}\APACrefatitle {{ConViT: I}mproving Vision Transformers with Soft
  Convolutional Inductive Biases} {{ConViT: I}mproving vision transformers with
  soft convolutional inductive biases}.{\BBCQ}
\newblock
 \APACrefbtitle {ICML.} {Icml.}
\PrintBackRefs{\CurrentBib}

\bibitem [\protect \citeauthoryear {%
Devlin%
, Chang%
, Lee%
\BCBL {}\ \BBA {} Toutanova%
}{%
Devlin%
\ \protect \BOthers {.}}{%
{\protect \APACyear {2019}}%
}]{%
devlin-etal-2019-bert}
\APACinsertmetastar {%
devlin-etal-2019-bert}%
\begin{APACrefauthors}%
Devlin, J.%
, Chang, M\BHBI W.%
, Lee, K.%
\BCBL {} Toutanova, K.%
\end{APACrefauthors}%
\unskip\
\newblock
\APACrefYearMonthDay{2019}{}{}.
\newblock
{\BBOQ}\APACrefatitle {{BERT}: Pre-training of Deep Bidirectional Transformers
  for Language Understanding} {{BERT}: Pre-training of deep bidirectional
  transformers for language understanding}.{\BBCQ}
\newblock
 \APACrefbtitle {NAACL,.} {Naacl,.}
\PrintBackRefs{\CurrentBib}

\bibitem [\protect \citeauthoryear {%
Didolkar%
\ \protect \BOthers {.}}{%
Didolkar%
\ \protect \BOthers {.}}{%
{\protect \APACyear {2021}}%
}]{%
DBLP:conf/nips/DidolkarGKBBHMB21}
\APACinsertmetastar {%
DBLP:conf/nips/DidolkarGKBBHMB21}%
\begin{APACrefauthors}%
Didolkar, A.%
, Goyal, A.%
, Ke, N.R.%
, Blundell, C.%
, Beaudoin, P.%
, Heess, N.%
\BDBL {}Bengio, Y.%
\end{APACrefauthors}%
\unskip\
\newblock
\APACrefYearMonthDay{2021}{}{}.
\newblock
{\BBOQ}\APACrefatitle {Neural Production Systems} {Neural production
  systems}.{\BBCQ}
\newblock
 \APACrefbtitle {NIPS.} {Nips.}
\PrintBackRefs{\CurrentBib}

\bibitem [\protect \citeauthoryear {%
Dong%
\ \protect \BOthers {.}}{%
Dong%
\ \protect \BOthers {.}}{%
{\protect \APACyear {2021}}%
}]{%
DBLP:journals/corr/abs-2111-12710}
\APACinsertmetastar {%
DBLP:journals/corr/abs-2111-12710}%
\begin{APACrefauthors}%
Dong, X.%
, Bao, J.%
, Zhang, T.%
, Chen, D.%
, Zhang, W.%
, Yuan, L.%
\BDBL {}Yu, N.%
\end{APACrefauthors}%
\unskip\
\newblock
\APACrefYearMonthDay{2021}{}{}.
\newblock
{\BBOQ}\APACrefatitle {{PeCo: Perceptual Codebook for BERT Pre-training of
  Vision Transformers}} {{PeCo: Perceptual Codebook for BERT Pre-training of
  Vision Transformers}}.{\BBCQ}
\newblock
\APACjournalVolNumPages{arXiv:2111.12710}{}{}{}.
\newblock

\newblock

\PrintBackRefs{\CurrentBib}

\bibitem [\protect \citeauthoryear {%
Dosovitskiy%
\ \protect \BOthers {.}}{%
Dosovitskiy%
\ \protect \BOthers {.}}{%
{\protect \APACyear {2021}}%
}]{%
ViT}
\APACinsertmetastar {%
ViT}%
\begin{APACrefauthors}%
Dosovitskiy, A.%
, Beyer, L.%
, Kolesnikov, A.%
, Weissenborn, D.%
, Zhai, X.%
, Unterthiner, T.%
\BDBL {}Houlsby, N.%
\end{APACrefauthors}%
\unskip\
\newblock
\APACrefYearMonthDay{2021}{}{}.
\newblock
{\BBOQ}\APACrefatitle {An Image is Worth 16x16 Words: Transformers for Image
  Recognition at Scale} {An image is worth 16x16 words: Transformers for image
  recognition at scale}.{\BBCQ}
\newblock
 \APACrefbtitle {ICLR.} {Iclr.}
\PrintBackRefs{\CurrentBib}

\bibitem [\protect \citeauthoryear {%
Dwibedi%
, Aytar%
, Tompson%
, Sermanet%
\BCBL {}\ \BBA {} Zisserman%
}{%
Dwibedi%
\ \protect \BOthers {.}}{%
{\protect \APACyear {2021}}%
}]{%
little-friends}
\APACinsertmetastar {%
little-friends}%
\begin{APACrefauthors}%
Dwibedi, D.%
, Aytar, Y.%
, Tompson, J.%
, Sermanet, P.%
\BCBL {} Zisserman, A.%
\end{APACrefauthors}%
\unskip\
\newblock
\APACrefYearMonthDay{2021}{}{}.
\newblock
{\BBOQ}\APACrefatitle {With a Little Help from My Friends: Nearest-Neighbor
  Contrastive Learning of Visual Representations} {With a little help from my
  friends: Nearest-neighbor contrastive learning of visual
  representations}.{\BBCQ}
\newblock
 \APACrefbtitle {ICCV.} {Iccv.}
\PrintBackRefs{\CurrentBib}

\bibitem [\protect \citeauthoryear {%
El{-}Nouby%
\ \protect \BOthers {.}}{%
El{-}Nouby%
\ \protect \BOthers {.}}{%
{\protect \APACyear {2021}}%
}]{%
DBLP:journals/corr/abs-2112-10740}
\APACinsertmetastar {%
DBLP:journals/corr/abs-2112-10740}%
\begin{APACrefauthors}%
El{-}Nouby, A.%
, Izacard, G.%
, Touvron, H.%
, Laptev, I.%
, J{\'{e}}gou, H.%
\BCBL {} Grave, E.%
\end{APACrefauthors}%
\unskip\
\newblock
\APACrefYearMonthDay{2021}{}{}.
\newblock
{\BBOQ}\APACrefatitle {Are Large-scale Datasets Necessary for Self-Supervised
  Pre-training?} {Are large-scale datasets necessary for self-supervised
  pre-training?}{\BBCQ}
\newblock
\APACjournalVolNumPages{arXiv:2112.10740}{}{}{}.
\newblock

\newblock

\PrintBackRefs{\CurrentBib}

\bibitem [\protect \citeauthoryear {%
Engelcke%
, Parker~Jones%
\BCBL {}\ \BBA {} Posner%
}{%
Engelcke%
\ \protect \BOthers {.}}{%
{\protect \APACyear {2021}}%
}]{%
Genesisv2}
\APACinsertmetastar {%
Genesisv2}%
\begin{APACrefauthors}%
Engelcke, M.%
, Parker~Jones, O.%
\BCBL {} Posner, I.%
\end{APACrefauthors}%
\unskip\
\newblock
\APACrefYearMonthDay{2021}{}{}.
\newblock
{\BBOQ}\APACrefatitle {Genesis-v2: Inferring unordered object representations
  without iterative refinement} {Genesis-v2: Inferring unordered object
  representations without iterative refinement}.{\BBCQ}
\newblock
\APACjournalVolNumPages{NeurIPS}{34}{}{8085--8094}.
\newblock

\newblock

\PrintBackRefs{\CurrentBib}

\bibitem [\protect \citeauthoryear {%
Ermolov%
, Siarohin%
, Sangineto%
\BCBL {}\ \BBA {} Sebe%
}{%
Ermolov%
\ \protect \BOthers {.}}{%
{\protect \APACyear {2021}}%
}]{%
w-mse}
\APACinsertmetastar {%
w-mse}%
\begin{APACrefauthors}%
Ermolov, A.%
, Siarohin, A.%
, Sangineto, E.%
\BCBL {} Sebe, N.%
\end{APACrefauthors}%
\unskip\
\newblock
\APACrefYearMonthDay{2021}{}{}.
\newblock
{\BBOQ}\APACrefatitle {Whitening for Self-Supervised Representation Learning}
  {Whitening for self-supervised representation learning}.{\BBCQ}
\newblock
 \APACrefbtitle {ICML.} {Icml.}
\PrintBackRefs{\CurrentBib}

\bibitem [\protect \citeauthoryear {%
Everingham%
, Gool%
, Williams%
, Winn%
\BCBL {}\ \BBA {} Zisserman%
}{%
Everingham%
\ \protect \BOthers {.}}{%
{\protect \APACyear {2010}}%
}]{%
Pascal-voc-dataset}
\APACinsertmetastar {%
Pascal-voc-dataset}%
\begin{APACrefauthors}%
Everingham, M.%
, Gool, L.V.%
, Williams, C.K.I.%
, Winn, J.M.%
\BCBL {} Zisserman, A.%
\end{APACrefauthors}%
\unskip\
\newblock
\APACrefYearMonthDay{2010}{}{}.
\newblock
{\BBOQ}\APACrefatitle {The Pascal Visual Object Classes {(VOC)} Challenge} {The
  pascal visual object classes {(VOC)} challenge}.{\BBCQ}
\newblock
\APACjournalVolNumPages{Int. J. Comput. Vis.}{88}{2}{303--338}.
\newblock

\newblock

\PrintBackRefs{\CurrentBib}

\bibitem [\protect \citeauthoryear {%
Gansbeke%
, Vandenhende%
, Georgoulis%
, Proesmans%
\BCBL {}\ \BBA {} Gool%
}{%
Gansbeke%
\ \protect \BOthers {.}}{%
{\protect \APACyear {2020}}%
}]{%
DBLP:conf/eccv/GansbekeVGPG20}
\APACinsertmetastar {%
DBLP:conf/eccv/GansbekeVGPG20}%
\begin{APACrefauthors}%
Gansbeke, W.V.%
, Vandenhende, S.%
, Georgoulis, S.%
, Proesmans, M.%
\BCBL {} Gool, L.V.%
\end{APACrefauthors}%
\unskip\
\newblock
\APACrefYearMonthDay{2020}{}{}.
\newblock
{\BBOQ}\APACrefatitle {{SCAN:} Learning to Classify Images Without Labels}
  {{SCAN:} learning to classify images without labels}.{\BBCQ}
\newblock
 \APACrefbtitle {ECCV.} {Eccv.}
\PrintBackRefs{\CurrentBib}

\bibitem [\protect \citeauthoryear {%
Girshick%
}{%
Girshick%
}{%
{\protect \APACyear {2015}}%
}]{%
DBLP:conf/iccv/Girshick15}
\APACinsertmetastar {%
DBLP:conf/iccv/Girshick15}%
\begin{APACrefauthors}%
Girshick, R.B.%
\end{APACrefauthors}%
\unskip\
\newblock
\APACrefYearMonthDay{2015}{}{}.
\newblock
{\BBOQ}\APACrefatitle {Fast {R-CNN}} {Fast {R-CNN}}.{\BBCQ}
\newblock
 \APACrefbtitle {ICCV} {Iccv}\ (\BPGS\ 1440--1448).
\PrintBackRefs{\CurrentBib}

\bibitem [\protect \citeauthoryear {%
Goyal%
\ \protect \BOthers {.}}{%
Goyal%
\ \protect \BOthers {.}}{%
{\protect \APACyear {2020}}%
}]{%
DBLP:journals/corr/abs-2006-16225}
\APACinsertmetastar {%
DBLP:journals/corr/abs-2006-16225}%
\begin{APACrefauthors}%
Goyal, A.%
, Lamb, A.%
, Gampa, P.%
, Beaudoin, P.%
, Levine, S.%
, Blundell, C.%
\BDBL {}Mozer, M.%
\end{APACrefauthors}%
\unskip\
\newblock
\APACrefYearMonthDay{2020}{}{}.
\newblock
{\BBOQ}\APACrefatitle {Object Files and Schemata: Factorizing Declarative and
  Procedural Knowledge in Dynamical Systems} {Object files and schemata:
  Factorizing declarative and procedural knowledge in dynamical
  systems}.{\BBCQ}
\newblock
\APACjournalVolNumPages{arXiv:2006.16225}{}{}{}.
\newblock

\newblock

\PrintBackRefs{\CurrentBib}

\bibitem [\protect \citeauthoryear {%
Grana%
, Borghesani%
\BCBL {}\ \BBA {} Cucchiara%
}{%
Grana%
\ \protect \BOthers {.}}{%
{\protect \APACyear {2010}}%
}]{%
DBLP:journals/tip/GranaBC10}
\APACinsertmetastar {%
DBLP:journals/tip/GranaBC10}%
\begin{APACrefauthors}%
Grana, C.%
, Borghesani, D.%
\BCBL {} Cucchiara, R.%
\end{APACrefauthors}%
\unskip\
\newblock
\APACrefYearMonthDay{2010}{}{}.
\newblock
{\BBOQ}\APACrefatitle {Optimized Block-Based Connected Components Labeling With
  Decision Trees} {Optimized block-based connected components labeling with
  decision trees}.{\BBCQ}
\newblock
\APACjournalVolNumPages{{IEEE} Trans. Image Process.}{19}{6}{1596--1609}.
\newblock

\newblock

\PrintBackRefs{\CurrentBib}

\bibitem [\protect \citeauthoryear {%
Grill%
\ \protect \BOthers {.}}{%
Grill%
\ \protect \BOthers {.}}{%
{\protect \APACyear {2020}}%
}]{%
byol}
\APACinsertmetastar {%
byol}%
\begin{APACrefauthors}%
Grill, J\BHBI B.%
, Strub, F.%
, Altché, F.%
, Tallec, C.%
, Richemond, P.H.%
, Buchatskaya, E.%
\BDBL {}Valko, M.%
\end{APACrefauthors}%
\unskip\
\newblock
\APACrefYearMonthDay{2020}{}{}.
\newblock
{\BBOQ}\APACrefatitle {Bootstrap your own latent: A new approach to
  self-supervised Learning} {Bootstrap your own latent: A new approach to
  self-supervised learning}.{\BBCQ}
\newblock
 \APACrefbtitle {NeurIPS.} {Neurips.}
\PrintBackRefs{\CurrentBib}

\bibitem [\protect \citeauthoryear {%
Guo%
\ \protect \BOthers {.}}{%
Guo%
\ \protect \BOthers {.}}{%
{\protect \APACyear {2021}}%
}]{%
DBLP:journals/cvm/GuoCLMMH21}
\APACinsertmetastar {%
DBLP:journals/cvm/GuoCLMMH21}%
\begin{APACrefauthors}%
Guo, M.%
, Cai, J.%
, Liu, Z.%
, Mu, T.%
, Martin, R.R.%
\BCBL {} Hu, S.%
\end{APACrefauthors}%
\unskip\
\newblock
\APACrefYearMonthDay{2021}{}{}.
\newblock
{\BBOQ}\APACrefatitle {{PCT: P}oint cloud transformer} {{PCT: P}oint cloud
  transformer}.{\BBCQ}
\newblock
\APACjournalVolNumPages{Comput. Vis. Media}{7}{2}{187--199}.
\newblock

\newblock

\PrintBackRefs{\CurrentBib}

\bibitem [\protect \citeauthoryear {%
Hassani%
, Walton%
, Li%
, Li%
\BCBL {}\ \BBA {} Shi%
}{%
Hassani%
\ \protect \BOthers {.}}{%
{\protect \APACyear {2022}}%
}]{%
DBLP:journals/corr/abs-2204-07143}
\APACinsertmetastar {%
DBLP:journals/corr/abs-2204-07143}%
\begin{APACrefauthors}%
Hassani, A.%
, Walton, S.%
, Li, J.%
, Li, S.%
\BCBL {} Shi, H.%
\end{APACrefauthors}%
\unskip\
\newblock
\APACrefYearMonthDay{2022}{}{}.
\newblock
{\BBOQ}\APACrefatitle {Neighborhood Attention Transformer} {Neighborhood
  attention transformer}.{\BBCQ}
\newblock
\APACjournalVolNumPages{arXiv:2204.07143}{}{}{}.
\newblock

\newblock

\PrintBackRefs{\CurrentBib}

\bibitem [\protect \citeauthoryear {%
He%
\ \protect \BOthers {.}}{%
He%
\ \protect \BOthers {.}}{%
{\protect \APACyear {2022}}%
}]{%
DBLP:journals/corr/abs-2111-06377}
\APACinsertmetastar {%
DBLP:journals/corr/abs-2111-06377}%
\begin{APACrefauthors}%
He, K.%
, Chen, X.%
, Xie, S.%
, Li, Y.%
, Doll{\'{a}}r, P.%
\BCBL {} Girshick, R.B.%
\end{APACrefauthors}%
\unskip\
\newblock
\APACrefYearMonthDay{2022}{}{}.
\newblock
{\BBOQ}\APACrefatitle {Masked Autoencoders Are Scalable Vision Learners}
  {Masked autoencoders are scalable vision learners}.{\BBCQ}
\newblock
\APACjournalVolNumPages{CVPR}{}{}{}.
\newblock

\newblock

\PrintBackRefs{\CurrentBib}

\bibitem [\protect \citeauthoryear {%
He%
, Fan%
, Wu%
, Xie%
\BCBL {}\ \BBA {} Girshick%
}{%
He%
\ \protect \BOthers {.}}{%
{\protect \APACyear {2020}}%
}]{%
MoCo}
\APACinsertmetastar {%
MoCo}%
\begin{APACrefauthors}%
He, K.%
, Fan, H.%
, Wu, Y.%
, Xie, S.%
\BCBL {} Girshick, R.%
\end{APACrefauthors}%
\unskip\
\newblock
\APACrefYearMonthDay{2020}{}{}.
\newblock
{\BBOQ}\APACrefatitle {Momentum Contrast for Unsupervised Visual Representation
  Learning} {Momentum contrast for unsupervised visual representation
  learning}.{\BBCQ}
\newblock
 \APACrefbtitle {CVPR.} {Cvpr.}
\PrintBackRefs{\CurrentBib}

\bibitem [\protect \citeauthoryear {%
He%
, Gkioxari%
, Doll{\'a}r%
\BCBL {}\ \BBA {} Girshick%
}{%
He%
\ \protect \BOthers {.}}{%
{\protect \APACyear {2017}}%
}]{%
he2017mask}
\APACinsertmetastar {%
he2017mask}%
\begin{APACrefauthors}%
He, K.%
, Gkioxari, G.%
, Doll{\'a}r, P.%
\BCBL {} Girshick, R.%
\end{APACrefauthors}%
\unskip\
\newblock
\APACrefYearMonthDay{2017}{}{}.
\newblock
{\BBOQ}\APACrefatitle {Mask {R-CNN}} {Mask {R-CNN}}.{\BBCQ}
\newblock
 \APACrefbtitle {ICCV.} {Iccv.}
\PrintBackRefs{\CurrentBib}

\bibitem [\protect \citeauthoryear {%
He%
, Zhang%
, Ren%
\BCBL {}\ \BBA {} Sun%
}{%
He%
\ \protect \BOthers {.}}{%
{\protect \APACyear {2016}}%
}]{%
he2016deep}
\APACinsertmetastar {%
he2016deep}%
\begin{APACrefauthors}%
He, K.%
, Zhang, X.%
, Ren, S.%
\BCBL {} Sun, J.%
\end{APACrefauthors}%
\unskip\
\newblock
\APACrefYearMonthDay{2016}{}{}.
\newblock
{\BBOQ}\APACrefatitle {Deep residual learning for image recognition} {Deep
  residual learning for image recognition}.{\BBCQ}
\newblock
 \APACrefbtitle {CVPR.} {Cvpr.}
\PrintBackRefs{\CurrentBib}

\bibitem [\protect \citeauthoryear {%
Hendrycks%
\ \BBA {} Dietterich%
}{%
Hendrycks%
\ \BBA {} Dietterich%
}{%
{\protect \APACyear {2018}}%
}]{%
DBLP:journals/corr/abs-1807-01697}
\APACinsertmetastar {%
DBLP:journals/corr/abs-1807-01697}%
\begin{APACrefauthors}%
Hendrycks, D.%
\BCBT {}\ \BBA {} Dietterich, T.G.%
\end{APACrefauthors}%
\unskip\
\newblock
\APACrefYearMonthDay{2018}{}{}.
\newblock
{\BBOQ}\APACrefatitle {Benchmarking Neural Network Robustness to Common
  Corruptions and Perturbations} {Benchmarking neural network robustness to
  common corruptions and perturbations}.{\BBCQ}
\newblock
 \APACrefbtitle {ICLR.} {Iclr.}
\PrintBackRefs{\CurrentBib}

\bibitem [\protect \citeauthoryear {%
Hendrycks%
, Zhao%
, Basart%
, Steinhardt%
\BCBL {}\ \BBA {} Song%
}{%
Hendrycks%
\ \protect \BOthers {.}}{%
{\protect \APACyear {2019}}%
}]{%
DBLP:journals/corr/abs-1907-07174}
\APACinsertmetastar {%
DBLP:journals/corr/abs-1907-07174}%
\begin{APACrefauthors}%
Hendrycks, D.%
, Zhao, K.%
, Basart, S.%
, Steinhardt, J.%
\BCBL {} Song, D.%
\end{APACrefauthors}%
\unskip\
\newblock
\APACrefYearMonthDay{2019}{}{}.
\newblock
{\BBOQ}\APACrefatitle {Natural Adversarial Examples} {Natural adversarial
  examples}.{\BBCQ}
\newblock
 \APACrefbtitle {CVPR.} {Cvpr.}
\PrintBackRefs{\CurrentBib}

\bibitem [\protect \citeauthoryear {%
Herzig%
\ \protect \BOthers {.}}{%
Herzig%
\ \protect \BOthers {.}}{%
{\protect \APACyear {2022}}%
}]{%
ObjectRegionVideoTransformer}
\APACinsertmetastar {%
ObjectRegionVideoTransformer}%
\begin{APACrefauthors}%
Herzig, R.%
, Ben-Avraham, E.%
, Mangalam, K.%
, Bar, A.%
, Chechik, G.%
, Rohrbach, A.%
\BDBL {}Globerson, A.%
\end{APACrefauthors}%
\unskip\
\newblock
\APACrefYearMonthDay{2022}{}{}.
\newblock
{\BBOQ}\APACrefatitle {Object-region video transformers} {Object-region video
  transformers}.{\BBCQ}
\newblock
 \APACrefbtitle {CVPR} {Cvpr}\ (\BPGS\ 3148--3159).
\PrintBackRefs{\CurrentBib}

\bibitem [\protect \citeauthoryear {%
Hjelm%
\ \protect \BOthers {.}}{%
Hjelm%
\ \protect \BOthers {.}}{%
{\protect \APACyear {2019}}%
}]{%
DIM}
\APACinsertmetastar {%
DIM}%
\begin{APACrefauthors}%
Hjelm, R.D.%
, Fedorov, A.%
, Lavoie{-}Marchildon, S.%
, Grewal, K.%
, Bachman, P.%
, Trischler, A.%
\BCBL {} Bengio, Y.%
\end{APACrefauthors}%
\unskip\
\newblock
\APACrefYearMonthDay{2019}{}{}.
\newblock
{\BBOQ}\APACrefatitle {Learning deep representations by mutual information
  estimation and maximization} {Learning deep representations by mutual
  information estimation and maximization}.{\BBCQ}
\newblock
 \APACrefbtitle {ICLR.} {Iclr.}
\PrintBackRefs{\CurrentBib}

\bibitem [\protect \citeauthoryear {%
Hua%
, Tian%
, Ren%
, Zhao%
\BCBL {}\ \BBA {} Sigal%
}{%
Hua%
\ \protect \BOthers {.}}{%
{\protect \APACyear {2022}}%
}]{%
RandSAC}
\APACinsertmetastar {%
RandSAC}%
\begin{APACrefauthors}%
Hua, T.%
, Tian, Y.%
, Ren, S.%
, Zhao, H.%
\BCBL {} Sigal, L.%
\end{APACrefauthors}%
\unskip\
\newblock
\APACrefYearMonthDay{2022}{}{}.
\newblock
{\BBOQ}\APACrefatitle {{Self-supervision through Random Segments with
  Autoregressive Coding (RandSAC)}} {{Self-supervision through Random Segments
  with Autoregressive Coding (RandSAC)}}.{\BBCQ}
\newblock
\APACjournalVolNumPages{arXiv:2203.12054}{}{}{}.
\newblock

\newblock

\PrintBackRefs{\CurrentBib}

\bibitem [\protect \citeauthoryear {%
Hua%
\ \protect \BOthers {.}}{%
Hua%
\ \protect \BOthers {.}}{%
{\protect \APACyear {2021}}%
}]{%
hua2021feature}
\APACinsertmetastar {%
hua2021feature}%
\begin{APACrefauthors}%
Hua, T.%
, Wang, W.%
, Xue, Z.%
, Wang, Y.%
, Ren, S.%
\BCBL {} Zhao, H.%
\end{APACrefauthors}%
\unskip\
\newblock
\APACrefYearMonthDay{2021}{}{}.
\newblock
{\BBOQ}\APACrefatitle {On Feature Decorrelation in Self-Supervised Learning}
  {On feature decorrelation in self-supervised learning}.{\BBCQ}
\newblock
\APACjournalVolNumPages{arXiv:2105.00470}{}{}{}.
\newblock

\newblock

\PrintBackRefs{\CurrentBib}

\bibitem [\protect \citeauthoryear {%
Hudson%
\ \BBA {} Zitnick%
}{%
Hudson%
\ \BBA {} Zitnick%
}{%
{\protect \APACyear {2021}}%
}]{%
GAT}
\APACinsertmetastar {%
GAT}%
\begin{APACrefauthors}%
Hudson, D.A.%
\BCBT {}\ \BBA {} Zitnick, C.L.%
\end{APACrefauthors}%
\unskip\
\newblock
\APACrefYearMonthDay{2021}{}{}.
\newblock
{\BBOQ}\APACrefatitle {{Generative Adversarial Transformers}} {{Generative
  Adversarial Transformers}}.{\BBCQ}
\newblock
 \APACrefbtitle {ICML.} {Icml.}
\PrintBackRefs{\CurrentBib}

\bibitem [\protect \citeauthoryear {%
Ji%
, Henriques%
\BCBL {}\ \BBA {} Vedaldi%
}{%
Ji%
\ \protect \BOthers {.}}{%
{\protect \APACyear {2019}}%
}]{%
IIC}
\APACinsertmetastar {%
IIC}%
\begin{APACrefauthors}%
Ji, X.%
, Henriques, J.F.%
\BCBL {} Vedaldi, A.%
\end{APACrefauthors}%
\unskip\
\newblock
\APACrefYearMonthDay{2019}{}{}.
\newblock
{\BBOQ}\APACrefatitle {Invariant Information Clustering for Unsupervised Image
  Classification and Segmentation} {Invariant information clustering for
  unsupervised image classification and segmentation}.{\BBCQ}
\newblock
 \APACrefbtitle {ICCV.} {Iccv.}
\PrintBackRefs{\CurrentBib}

\bibitem [\protect \citeauthoryear {%
Jiang%
, Chang%
\BCBL {}\ \BBA {} Wang%
}{%
Jiang%
\ \protect \BOthers {.}}{%
{\protect \APACyear {2021}}%
}]{%
TransGAN}
\APACinsertmetastar {%
TransGAN}%
\begin{APACrefauthors}%
Jiang, Y.%
, Chang, S.%
\BCBL {} Wang, Z.%
\end{APACrefauthors}%
\unskip\
\newblock
\APACrefYearMonthDay{2021}{}{}.
\newblock
{\BBOQ}\APACrefatitle {{TransGAN}: Two Transformers Can Make One Strong {GAN}}
  {{TransGAN}: Two transformers can make one strong {GAN}}.{\BBCQ}
\newblock
\APACjournalVolNumPages{NIPS}{}{}{}.
\newblock

\newblock

\PrintBackRefs{\CurrentBib}

\bibitem [\protect \citeauthoryear {%
Kakogeorgiou%
\ \protect \BOthers {.}}{%
Kakogeorgiou%
\ \protect \BOthers {.}}{%
{\protect \APACyear {2022}}%
}]{%
What-to-Hide}
\APACinsertmetastar {%
What-to-Hide}%
\begin{APACrefauthors}%
Kakogeorgiou, I.%
, Gidaris, S.%
, Psomas, B.%
, Avrithis, Y.%
, Bursuc, A.%
, Karantzalos, K.%
\BCBL {} Komodakis, N.%
\end{APACrefauthors}%
\unskip\
\newblock
\APACrefYearMonthDay{2022}{}{}.
\newblock
{\BBOQ}\APACrefatitle {{What to Hide from Your Students: Attention-Guided
  Masked Image Modeling}} {{What to Hide from Your Students: Attention-Guided
  Masked Image Modeling}}.{\BBCQ}
\newblock
\APACjournalVolNumPages{arXiv:2203.12719}{}{}{}.
\newblock

\newblock

\PrintBackRefs{\CurrentBib}

\bibitem [\protect \citeauthoryear {%
Kang%
, Mo%
\BCBL {}\ \BBA {} Shin%
}{%
Kang%
\ \protect \BOthers {.}}{%
{\protect \APACyear {2022}}%
}]{%
ReMixer}
\APACinsertmetastar {%
ReMixer}%
\begin{APACrefauthors}%
Kang, H.%
, Mo, S.%
\BCBL {} Shin, J.%
\end{APACrefauthors}%
\unskip\
\newblock
\APACrefYearMonthDay{2022}{}{}.
\newblock
{\BBOQ}\APACrefatitle {ReMixer: Object-aware Mixing Layer for Vision
  Transformers and Mixers} {Remixer: Object-aware mixing layer for vision
  transformers and mixers}.{\BBCQ}
\newblock
 \APACrefbtitle {ICLR2022 Workshop on the Elements of Reasoning: Objects,
  Structure and Causality.} {Iclr2022 workshop on the elements of reasoning:
  Objects, structure and causality.}
\PrintBackRefs{\CurrentBib}

\bibitem [\protect \citeauthoryear {%
Krizhevsky%
, Hinton%
\BCBL {}\ \protect \BOthers {.}}{%
Krizhevsky%
\ \protect \BOthers {.}}{%
{\protect \APACyear {2009}}%
}]{%
cifar}
\APACinsertmetastar {%
cifar}%
\begin{APACrefauthors}%
Krizhevsky, A.%
, Hinton, G.%
\BCBL {}\ \BOthersPeriod {.}\end{APACrefauthors}%
\unskip\
\newblock
\APACrefYearMonthDay{2009}{}{}.
\newblock
{\BBOQ}\APACrefatitle {Learning multiple layers of features from tiny images}
  {Learning multiple layers of features from tiny images}.{\BBCQ}
\newblock

\newblock

\newblock

\PrintBackRefs{\CurrentBib}

\bibitem [\protect \citeauthoryear {%
K.~Li%
, Wu%
, Peng%
, Ernst%
\BCBL {}\ \BBA {} Fu%
}{%
K.~Li%
\ \protect \BOthers {.}}{%
{\protect \APACyear {2020}}%
}]{%
DBLP:journals/pami/LiWPEF20}
\APACinsertmetastar {%
DBLP:journals/pami/LiWPEF20}%
\begin{APACrefauthors}%
Li, K.%
, Wu, Z.%
, Peng, K.%
, Ernst, J.%
\BCBL {} Fu, Y.%
\end{APACrefauthors}%
\unskip\
\newblock
\APACrefYearMonthDay{2020}{}{}.
\newblock
{\BBOQ}\APACrefatitle {Guided Attention Inference Network} {Guided attention
  inference network}.{\BBCQ}
\newblock
\APACjournalVolNumPages{{IEEE} Trans. Pattern Anal. Mach.
  Intell.}{42}{12}{2996--3010}.
\newblock

\newblock

\PrintBackRefs{\CurrentBib}

\bibitem [\protect \citeauthoryear {%
Y.~Li%
, Mao%
, Girshick%
\BCBL {}\ \BBA {} He%
}{%
Y.~Li%
\ \protect \BOthers {.}}{%
{\protect \APACyear {2022}}%
}]{%
ViTDet}
\APACinsertmetastar {%
ViTDet}%
\begin{APACrefauthors}%
Li, Y.%
, Mao, H.%
, Girshick, R.%
\BCBL {} He, K.%
\end{APACrefauthors}%
\unskip\
\newblock
\APACrefYearMonthDay{2022}{}{}.
\newblock
{\BBOQ}\APACrefatitle {Exploring plain vision transformer backbones for object
  detection} {Exploring plain vision transformer backbones for object
  detection}.{\BBCQ}
\newblock
\APACjournalVolNumPages{arXiv:2203.16527}{}{}{}.
\newblock

\newblock

\PrintBackRefs{\CurrentBib}

\bibitem [\protect \citeauthoryear {%
Y.~Li%
, Wu%
\BCBL {}\ \protect \BOthers {.}}{%
Y.~Li%
, Wu%
\BCBL {}\ \protect \BOthers {.}}{%
{\protect \APACyear {2021}}%
}]{%
DBLP:journals/corr/abs-2112-01526}
\APACinsertmetastar {%
DBLP:journals/corr/abs-2112-01526}%
\begin{APACrefauthors}%
Li, Y.%
, Wu, C.%
, Fan, H.%
, Mangalam, K.%
, Xiong, B.%
, Malik, J.%
\BCBL {} Feichtenhofer, C.%
\end{APACrefauthors}%
\unskip\
\newblock
\APACrefYearMonthDay{2021}{}{}.
\newblock
{\BBOQ}\APACrefatitle {Improved Multiscale Vision Transformers for
  Classification and Detection} {Improved multiscale vision transformers for
  classification and detection}.{\BBCQ}
\newblock
\APACjournalVolNumPages{arXiv:2112.01526}{}{}{}.
\newblock

\newblock

\PrintBackRefs{\CurrentBib}

\bibitem [\protect \citeauthoryear {%
Y.~Li%
, Zhang%
, Cao%
, Timofte%
\BCBL {}\ \BBA {} Gool%
}{%
Y.~Li%
, Zhang%
\BCBL {}\ \protect \BOthers {.}}{%
{\protect \APACyear {2021}}%
}]{%
LocalViT}
\APACinsertmetastar {%
LocalViT}%
\begin{APACrefauthors}%
Li, Y.%
, Zhang, K.%
, Cao, J.%
, Timofte, R.%
\BCBL {} Gool, L.V.%
\end{APACrefauthors}%
\unskip\
\newblock
\APACrefYearMonthDay{2021}{}{}.
\newblock
{\BBOQ}\APACrefatitle {{LocalViT}: Bringing Locality to Vision Transformers}
  {{LocalViT}: Bringing locality to vision transformers}.{\BBCQ}
\newblock
\APACjournalVolNumPages{arXiv:2104.05707}{}{}{}.
\newblock

\newblock

\PrintBackRefs{\CurrentBib}

\bibitem [\protect \citeauthoryear {%
Y.~Liu%
\ \protect \BOthers {.}}{%
Y.~Liu%
\ \protect \BOthers {.}}{%
{\protect \APACyear {2021}}%
}]{%
OurNIPS}
\APACinsertmetastar {%
OurNIPS}%
\begin{APACrefauthors}%
Liu, Y.%
, Sangineto, E.%
, Bi, W.%
, Sebe, N.%
, Lepri, B.%
\BCBL {} Nadai, M.D.%
\end{APACrefauthors}%
\unskip\
\newblock
\APACrefYearMonthDay{2021}{}{}.
\newblock
{\BBOQ}\APACrefatitle {Efficient Training of Visual Transformers with Small
  Datasets} {Efficient training of visual transformers with small
  datasets}.{\BBCQ}
\newblock
\APACjournalVolNumPages{NeurIPS}{}{}{}.
\newblock

\newblock

\PrintBackRefs{\CurrentBib}

\bibitem [\protect \citeauthoryear {%
Z.~Liu%
\ \protect \BOthers {.}}{%
Z.~Liu%
\ \protect \BOthers {.}}{%
{\protect \APACyear {2021}}%
}]{%
Swin}
\APACinsertmetastar {%
Swin}%
\begin{APACrefauthors}%
Liu, Z.%
, Lin, Y.%
, Cao, Y.%
, Hu, H.%
, Wei, Y.%
, Zhang, Z.%
\BDBL {}Guo, B.%
\end{APACrefauthors}%
\unskip\
\newblock
\APACrefYearMonthDay{2021}{}{}.
\newblock
{\BBOQ}\APACrefatitle {Swin transformer: Hierarchical vision transformer using
  shifted windows} {Swin transformer: Hierarchical vision transformer using
  shifted windows}.{\BBCQ}
\newblock
\APACjournalVolNumPages{ICCV}{}{}{}.
\newblock

\newblock

\PrintBackRefs{\CurrentBib}

\bibitem [\protect \citeauthoryear {%
Locatello%
\ \protect \BOthers {.}}{%
Locatello%
\ \protect \BOthers {.}}{%
{\protect \APACyear {2020}}%
}]{%
SlotAttention}
\APACinsertmetastar {%
SlotAttention}%
\begin{APACrefauthors}%
Locatello, F.%
, Weissenborn, D.%
, Unterthiner, T.%
, Mahendran, A.%
, Heigold, G.%
, Uszkoreit, J.%
\BDBL {}Kipf, T.%
\end{APACrefauthors}%
\unskip\
\newblock
\APACrefYearMonthDay{2020}{}{}.
\newblock
{\BBOQ}\APACrefatitle {Object-centric learning with slot attention}
  {Object-centric learning with slot attention}.{\BBCQ}
\newblock
\APACjournalVolNumPages{NeurIPS}{33}{}{11525--11538}.
\newblock

\newblock

\PrintBackRefs{\CurrentBib}

\bibitem [\protect \citeauthoryear {%
Loshchilov%
\ \BBA {} Hutter%
}{%
Loshchilov%
\ \BBA {} Hutter%
}{%
{\protect \APACyear {2019}}%
}]{%
LoshchilovH19}
\APACinsertmetastar {%
LoshchilovH19}%
\begin{APACrefauthors}%
Loshchilov, I.%
\BCBT {}\ \BBA {} Hutter, F.%
\end{APACrefauthors}%
\unskip\
\newblock
\APACrefYearMonthDay{2019}{}{}.
\newblock
{\BBOQ}\APACrefatitle {Decoupled Weight Decay Regularization} {Decoupled weight
  decay regularization}.{\BBCQ}
\newblock
 \APACrefbtitle {7th International Conference on Learning Representations,
  {ICLR} 2019, New Orleans, LA, USA, May 6-9, 2019.} {7th international
  conference on learning representations, {ICLR} 2019, new orleans, la, usa,
  may 6-9, 2019.}
\PrintBackRefs{\CurrentBib}

\bibitem [\protect \citeauthoryear {%
Luo%
, Li%
, Urtasun%
\BCBL {}\ \BBA {} Zemel%
}{%
Luo%
\ \protect \BOthers {.}}{%
{\protect \APACyear {2017}}%
}]{%
DBLP:journals/corr/LuoLUZ17}
\APACinsertmetastar {%
DBLP:journals/corr/LuoLUZ17}%
\begin{APACrefauthors}%
Luo, W.%
, Li, Y.%
, Urtasun, R.%
\BCBL {} Zemel, R.S.%
\end{APACrefauthors}%
\unskip\
\newblock
\APACrefYearMonthDay{2017}{}{}.
\newblock
{\BBOQ}\APACrefatitle {Understanding the Effective Receptive Field in Deep
  Convolutional Neural Networks} {Understanding the effective receptive field
  in deep convolutional neural networks}.{\BBCQ}
\newblock
\APACjournalVolNumPages{arXiv:1701.04128}{}{}{}.
\newblock

\newblock

\PrintBackRefs{\CurrentBib}

\bibitem [\protect \citeauthoryear {%
Meinhardt%
, Kirillov%
, Leal{-}Taix{\'{e}}%
\BCBL {}\ \BBA {} Feichtenhofer%
}{%
Meinhardt%
\ \protect \BOthers {.}}{%
{\protect \APACyear {2022}}%
}]{%
TrackFormer}
\APACinsertmetastar {%
TrackFormer}%
\begin{APACrefauthors}%
Meinhardt, T.%
, Kirillov, A.%
, Leal{-}Taix{\'{e}}, L.%
\BCBL {} Feichtenhofer, C.%
\end{APACrefauthors}%
\unskip\
\newblock
\APACrefYearMonthDay{2022}{}{}.
\newblock
{\BBOQ}\APACrefatitle {{TrackFormer}: Multi-Object Tracking with Transformers}
  {{TrackFormer}: Multi-object tracking with transformers}.{\BBCQ}
\newblock
\APACjournalVolNumPages{CVPR}{}{}{}.
\newblock

\newblock

\PrintBackRefs{\CurrentBib}

\bibitem [\protect \citeauthoryear {%
Naseer%
\ \protect \BOthers {.}}{%
Naseer%
\ \protect \BOthers {.}}{%
{\protect \APACyear {2021}}%
}]{%
IntriguingProperties-of-VTs}
\APACinsertmetastar {%
IntriguingProperties-of-VTs}%
\begin{APACrefauthors}%
Naseer, M.%
, Ranasinghe, K.%
, Khan, S.H.%
, Hayat, M.%
, Khan, F.S.%
\BCBL {} Yang, M.%
\end{APACrefauthors}%
\unskip\
\newblock
\APACrefYearMonthDay{2021}{}{}.
\newblock
{\BBOQ}\APACrefatitle {Intriguing Properties of Vision Transformers}
  {Intriguing properties of vision transformers}.{\BBCQ}
\newblock
\APACjournalVolNumPages{NIPS}{}{}{}.
\newblock

\newblock

\PrintBackRefs{\CurrentBib}

\bibitem [\protect \citeauthoryear {%
Neimark%
, Bar%
, Zohar%
\BCBL {}\ \BBA {} Asselmann%
}{%
Neimark%
\ \protect \BOthers {.}}{%
{\protect \APACyear {2021}}%
}]{%
DBLP:journals/corr/abs-2102-00719}
\APACinsertmetastar {%
DBLP:journals/corr/abs-2102-00719}%
\begin{APACrefauthors}%
Neimark, D.%
, Bar, O.%
, Zohar, M.%
\BCBL {} Asselmann, D.%
\end{APACrefauthors}%
\unskip\
\newblock
\APACrefYearMonthDay{2021}{}{}.
\newblock
{\BBOQ}\APACrefatitle {Video Transformer Network} {Video transformer
  network}.{\BBCQ}
\newblock
\APACjournalVolNumPages{arXiv:2102.00719}{}{}{}.
\newblock

\newblock

\PrintBackRefs{\CurrentBib}

\bibitem [\protect \citeauthoryear {%
Nilsback%
\ \BBA {} Zisserman%
}{%
Nilsback%
\ \BBA {} Zisserman%
}{%
{\protect \APACyear {2008}}%
}]{%
flower}
\APACinsertmetastar {%
flower}%
\begin{APACrefauthors}%
Nilsback, M\BHBI E.%
\BCBT {}\ \BBA {} Zisserman, A.%
\end{APACrefauthors}%
\unskip\
\newblock
\APACrefYearMonthDay{2008}{}{}.
\newblock
{\BBOQ}\APACrefatitle {Automated flower classification over a large number of
  classes} {Automated flower classification over a large number of
  classes}.{\BBCQ}
\newblock
 \APACrefbtitle {Indian Conference on Computer Vision, Graphics \& Image
  Processing.} {Indian conference on computer vision, graphics \& image
  processing.}
\PrintBackRefs{\CurrentBib}

\bibitem [\protect \citeauthoryear {%
Parkhi%
, Vedaldi%
, Zisserman%
\BCBL {}\ \BBA {} Jawahar%
}{%
Parkhi%
\ \protect \BOthers {.}}{%
{\protect \APACyear {2012}}%
}]{%
Pets-dataset}
\APACinsertmetastar {%
Pets-dataset}%
\begin{APACrefauthors}%
Parkhi, O.M.%
, Vedaldi, A.%
, Zisserman, A.%
\BCBL {} Jawahar, C.V.%
\end{APACrefauthors}%
\unskip\
\newblock
\APACrefYearMonthDay{2012}{}{}.
\newblock
{\BBOQ}\APACrefatitle {Cats and dogs} {Cats and dogs}.{\BBCQ}
\newblock
 \APACrefbtitle {CVPR.} {Cvpr.}
\PrintBackRefs{\CurrentBib}

\bibitem [\protect \citeauthoryear {%
Radford%
\ \BBA {} Narasimhan%
}{%
Radford%
\ \BBA {} Narasimhan%
}{%
{\protect \APACyear {2018}}%
}]{%
Radford2018ImprovingLU}
\APACinsertmetastar {%
Radford2018ImprovingLU}%
\begin{APACrefauthors}%
Radford, A.%
\BCBT {}\ \BBA {} Narasimhan, K.%
\end{APACrefauthors}%
\unskip\
\newblock
\APACrefYearMonthDay{2018}{}{}.
\newblock
{\BBOQ}\APACrefatitle {Improving Language Understanding by Generative
  Pre-Training} {Improving language understanding by generative
  pre-training}.{\BBCQ}.
\PrintBackRefs{\CurrentBib}

\bibitem [\protect \citeauthoryear {%
Raghu%
, Unterthiner%
, Kornblith%
, Zhang%
\BCBL {}\ \BBA {} Dosovitskiy%
}{%
Raghu%
\ \protect \BOthers {.}}{%
{\protect \APACyear {2021}}%
}]{%
do-VTs-see-like-CNNs}
\APACinsertmetastar {%
do-VTs-see-like-CNNs}%
\begin{APACrefauthors}%
Raghu, M.%
, Unterthiner, T.%
, Kornblith, S.%
, Zhang, C.%
\BCBL {} Dosovitskiy, A.%
\end{APACrefauthors}%
\unskip\
\newblock
\APACrefYearMonthDay{2021}{}{}.
\newblock
{\BBOQ}\APACrefatitle {Do Vision Transformers See Like Convolutional Neural
  Networks?} {Do vision transformers see like convolutional neural
  networks?}{\BBCQ}
\newblock
\APACjournalVolNumPages{arXiv:2108.08810}{}{}{}.
\newblock

\newblock

\PrintBackRefs{\CurrentBib}

\bibitem [\protect \citeauthoryear {%
Ramesh%
, Dhariwal%
, Nichol%
, Chu%
\BCBL {}\ \BBA {} Chen%
}{%
Ramesh%
\ \protect \BOthers {.}}{%
{\protect \APACyear {2022}}%
}]{%
DALL-E-2}
\APACinsertmetastar {%
DALL-E-2}%
\begin{APACrefauthors}%
Ramesh, A.%
, Dhariwal, P.%
, Nichol, A.%
, Chu, C.%
\BCBL {} Chen, M.%
\end{APACrefauthors}%
\unskip\
\newblock
\APACrefYearMonthDay{2022}{}{}.
\newblock
{\BBOQ}\APACrefatitle {Hierarchical Text-Conditional Image Generation with
  {CLIP} Latents} {Hierarchical text-conditional image generation with {CLIP}
  latents}.{\BBCQ}
\newblock
\APACjournalVolNumPages{arXiv:2204.06125}{}{}{}.
\newblock

\newblock

\PrintBackRefs{\CurrentBib}

\bibitem [\protect \citeauthoryear {%
Ramesh%
\ \protect \BOthers {.}}{%
Ramesh%
\ \protect \BOthers {.}}{%
{\protect \APACyear {2021}}%
}]{%
DALL-E}
\APACinsertmetastar {%
DALL-E}%
\begin{APACrefauthors}%
Ramesh, A.%
, Pavlov, M.%
, Goh, G.%
, Gray, S.%
, Voss, C.%
, Radford, A.%
\BDBL {}Sutskever, I.%
\end{APACrefauthors}%
\unskip\
\newblock
\APACrefYearMonthDay{2021}{}{}.
\newblock
{\BBOQ}\APACrefatitle {Zero-Shot Text-to-Image Generation} {Zero-shot
  text-to-image generation}.{\BBCQ}
\newblock
 \APACrefbtitle {ICML.} {Icml.}
\PrintBackRefs{\CurrentBib}

\bibitem [\protect \citeauthoryear {%
Rao%
\ \protect \BOthers {.}}{%
Rao%
\ \protect \BOthers {.}}{%
{\protect \APACyear {2022}}%
}]{%
DenseCLIP}
\APACinsertmetastar {%
DenseCLIP}%
\begin{APACrefauthors}%
Rao, Y.%
, Zhao, W.%
, Chen, G.%
, Tang, Y.%
, Zhu, Z.%
, Huang, G.%
\BDBL {}Lu, J.%
\end{APACrefauthors}%
\unskip\
\newblock
\APACrefYearMonthDay{2022}{}{}.
\newblock
{\BBOQ}\APACrefatitle {{DenseCLIP}: Language-Guided Dense Prediction with
  Context-Aware Prompting} {{DenseCLIP}: Language-guided dense prediction with
  context-aware prompting}.{\BBCQ}
\newblock
\APACjournalVolNumPages{CVPR}{}{}{}.
\newblock

\newblock

\PrintBackRefs{\CurrentBib}

\bibitem [\protect \citeauthoryear {%
Razlighi%
\ \BBA {} Kehtarnavaz%
}{%
Razlighi%
\ \BBA {} Kehtarnavaz%
}{%
{\protect \APACyear {2009}}%
}]{%
Razlighi2009ACS}
\APACinsertmetastar {%
Razlighi2009ACS}%
\begin{APACrefauthors}%
Razlighi, Q.%
\BCBT {}\ \BBA {} Kehtarnavaz, N.%
\end{APACrefauthors}%
\unskip\
\newblock
\APACrefYearMonthDay{2009}{}{}.
\newblock
{\BBOQ}\APACrefatitle {A comparison study of image spatial entropy} {A
  comparison study of image spatial entropy}.{\BBCQ}
\newblock
 \APACrefbtitle {Electronic Imaging.} {Electronic imaging.}
\PrintBackRefs{\CurrentBib}

\bibitem [\protect \citeauthoryear {%
Rudin%
, Osher%
\BCBL {}\ \BBA {} Fatemi%
}{%
Rudin%
\ \protect \BOthers {.}}{%
{\protect \APACyear {1992}}%
}]{%
RUDIN1992259}
\APACinsertmetastar {%
RUDIN1992259}%
\begin{APACrefauthors}%
Rudin, L.I.%
, Osher, S.%
\BCBL {} Fatemi, E.%
\end{APACrefauthors}%
\unskip\
\newblock
\APACrefYearMonthDay{1992}{}{}.
\newblock
{\BBOQ}\APACrefatitle {Nonlinear total variation based noise removal
  algorithms} {Nonlinear total variation based noise removal
  algorithms}.{\BBCQ}
\newblock
\APACjournalVolNumPages{Physica D: Nonlinear Phenomena}{60}{1}{259-268}.
\newblock

\newblock

\PrintBackRefs{\CurrentBib}

\bibitem [\protect \citeauthoryear {%
Russakovsky%
\ \protect \BOthers {.}}{%
Russakovsky%
\ \protect \BOthers {.}}{%
{\protect \APACyear {2015}}%
}]{%
imagenet}
\APACinsertmetastar {%
imagenet}%
\begin{APACrefauthors}%
Russakovsky, O.%
, Deng, J.%
, Su, H.%
, Krause, J.%
, Satheesh, S.%
, Ma, S.%
\BDBL {}others%
\end{APACrefauthors}%
\unskip\
\newblock
\APACrefYearMonthDay{2015}{}{}.
\newblock
{\BBOQ}\APACrefatitle {Imagenet large scale visual recognition challenge}
  {Imagenet large scale visual recognition challenge}.{\BBCQ}
\newblock
\APACjournalVolNumPages{International Journal of Computer
  Vision}{115}{3}{211--252}.
\newblock

\newblock

\PrintBackRefs{\CurrentBib}

\bibitem [\protect \citeauthoryear {%
Sajjadi%
\ \protect \BOthers {.}}{%
Sajjadi%
\ \protect \BOthers {.}}{%
{\protect \APACyear {2022}}%
}]{%
ObjectSceneRepresentation}
\APACinsertmetastar {%
ObjectSceneRepresentation}%
\begin{APACrefauthors}%
Sajjadi, M.S.%
, Duckworth, D.%
, Mahendran, A.%
, van Steenkiste, S.%
, Paveti{\'c}, F.%
, Lu{\v{c}}i{\'c}, M.%
\BDBL {}Kipf, T.%
\end{APACrefauthors}%
\unskip\
\newblock
\APACrefYearMonthDay{2022}{}{}.
\newblock
{\BBOQ}\APACrefatitle {Object scene representation transformer} {Object scene
  representation transformer}.{\BBCQ}
\newblock
\APACjournalVolNumPages{arXiv:2206.06922}{}{}{}.
\newblock

\newblock

\PrintBackRefs{\CurrentBib}

\bibitem [\protect \citeauthoryear {%
Strudel%
, Garcia%
, Laptev%
\BCBL {}\ \BBA {} Schmid%
}{%
Strudel%
\ \protect \BOthers {.}}{%
{\protect \APACyear {2021}}%
}]{%
strudel2021segmenter}
\APACinsertmetastar {%
strudel2021segmenter}%
\begin{APACrefauthors}%
Strudel, R.%
, Garcia, R.%
, Laptev, I.%
\BCBL {} Schmid, C.%
\end{APACrefauthors}%
\unskip\
\newblock
\APACrefYearMonthDay{2021}{}{}.
\newblock
{\BBOQ}\APACrefatitle {Segmenter: Transformer for Semantic Segmentation}
  {Segmenter: Transformer for semantic segmentation}.{\BBCQ}
\newblock
 \APACrefbtitle {ICCV.} {Iccv.}
\PrintBackRefs{\CurrentBib}

\bibitem [\protect \citeauthoryear {%
Tian%
, Krishnan%
\BCBL {}\ \BBA {} Isola%
}{%
Tian%
\ \protect \BOthers {.}}{%
{\protect \APACyear {2020}}%
}]{%
tian2019contrastive}
\APACinsertmetastar {%
tian2019contrastive}%
\begin{APACrefauthors}%
Tian, Y.%
, Krishnan, D.%
\BCBL {} Isola, P.%
\end{APACrefauthors}%
\unskip\
\newblock
\APACrefYearMonthDay{2020}{}{}.
\newblock
{\BBOQ}\APACrefatitle {Contrastive Multiview Coding} {Contrastive multiview
  coding}.{\BBCQ}
\newblock
 \APACrefbtitle {ECCV.} {Eccv.}
\PrintBackRefs{\CurrentBib}

\bibitem [\protect \citeauthoryear {%
Touvron%
\ \protect \BOthers {.}}{%
Touvron%
\ \protect \BOthers {.}}{%
{\protect \APACyear {2020}}%
}]{%
DeiT}
\APACinsertmetastar {%
DeiT}%
\begin{APACrefauthors}%
Touvron, H.%
, Cord, M.%
, Douze, M.%
, Massa, F.%
, Sablayrolles, A.%
\BCBL {} J{\'e}gou, H.%
\end{APACrefauthors}%
\unskip\
\newblock
\APACrefYearMonthDay{2020}{}{}.
\newblock
{\BBOQ}\APACrefatitle {Training data-efficient image transformers \&
  distillation through attention} {Training data-efficient image transformers
  \& distillation through attention}.{\BBCQ}
\newblock
\APACjournalVolNumPages{arXiv:2012.12877}{}{}{}.
\newblock

\newblock

\PrintBackRefs{\CurrentBib}

\bibitem [\protect \citeauthoryear {%
Tupin%
, Sigelle%
\BCBL {}\ \BBA {} Maitre%
}{%
Tupin%
\ \protect \BOthers {.}}{%
{\protect \APACyear {2000}}%
}]{%
901061}
\APACinsertmetastar {%
901061}%
\begin{APACrefauthors}%
Tupin, F.%
, Sigelle, M.%
\BCBL {} Maitre, H.%
\end{APACrefauthors}%
\unskip\
\newblock
\APACrefYearMonthDay{2000}{}{}.
\newblock
{\BBOQ}\APACrefatitle {Definition of a spatial entropy and its use for texture
  discrimination} {Definition of a spatial entropy and its use for texture
  discrimination}.{\BBCQ}
\newblock
 \APACrefbtitle {ICIP.} {Icip.}
\PrintBackRefs{\CurrentBib}

\bibitem [\protect \citeauthoryear {%
van~den Oord%
, Li%
\BCBL {}\ \BBA {} Vinyals%
}{%
van~den Oord%
\ \protect \BOthers {.}}{%
{\protect \APACyear {2018}}%
}]{%
CPC}
\APACinsertmetastar {%
CPC}%
\begin{APACrefauthors}%
van~den Oord, A.%
, Li, Y.%
\BCBL {} Vinyals, O.%
\end{APACrefauthors}%
\unskip\
\newblock
\APACrefYearMonthDay{2018}{}{}.
\newblock
{\BBOQ}\APACrefatitle {Representation Learning with Contrastive Predictive
  Coding} {Representation learning with contrastive predictive coding}.{\BBCQ}
\newblock
\APACjournalVolNumPages{arXiv:1807.03748}{}{}{}.
\newblock

\newblock

\PrintBackRefs{\CurrentBib}

\bibitem [\protect \citeauthoryear {%
Vaswani%
\ \protect \BOthers {.}}{%
Vaswani%
\ \protect \BOthers {.}}{%
{\protect \APACyear {2017}}%
}]{%
attention-is-all-you-need}
\APACinsertmetastar {%
attention-is-all-you-need}%
\begin{APACrefauthors}%
Vaswani, A.%
, Shazeer, N.%
, Parmar, N.%
, Uszkoreit, J.%
, Jones, L.%
, Gomez, A.N.%
\BDBL {}Polosukhin, I.%
\end{APACrefauthors}%
\unskip\
\newblock
\APACrefYearMonthDay{2017}{}{}.
\newblock
{\BBOQ}\APACrefatitle {Attention is All you Need} {Attention is all you
  need}.{\BBCQ}
\newblock
 \APACrefbtitle {NeurIPS.} {Neurips.}
\PrintBackRefs{\CurrentBib}

\bibitem [\protect \citeauthoryear {%
T.~Wang%
\ \BBA {} Isola%
}{%
T.~Wang%
\ \BBA {} Isola%
}{%
{\protect \APACyear {2020}}%
}]{%
wang2020understanding}
\APACinsertmetastar {%
wang2020understanding}%
\begin{APACrefauthors}%
Wang, T.%
\BCBT {}\ \BBA {} Isola, P.%
\end{APACrefauthors}%
\unskip\
\newblock
\APACrefYearMonthDay{2020}{}{}.
\newblock
{\BBOQ}\APACrefatitle {Understanding contrastive representation learning
  through alignment and uniformity on the hypersphere} {Understanding
  contrastive representation learning through alignment and uniformity on the
  hypersphere}.{\BBCQ}
\newblock
 \APACrefbtitle {ICML.} {Icml.}
\PrintBackRefs{\CurrentBib}

\bibitem [\protect \citeauthoryear {%
W.~Wang%
\ \protect \BOthers {.}}{%
W.~Wang%
\ \protect \BOthers {.}}{%
{\protect \APACyear {2021}}%
}]{%
PVT}
\APACinsertmetastar {%
PVT}%
\begin{APACrefauthors}%
Wang, W.%
, Xie, E.%
, Li, X.%
, Fan, D\BHBI P.%
, Song, K.%
, Liang, D.%
\BDBL {}Shao, L.%
\end{APACrefauthors}%
\unskip\
\newblock
\APACrefYearMonthDay{2021}{}{}.
\newblock
{\BBOQ}\APACrefatitle {Pyramid Vision Transformer: A Versatile Backbone for
  Dense Prediction Without Convolutions} {Pyramid vision transformer: A
  versatile backbone for dense prediction without convolutions}.{\BBCQ}
\newblock
 \APACrefbtitle {ICCV.} {Iccv.}
\PrintBackRefs{\CurrentBib}

\bibitem [\protect \citeauthoryear {%
Z.~Wang%
\ \protect \BOthers {.}}{%
Z.~Wang%
\ \protect \BOthers {.}}{%
{\protect \APACyear {2022}}%
}]{%
DBLP:conf/iclr/WangYYDT022}
\APACinsertmetastar {%
DBLP:conf/iclr/WangYYDT022}%
\begin{APACrefauthors}%
Wang, Z.%
, Yu, J.%
, Yu, A.W.%
, Dai, Z.%
, Tsvetkov, Y.%
\BCBL {} Cao, Y.%
\end{APACrefauthors}%
\unskip\
\newblock
\APACrefYearMonthDay{2022}{}{}.
\newblock
{\BBOQ}\APACrefatitle {{SimVLM:} Simple Visual Language Model Pretraining with
  Weak Supervision} {{SimVLM:} simple visual language model pretraining with
  weak supervision}.{\BBCQ}
\newblock
 \APACrefbtitle {ICLR.} {Iclr.}
\PrintBackRefs{\CurrentBib}

\bibitem [\protect \citeauthoryear {%
Wei%
\ \protect \BOthers {.}}{%
Wei%
\ \protect \BOthers {.}}{%
{\protect \APACyear {2021}}%
}]{%
DBLP:journals/corr/abs-2112-09133}
\APACinsertmetastar {%
DBLP:journals/corr/abs-2112-09133}%
\begin{APACrefauthors}%
Wei, C.%
, Fan, H.%
, Xie, S.%
, Wu, C.%
, Yuille, A.L.%
\BCBL {} Feichtenhofer, C.%
\end{APACrefauthors}%
\unskip\
\newblock
\APACrefYearMonthDay{2021}{}{}.
\newblock
{\BBOQ}\APACrefatitle {Masked Feature Prediction for Self-Supervised Visual
  Pre-Training} {Masked feature prediction for self-supervised visual
  pre-training}.{\BBCQ}
\newblock
\APACjournalVolNumPages{arXiv:2112.09133}{}{}{}.
\newblock

\newblock

\PrintBackRefs{\CurrentBib}

\bibitem [\protect \citeauthoryear {%
Wu%
\ \protect \BOthers {.}}{%
Wu%
\ \protect \BOthers {.}}{%
{\protect \APACyear {2021}}%
}]{%
CvT}
\APACinsertmetastar {%
CvT}%
\begin{APACrefauthors}%
Wu, H.%
, Xiao, B.%
, Codella, N.%
, Liu, M.%
, Dai, X.%
, Yuan, L.%
\BCBL {} Zhang, L.%
\end{APACrefauthors}%
\unskip\
\newblock
\APACrefYearMonthDay{2021}{}{}.
\newblock
{\BBOQ}\APACrefatitle {{CvT}: Introducing convolutions to vision transformers}
  {{CvT}: Introducing convolutions to vision transformers}.{\BBCQ}
\newblock
\APACjournalVolNumPages{ICCV}{}{}{}.
\newblock

\newblock

\PrintBackRefs{\CurrentBib}

\bibitem [\protect \citeauthoryear {%
Xie%
\ \protect \BOthers {.}}{%
Xie%
\ \protect \BOthers {.}}{%
{\protect \APACyear {2022}}%
}]{%
DBLP:journals/corr/abs-2111-09886}
\APACinsertmetastar {%
DBLP:journals/corr/abs-2111-09886}%
\begin{APACrefauthors}%
Xie, Z.%
, Zhang, Z.%
, Cao, Y.%
, Lin, Y.%
, Bao, J.%
, Yao, Z.%
\BDBL {}Hu, H.%
\end{APACrefauthors}%
\unskip\
\newblock
\APACrefYearMonthDay{2022}{}{}.
\newblock
{\BBOQ}\APACrefatitle {{SimMIM: A} Simple Framework for Masked Image Modeling}
  {{SimMIM: A} simple framework for masked image modeling}.{\BBCQ}
\newblock
\APACjournalVolNumPages{CVPR}{}{}{}.
\newblock

\newblock

\PrintBackRefs{\CurrentBib}

\bibitem [\protect \citeauthoryear {%
Xu%
, Xu%
, Chang%
\BCBL {}\ \BBA {} Tu%
}{%
Xu%
\ \protect \BOthers {.}}{%
{\protect \APACyear {2021}}%
}]{%
CoaT}
\APACinsertmetastar {%
CoaT}%
\begin{APACrefauthors}%
Xu, W.%
, Xu, Y.%
, Chang, T.%
\BCBL {} Tu, Z.%
\end{APACrefauthors}%
\unskip\
\newblock
\APACrefYearMonthDay{2021}{}{}.
\newblock
{\BBOQ}\APACrefatitle {Co-Scale Conv-Attentional Image Transformers} {Co-scale
  conv-attentional image transformers}.{\BBCQ}
\newblock
\APACjournalVolNumPages{ICCV}{}{}{}.
\newblock

\newblock

\PrintBackRefs{\CurrentBib}

\bibitem [\protect \citeauthoryear {%
K.~Yuan%
\ \protect \BOthers {.}}{%
K.~Yuan%
\ \protect \BOthers {.}}{%
{\protect \APACyear {2021}}%
}]{%
CeiT}
\APACinsertmetastar {%
CeiT}%
\begin{APACrefauthors}%
Yuan, K.%
, Guo, S.%
, Liu, Z.%
, Zhou, A.%
, Yu, F.%
\BCBL {} Wu, W.%
\end{APACrefauthors}%
\unskip\
\newblock
\APACrefYearMonthDay{2021}{}{}.
\newblock
{\BBOQ}\APACrefatitle {Incorporating Convolution Designs into Visual
  Transformers} {Incorporating convolution designs into visual
  transformers}.{\BBCQ}
\newblock
\APACjournalVolNumPages{ICCV}{}{}{}.
\newblock

\newblock

\PrintBackRefs{\CurrentBib}

\bibitem [\protect \citeauthoryear {%
L.~Yuan%
\ \protect \BOthers {.}}{%
L.~Yuan%
\ \protect \BOthers {.}}{%
{\protect \APACyear {2021}}%
}]{%
T2T}
\APACinsertmetastar {%
T2T}%
\begin{APACrefauthors}%
Yuan, L.%
, Chen, Y.%
, Wang, T.%
, Yu, W.%
, Shi, Y.%
, Jiang, Z.%
\BDBL {}Yan, S.%
\end{APACrefauthors}%
\unskip\
\newblock
\APACrefYearMonthDay{2021}{}{}.
\newblock
{\BBOQ}\APACrefatitle {Tokens-to-token {ViT}: Training vision transformers from
  scratch on {ImageNet}} {Tokens-to-token {ViT}: Training vision transformers
  from scratch on {ImageNet}}.{\BBCQ}
\newblock
 \APACrefbtitle {ICCV.} {Iccv.}
\PrintBackRefs{\CurrentBib}

\bibitem [\protect \citeauthoryear {%
Yun%
\ \protect \BOthers {.}}{%
Yun%
\ \protect \BOthers {.}}{%
{\protect \APACyear {2019}}%
}]{%
CutMix}
\APACinsertmetastar {%
CutMix}%
\begin{APACrefauthors}%
Yun, S.%
, Han, D.%
, Chun, S.%
, Oh, S.J.%
, Yoo, Y.%
\BCBL {} Choe, J.%
\end{APACrefauthors}%
\unskip\
\newblock
\APACrefYearMonthDay{2019}{}{}.
\newblock
{\BBOQ}\APACrefatitle {{CutMix: R}egularization Strategy to Train Strong
  Classifiers With Localizable Features} {{CutMix: R}egularization strategy to
  train strong classifiers with localizable features}.{\BBCQ}
\newblock
 \APACrefbtitle {ICCV.} {Iccv.}
\PrintBackRefs{\CurrentBib}

\bibitem [\protect \citeauthoryear {%
Yun%
, Lee%
, Kim%
\BCBL {}\ \BBA {} Shin%
}{%
Yun%
\ \protect \BOthers {.}}{%
{\protect \APACyear {2022}}%
}]{%
PatchLevelRepresentation}
\APACinsertmetastar {%
PatchLevelRepresentation}%
\begin{APACrefauthors}%
Yun, S.%
, Lee, H.%
, Kim, J.%
\BCBL {} Shin, J.%
\end{APACrefauthors}%
\unskip\
\newblock
\APACrefYearMonthDay{2022}{}{}.
\newblock
{\BBOQ}\APACrefatitle {Patch-Level Representation Learning for Self-Supervised
  Vision Transformers} {Patch-level representation learning for self-supervised
  vision transformers}.{\BBCQ}
\newblock
 \APACrefbtitle {CVPR} {Cvpr}\ (\BPGS\ 8354--8363).
\PrintBackRefs{\CurrentBib}

\bibitem [\protect \citeauthoryear {%
Zbontar%
, Jing%
, Misra%
, LeCun%
\BCBL {}\ \BBA {} Deny%
}{%
Zbontar%
\ \protect \BOthers {.}}{%
{\protect \APACyear {2021}}%
}]{%
barlow}
\APACinsertmetastar {%
barlow}%
\begin{APACrefauthors}%
Zbontar, J.%
, Jing, L.%
, Misra, I.%
, LeCun, Y.%
\BCBL {} Deny, S.%
\end{APACrefauthors}%
\unskip\
\newblock
\APACrefYearMonthDay{2021}{}{}.
\newblock
{\BBOQ}\APACrefatitle {Barlow Twins: Self-Supervised Learning via Redundancy
  Reduction} {Barlow twins: Self-supervised learning via redundancy
  reduction}.{\BBCQ}
\newblock
 \APACrefbtitle {ICML.} {Icml.}
\PrintBackRefs{\CurrentBib}

\bibitem [\protect \citeauthoryear {%
Zhang%
, Ciss{\'{e}}%
, Dauphin%
\BCBL {}\ \BBA {} Lopez{-}Paz%
}{%
Zhang%
\ \protect \BOthers {.}}{%
{\protect \APACyear {2018}}%
}]{%
mixup}
\APACinsertmetastar {%
mixup}%
\begin{APACrefauthors}%
Zhang, H.%
, Ciss{\'{e}}, M.%
, Dauphin, Y.N.%
\BCBL {} Lopez{-}Paz, D.%
\end{APACrefauthors}%
\unskip\
\newblock
\APACrefYearMonthDay{2018}{}{}.
\newblock
{\BBOQ}\APACrefatitle {mixup: Beyond Empirical Risk Minimization} {mixup:
  Beyond empirical risk minimization}.{\BBCQ}
\newblock
 \APACrefbtitle {ICLR.} {Iclr.}
\PrintBackRefs{\CurrentBib}

\bibitem [\protect \citeauthoryear {%
Zhao%
, Jiang%
, Jia%
, Torr%
\BCBL {}\ \BBA {} Koltun%
}{%
Zhao%
\ \protect \BOthers {.}}{%
{\protect \APACyear {2021}}%
}]{%
PointTransformer}
\APACinsertmetastar {%
PointTransformer}%
\begin{APACrefauthors}%
Zhao, H.%
, Jiang, L.%
, Jia, J.%
, Torr, P.H.S.%
\BCBL {} Koltun, V.%
\end{APACrefauthors}%
\unskip\
\newblock
\APACrefYearMonthDay{2021}{}{}.
\newblock
{\BBOQ}\APACrefatitle {Point Transformer} {Point transformer}.{\BBCQ}
\newblock
\APACjournalVolNumPages{ICCV}{}{}{}.
\newblock

\newblock

\PrintBackRefs{\CurrentBib}

\bibitem [\protect \citeauthoryear {%
Zheng%
\ \protect \BOthers {.}}{%
Zheng%
\ \protect \BOthers {.}}{%
{\protect \APACyear {2021}}%
}]{%
zheng20213d}
\APACinsertmetastar {%
zheng20213d}%
\begin{APACrefauthors}%
Zheng, C.%
, Zhu, S.%
, Mendieta, M.%
, Yang, T.%
, Chen, C.%
\BCBL {} Ding, Z.%
\end{APACrefauthors}%
\unskip\
\newblock
\APACrefYearMonthDay{2021}{}{}.
\newblock
{\BBOQ}\APACrefatitle {{3D H}uman Pose Estimation with Spatial and Temporal
  Transformers} {{3D H}uman pose estimation with spatial and temporal
  transformers}.{\BBCQ}
\newblock
\APACjournalVolNumPages{ICCV}{}{}{}.
\newblock

\newblock

\PrintBackRefs{\CurrentBib}

\bibitem [\protect \citeauthoryear {%
Zhou%
\ \protect \BOthers {.}}{%
Zhou%
\ \protect \BOthers {.}}{%
{\protect \APACyear {2022}}%
}]{%
DBLP:journals/corr/abs-2111-07832}
\APACinsertmetastar {%
DBLP:journals/corr/abs-2111-07832}%
\begin{APACrefauthors}%
Zhou, J.%
, Wei, C.%
, Wang, H.%
, Shen, W.%
, Xie, C.%
, Yuille, A.L.%
\BCBL {} Kong, T.%
\end{APACrefauthors}%
\unskip\
\newblock
\APACrefYearMonthDay{2022}{}{}.
\newblock
{\BBOQ}\APACrefatitle {{iBOT: Image {BERT} Pre-Training with Online Tokenizer}}
  {{iBOT: Image {BERT} Pre-Training with Online Tokenizer}}.{\BBCQ}
\newblock
\APACjournalVolNumPages{ICLR}{}{}{}.
\newblock

\newblock

\PrintBackRefs{\CurrentBib}

\bibitem [\protect \citeauthoryear {%
Zhu%
\ \protect \BOthers {.}}{%
Zhu%
\ \protect \BOthers {.}}{%
{\protect \APACyear {2021}}%
}]{%
Deformable-DETR}
\APACinsertmetastar {%
Deformable-DETR}%
\begin{APACrefauthors}%
Zhu, X.%
, Su, W.%
, Lu, L.%
, Li, B.%
, Wang, X.%
\BCBL {} Dai, J.%
\end{APACrefauthors}%
\unskip\
\newblock
\APACrefYearMonthDay{2021}{}{}.
\newblock
{\BBOQ}\APACrefatitle {Deformable {DETR}: Deformable Transformers for
  End-to-End Object Detection} {Deformable {DETR}: Deformable transformers for
  end-to-end object detection}.{\BBCQ}
\newblock
 \APACrefbtitle {ICLR.} {Iclr.}
\PrintBackRefs{\CurrentBib}

\bibitem [\protect \citeauthoryear {%
Zhuang%
, Zhai%
\BCBL {}\ \BBA {} Yamins%
}{%
Zhuang%
\ \protect \BOthers {.}}{%
{\protect \APACyear {2019}}%
}]{%
DBLP:journals/corr/abs-1903-12355}
\APACinsertmetastar {%
DBLP:journals/corr/abs-1903-12355}%
\begin{APACrefauthors}%
Zhuang, C.%
, Zhai, A.L.%
\BCBL {} Yamins, D.%
\end{APACrefauthors}%
\unskip\
\newblock
\APACrefYearMonthDay{2019}{}{}.
\newblock
{\BBOQ}\APACrefatitle {Local Aggregation for Unsupervised Learning of Visual
  Embeddings} {Local aggregation for unsupervised learning of visual
  embeddings}.{\BBCQ}
\newblock
 \APACrefbtitle {ICCV.} {Iccv.}
\PrintBackRefs{\CurrentBib}

\end{thebibliography}


\backmatter
\appendix
\renewcommand{\thesection}{\Alph{section}}

\section{Pseudo-code of the Spatial Entropy loss}
\label{sec:Pseudo}

\cref{fig:pytorch-code} shows the pseudo-code for computing $\mathcal{L}_{se}$  (Eq.~\ref{eq:spatial-entropy-loss}). 
The goal is twofold: to show how easy is to compute $\mathcal{L}_{se}$, and  how to make it differentiable. Specifically, 
Eq.~\ref{eq:connected-components} 
 is based on a connected component algorithm which is not differentiable. However, once $C^h$ is computed, each element $C_j \in C^h$ 
($C_j = \{ (x_1, y_1), ..., (x_{n_j}, y_{n_j})\}$) 
can be represented as a binary mask $M_j$, defined as:

\begin{equation}
\label{eq:binary-mask}
       M_j [x,y] = 
\begin{cases}
     1 ,            & \text{if } (x,y) \in C_j \\
    0              & \text{otherwise}
\end{cases}
\end{equation}

Using Eq.~\ref{eq:binary-mask}, we  compute the probability $P^h(C_j)$ of the component $C_j$ using matrix multiplication, which is done  efficiently on GPUs. In practice, Eq.~\ref{eq:single-component-probability} 
 is computed as:

\begin{equation}
    P^h (C_j) =  \frac{1}{\vert B^h \vert}  \sum_{(x, y)} B_{x,y}^h \odot M_j[x, y], 
\end{equation} 
\noindent
where $\odot$ is the element-wise product. This implementation makes it is possible to back-propagate the spatial entropy loss even if the connected component algorithm is not differentiable.

\begin{figure*}[!ht]
    \centering 
    \begin{minipage}{0.85\textwidth}

    \begin{minted}[
fontsize=\fontsize{8.5}{9},
frame=single,
framesep=2.5pt,
baselinestretch=1.05,
]{python}
# bs    : batch size
# H     : number of heads
# k x k : size of the embedding grid
# S     : defined in Eq. 4 (main paper), 
#         tensor with shape [bs, H, k, k]

def SpatialEntropy(S, eps=1e-9):
    bs, H, k, _ = S.size()
    m = torch.mean(S, dim=[2, 3], keepdim=True)
    B = torch.relu(S - m) + eps   
    
    with torch.no_grad():
        batch_idxs, head_idxs, M = connected_components(B)
    
    p_u = torch.sum(B[batch_idxs, head_idxs] * M, dim=[2, 3]) 
    p_n = p_u / torch.sum(B, dim=[2, 3])                     
    SE = -torch.sum((p_n * torch.log(p_n)) / (bs * H)
    return SE

\end{minted}
    \caption{PyTorch-like pseudocode of the Spatial Entropy loss.
     \label{fig:pytorch-code}}
    \end{minipage}
\end{figure*}

\section{Comparing the spatial entropy loss with other solutions}
\label{sec:total-variation}

In our preliminary experiments, we replaced the spatial entropy loss with a different loss based on a {\em total variation denoising}  \citep{RUDIN1992259} 
criterion, which we formulated as:
\begin{multline}
\label{eq:total-variation-loss}
\mathcal{L}_{tv} = \frac{1}{H}
\sum_{h=1}^H \sum_{(x,y) \in \{1,...,k\}^2} 
\vert S_{x,y}^h - S_{x,y+1}^h \vert + \\
\vert S_{x,y}^h - S_{x+1,y}^h \vert .
\end{multline}

\noindent
However, $\mathcal{L}_{tv}$ drastically underperforms $\mathcal{L}_{se}$ and tends to produce blurred (more uniform) attention maps.
The main difference between $\mathcal{L}_{tv}$ and $\mathcal{L}_{se}$ is that the thresholding (Eq.~\ref{eq:binary-map}) 
and the adjacency-based clustering (Eq.~\ref{eq:connected-components}) 
operations in $\mathcal{L}_{se}$ 
 group together
image regions of variable size and shape which have in common high attention scores 
for a specific head, and, therefore, they presumably represent the same semantics (e.g., a specific object). In contrast, $\mathcal{L}_{tv}$ compares 
patch tokens which are adjacent to each other (e.g., $S_{x,y}^h$ and $S_{x,y+1}^h$) but which do not necessarily share the same semantics
(e.g., $S_{x,y}^h$ may belong  to the background while  $S_{x,y+1}^h$ belongs to a foreground object). 
Thus, the implicit local inductive bias of the two losses is different: in case of $\mathcal{L}_{se}$, the inductive bias is that 
the image regions corresponding to the highest attention scores (for a specific head) 
should be spatially grouped in few, big ``blobs'', while, in the case of $\mathcal{L}_{tv}$, the inductive bias is that generic adjacent regions
should have  similar attention scores.

Finally, we have empirically tested  slightly different similarity metrics. For instance, replacing Eq.~\ref{eq:similarity-map} with a cosine similarity computed as:

\begin{equation}
\label{eq:similarity-map-normalized}
\hat{S}_{x,y}^h = \frac{<\pmb{q}_{CLS}^h, \pmb{k}_{x,y}^h>}{\vert\vert \pmb{q}_{CLS}^h \vert\vert \vert\vert\pmb{k}_{x,y}^h\vert\vert}, \quad (x,y) \in \{1,...,k\}^2,
\end{equation}

\noindent
and keeping all the rest unchanged (e.g., Eq.~\ref{eq:binary-map}-Eq.~\ref{eq:spatial-entropy-loss}),
we get a slightly lower  accuracy  when training on IN-100 (76.25 top-1 accuracy versus 76.72 in \cref{tab:ablation} (c)).
This is likely due to the fact that  $\hat{S}^h$ in Eq.~\ref{eq:similarity-map-normalized} does not correspond to the metric used
to compute the attention for the main task loss (Eq.~\ref{eq:attention-standard}). Thus, merging the main task loss (e.g., the cross entropy loss) with the spatial entropy loss may be more difficult.

\section{Self-supervised experiments}
\label{sec:appendix_ssl}

In this section, we 
 use SAR in a fully self-supervised scenario.
 Analogously to all the other experiments of this paper, we keep fixed 
 $\lambda = 0.01$ and we use the  hyperparameters of each tested VT-based method  without any hyperparameter tuning.
  Since self-supervised learning algorithms are very time consuming, we use IN-100, which is a medium-size dataset.
 We plug SAR on top of two state-of-the-art VT-based self-supervised learning algorithms:
 MoCo-v3 \citep{MoCo-v3} and DINO \citep{DINO} (\cref{sec:RelatedWork}).
 When we use MoCo-v3, in $\mathcal{L}_{tot}$ (\cref{sec:Spatial-entropy}),
we replace the cross-entropy loss ($\mathcal{L}_{ce}$) with the contrastive loss used in \citep{MoCo-v3}. Similarly, when we use DINO,
we use as the main task loss the ``self-distillation" proposed in \citep{DINO}, jointly with  its multi-crop strategy.
We use the official code of MoCo-v3 and DINO and we strictly follow the algorithms and the training protocols of the baseline methods, including all the default hyperparameters suggested by the corresponding authors. 
However, for computational reasons,   we used  a 1024 batch size for MoCo-v3 and MoCo-v3 + SAR, and
a batch size of 512
 for DINO and DINO + SAR.
The VT backbone is  ViT-S/16 for all the methods. 
More details in \cref{sec:Implementation}.
 
 We evaluate all  models (with and without SAR)
using the  
  standard self-supervised evaluation protocol, consisting  in freezing the network after training and then  training a linear classifier on top of the frozen features
\citep{DINO,MoCo-v3}. 
 The results are reported in \cref{tab:MoCov3DINO} (a), and show that, on IN-100, SAR significantly improves these state-of-the-art algorithms, including DINO (which inspired our work).

\begin{figure*}[p]
\centering
\begin{tabular}{C{0.49\linewidth}C{0.49\linewidth}}
    \multicolumn{2}{c}{\includegraphics[width=0.98\linewidth]{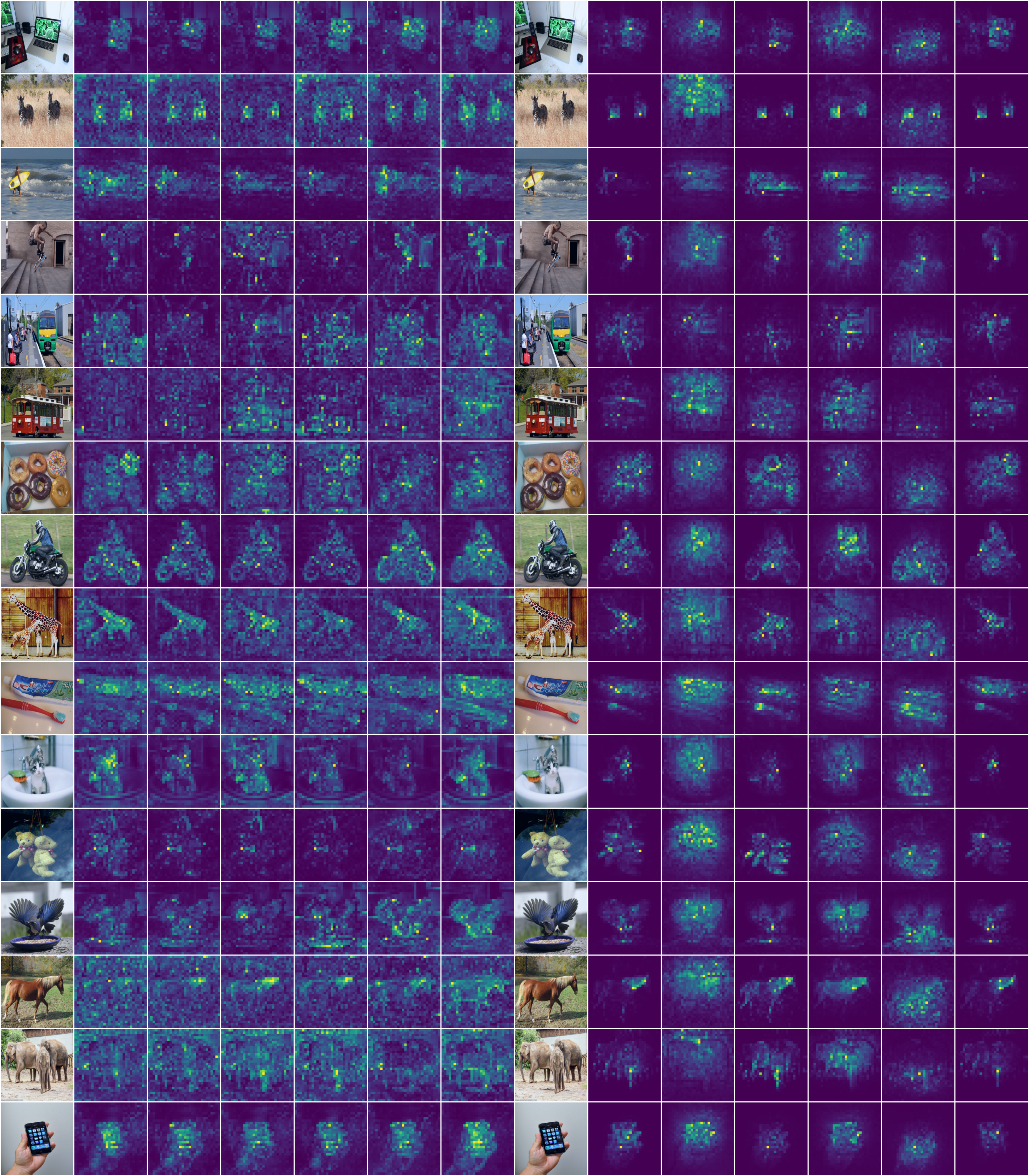}}\\
    MoCo-v3 & MoCo-v3 + SAR
\end{tabular}%
\caption{A qualitative comparison between the attention maps generated by MoCo-v3 and MoCo-v3 + SAR with the ViT-S/16 backbone (training on IN-100).
For each image, we show all the 6 attention maps ($A^h$) corresponding to the 6 last-block heads, computed using only the \texttt{[CLS]} token query.}
\label{fig:appendix-attention-maps-moco}
\end{figure*}

\begin{figure*}[p]
\centering
\begin{tabular}{C{0.49\linewidth}C{0.49\linewidth}}
    \multicolumn{2}{c}{\includegraphics[width=0.98\linewidth]{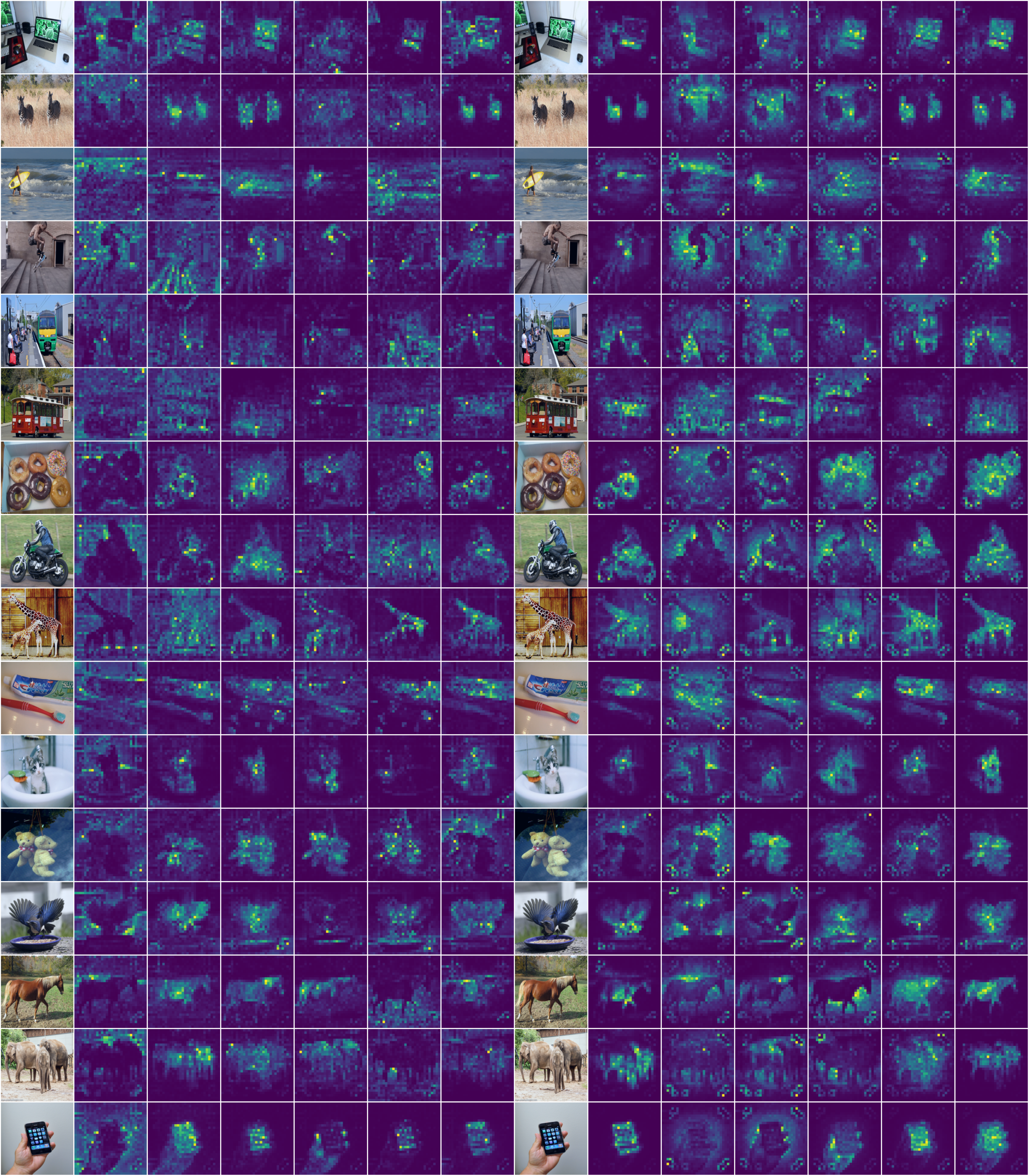}}\\
    DINO & DINO + SAR
\end{tabular}%
\caption{A qualitative comparison between the attention maps generated by DINO and DINO + SAR with the ViT-S/16 backbone (training on IN-100).}
\label{fig:appendix-attention-maps-dino}
\end{figure*}

We qualitatively compare the attention maps obtained with and without SAR in
\cref{fig:appendix-attention-maps-moco} (MoCo-v3) and \cref{fig:appendix-attention-maps-dino} (DINO).  \cref{fig:appendix-attention-maps-moco} shows that, in MoCo-v3 + SAR, the head-specific attention maps  focus on slightly different aspects of the main object, while in MoCo-v3, the attention is much more ``disordered'' (spread over  the whole image).
On the other hand, when comparing DINO with DINO + SAR (\cref{fig:appendix-attention-maps-dino}),
the attention map differences are more subtle. 
However, the higher inter-head variability in DINO + SAR is one interesting difference. For instance, while DINO's maps usually focus only on the main foreground object, in DINO + SAR, different heads cover different foreground objects (e.g., the cat and the sink in Row 11) or different background regions (e.g., the road and the sky in the ``train'' figure of Row 5).
This difference is probably due to the difference in how DINO and DINO + SAR are optimized. In fact, 
in DINO,
the only source of supervision is the comparison between two different views of the same image (\cref{sec:RelatedWork}), which likely
encourages the network to focus on the objects most frequently in common. On the other hand, in DINO + SAR, the creation of connected regions with large  attention values into each head's map is also encouraged  using $\mathcal{L}_{se}$. 

These qualitative observations are confirmed by the quantitative analysis reported  in \cref{tab:MoCov3DINO} (b), where we follow the protocol described in \cref{sec:Attention-map-analysis}.  SAR increases the Jaccard similarity of both self-supervised algorithms.

    
  

\begin{table}[t]
\vspace{2mm}
\setlength{\tabcolsep}{1.0pt}
\small
    \caption{Self-supervised training  on IN-100 (300  epochs). (a) Accuracy on IN-100 evaluated through linear probing and a single testing crop. (b) Segmentation properties of the corresponding attention maps evaluated on PASCAL VOC-12. The results of the baselines have been obtained by us using the corresponding publicly available code.}
    \begin{minipage}[b]{.48\textwidth}
    \captionsetup{labelformat=empty} 
    \label{tab:MoCov3DINO}
    \centering
    \begin{tabularx}{\linewidth}{Xl}
        \toprule
        \textbf{Method} & \textbf{Top-1 Acc.}\\ 
        \midrule
        MoCo-v3  & 77.60 \\ 
        \rowcolor{Light}MoCo-v3 + SAR & \textbf{78.88} \posimprov{1.28} \\ 
        \midrule
        DINO  & 77.42 \\ 
        \rowcolor{Light}DINO + SAR & \textbf{79.90} \posimprov{2.48}\\ 
        \bottomrule
    \end{tabularx}
    \vspace{0.5pt}
    \caption{(a)}
    \end{minipage}
    \hfill
    \begin{minipage}[b]{.49\textwidth}
    \captionsetup{labelformat=empty} 
    \centering
        \begin{tabularx}{\linewidth}{Xl}
            \toprule
            \textbf{Method} & \textbf{Jaccard similarity}  \\
            \midrule
            MoCo-v3 & 27.40 \\
            \rowcolor{Light}MoCo-v3 + SAR & \textbf{33.10} \posimprov{6.17}\\
            \midrule
            DINO  &  35.06 \\
            \rowcolor{Light}DINO + SAR & \textbf{35.28} \posimprov{0.22}\\ 
            \bottomrule
        \end{tabularx}
    \vspace{0.5pt}
    \caption{(b)}
    \end{minipage}
\end{table}


\section{An additional qualitative analysis of the attention maps}
\label{sec:Attention-map-analysis-new}

In this section, we extend the analysis of \cref{sec:Attention-map-analysis} providing additional visualizations of the attention maps obtained using supervised training on IN-1K.

\begin{figure*}[p]
\centering
\begin{tabular}{C{0.49\linewidth}C{0.49\linewidth}}
    \multicolumn{2}{c}{\includegraphics[width=0.98\linewidth]{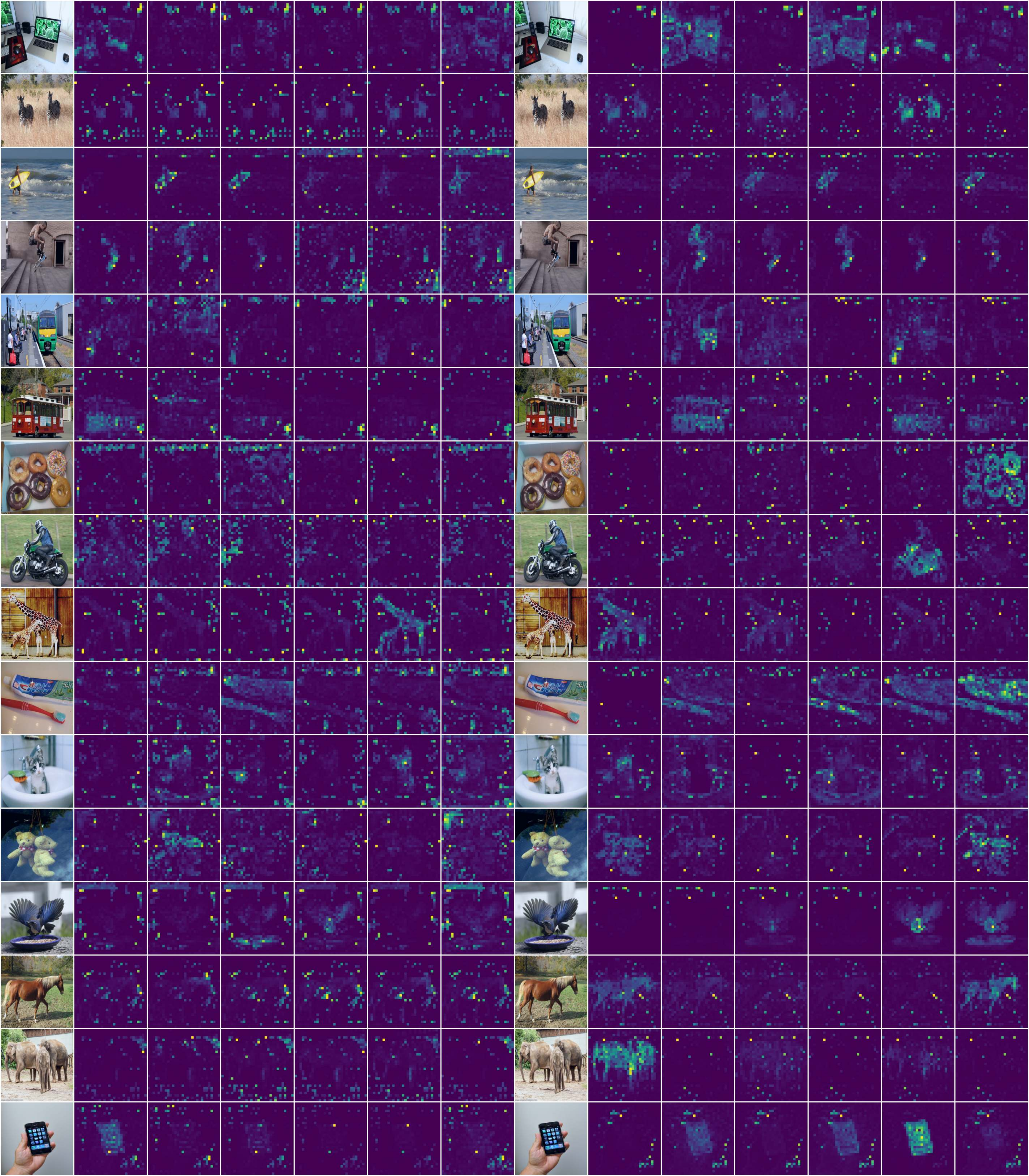}}\\
    ViT-S/16 & ViT-S/16 + SAR
\end{tabular}%
\caption{A qualitative comparison between the attention maps generated by ViT-S/16 and ViT-S/16 + SAR (supervised case, training on IN-1K).
}

\label{fig:appendix-attention-maps-supervised}
\end{figure*}

\cref{fig:appendix-attention-maps-supervised} shows the attention maps of ViT-S/16 and ViT-S/16 + SAR. These attention maps show that the ViT-S/16 attention scores are spread over all the image, while in ViT-S/16 + SAR they are much more spatially clustered and usually focused on the main object(s) of the input image. For example, the first row shows that the heads of ViT-S/16 focus on the upper part of the image, and only the keyboard of the laptop emerges. Vice versa, from the ViT-S/16 + SAR attention heads, it is possible to recognise the shape and size of the laptop precisely. 
Similarly, the second last row shows an example where the input image contains some elephants. While the different heads of ViT-S/16 seem to focus mainly on the background, the first head of ViT-S/16 + SAR is clearly focused on the elephants.  
Importantly,
these visualizations show that
different heads of ViT-S/16 + SAR usually focus on different semantic concepts, which shows that there are no collapse phenomena using the spatial entropy loss (see Eq.~\ref{eq:spatial-entropy} and the corresponding discussion in \cref{sec:Discussion}).

\section{Implementation details}
\label{sec:Implementation}
In the following, we list the implementation details. Each model was trained using 8  NVIDIA V100 32GB GPUs.
For computational reasons, the experiments of this paper have been done only once (and without hyperparameter tuning), thus we cannot report standard deviation values computed over multiple runs. However, this is the standard practice adopted by all the evaluation protocols used in this paper and by all the methods we have compared with.

\subsection{ViT based models}
We train our models using the public code of \citet{DeiT}\footnote{\url{https://github.com/facebookresearch/deit}} for ViT and we modify the original code when we use SAR, as described in \cref{sec:skip-connections}. The models are trained with a batch size of 1024, using the AdamW optimizer \citep{LoshchilovH19}  with initial learning rate of 0.001, a cosine learning rate schedule, a weight decay of 0.05 and using the original data-augmentation protocol, including the use of Mixup \citep{mixup} and CutMix \citep{CutMix} for the supervised classification tasks.

\subsection{Hybrid architectures}

In all the supervised experiments, we used the officially released code for PVT \citep{PVT}\footnote{\url{https://github.com/whai362/PVT/tree/v1}}, T2T \citep{T2T}\footnote{\url{https://github.com/yitu-opensource/T2T-ViT}} and CvT \citep{CvT}\footnote{\url{https://github.com/microsoft/CvT}}, strictly following the original training protocol for each architecture. 

PVT  and T2T  are trained with batch size of 1024, using the AdamW optimizer with an initial learning rate of 0.001, momentum 0.9 and weight decay of 0.05. 
CvT  is trained with a  batch size of 2048 and an initial learning rate of 0.02, decayed with a cosine schedule. The data augumentations of the original articles are based on the  DeiT protocol \citep{DeiT}. 
We refer the reader to the original papers for further details. When we use SAR, we  plug it on top of the original public code, following \cref{sec:skip-connections}.

\subsection{Object detection and semantic segmentation downstream tasks}
Starting from the pretrained ViT-S/16 and ViT-S/16 + SAR models of \cref{tab:ourneuipsvsthis}, we follow (\cite{ViTDet}) to obtain a feature pyramid and a Mask-RCNN detector \citep{he2017mask}.
In both the object detection and the segmentation task, the models are fine-tuned for 25 epochs with the AdamW optimizer, learning rate of 0.0001 and linear warmup for the first 250 iterations. The input image is 512$\times$512 during training, augmented with random cropping and flipping. We used a batch size of 8 distributed on 4 GPUs. 

\subsection{Transfer learning with different fine-tuning
protocols}

We fine-tune the ViT-S/16 models pretrained on IN-1K (see \cref{tab:ourneuipsvsthis}), always keeping unchanged the VT architecture used in the pre-training stage.
This is done to avoid making  the adaptation task more difficult, since each of the four datasets used in \cref{tab:fine-tuning-new} is composed of a relatively small number of samples.
For instance, when SAR is used during pre-training but removed during fine-tuning (second row of \cref{tab:fine-tuning-new}), in the fine-tuning stage
we use only $\mathcal{L}_{ce}$ for training but we 
do {\em not} re-introduce skip connections or LN layers in the last block
(i.e., we use Eq.~\ref{eq:skip-connection-removal-1}-Eq.~\ref{eq:skip-connection-removal-2}
 when fine-tuning).
Conversely, when  pre-training is done without SAR, which is introduced only in the fine-tuning stage (third row of \cref{tab:fine-tuning-new}),  when fine-tuning,
we use $\mathcal{L}_{tot} = \mathcal{L}_{ce} + \lambda \mathcal{L}_{se}$, but keeping the standard skip connections 
and the LN layer
in the last block (Eq.~\ref{eq:transformer-standard-1}-Eq.~\ref{eq:transformer-standard-2}). 

The models are fine-tuned for 100 epochs with a batch size of 512, and an initial learning rate of 0.0005 decayed with a cosine schedule.

\subsection{Self-supervised experiments} 
In our self-supervised experiments, we adopt the original code for MoCo-v3 \citep{MoCo-v3}\footnote{\url{https://github.com/facebookresearch/moco-v3}} and DINO \citep{DINO}\footnote{\url{https://github.com/facebookresearch/dino}} with a ViT-S/16 backbone. For computational reasons, we restrict our experiments training the models on IN-100 for 300 epochs.
Moreover, to fit the available computational resources, we reduce the batch size to 1024 for MoCo-v3, while DINO is trained with the default multi-crop strategy $(2 \times 224^2 + 10 \times 96^2)$, but with a  batch size of 512.
We thoroughly follow the authors' specifications for the other hyperparameters. 
The results in \cref{tab:MoCov3DINO} are obtained using a standard linear evaluation protocol in which the pretrained backbone is frozen, and a linear classifier is trained on top of it, using SGD for 100 epochs on IN-100.

\section{Dataset licensing details}
CIFAR-10, CIFAR-100 are released to the public with a non-commercial research and/or educational use\footnote{\url{https://www.cs.toronto.edu/~kriz/cifar.html}}. Oxford flower102 is released to the public with an unknown license through its website\footnote{\url{https://www.robots.ox.ac.uk/~vgg/data/flowers/102/}}, and we assume a non-commercial research and/or educational use. ImageNet annotations have a non-commercial research and educational license\footnote{\url{https://image-net.org/download}}.
PASCAL VOC 2012 images abide by
the Flickr Terms of Use\footnote{\url{http://host.robots.ox.ac.uk/pascal/VOC/voc2012/}}.
Stanford Cars images have a non-commercial research and educational license\footnote{\url{http://ai.stanford.edu/~jkrause/cars/car\_dataset.html}}
ClipArt, Painting and Sketches are part of the DomainNet dataset which is released under a fair use license\footnote{\url{http://ai.bu.edu/DomainNet/\#dataset}}.
The ImageNet-A\footnote{\url{https://github.com/hendrycks/natural-adv-examples}} and ImageNet-C\footnote{\url{https://github.com/hendrycks/robustness}} images are released with unknown licence, so we refer to the original authors to use these datasets.

\end{document}